# Exploring text representations for online misinformation

Dogo, Martins Samuel

# EXPLORING TEXT REPRESENTATIONS *for* ONLINE MISINFORMATION

MARTINS SAMUEL DOGO

April 2023

*Thesis presented for the degree of*
*Master of Philosophy*

*to the*

*School of Electronics, Electrical Engineering*
*and Computer Science*

QUEEN'S UNIVERSITY BELFAST

For Favour Dogo

# ABSTRACT

Mis- and disinformation, commonly collectively called *fake news*, continue to menace society. Perhaps, the impact of this age-old problem is presently most plain in politics and healthcare. However, fake news is affecting an increasing number of domains. It takes many different forms and continues to shapeshift as technology advances. Though it arguably most widely spreads in textual form, *e.g.*, through social media posts and blog articles.

Thus, it is imperative to thwart the spread of textual misinformation, which necessitates its initial detection. This thesis contributes to the creation of representations that are useful for detecting misinformation.

Firstly, it develops a novel method for extracting textual features from news articles for misinformation detection. These features harness the disparity between the *thematic coherence* of authentic and false news stories. In other words, the composition of themes discussed in both groups significantly differs as the story progresses.

Secondly, it demonstrates the effectiveness of topic features for fake news detection, using classification and clustering. Clustering is particularly useful because it alleviates the need for a labelled dataset, which can be labour-intensive and time-consuming to amass.

More generally, it contributes towards a better understanding of misinformation and ways of detecting it using Machine Learning and Natural Language Processing.

# ACKNOWLEDGMENTS

Above all, I am most grateful to the LORD, my rock, my fortress, and my deliverer. He is my King from of old. Palpable was His steadfast love, grace, and direction throughout my research.

I am greatly indebted to my supervisors, Dr Deepak Padmanabhan and Dr Anna Jurek-Loughrey, for their innumerable contributions and insights, patience, encouragement, supervision and intellectual support.

I express my gratitude to Dr Paul Miller and Dr Barry Devereux, the latter of whom served as the internal viva examiner, for providing me with constructive criticism and valuable recommendations during my fair share of Annual Progress Review meetings. I am grateful to the external viva examiner, Dr Seun Ajao, for taking the time to examine my thesis, and providing detailed feedback and suggestions, which have enhanced the quality of this work.

I am grateful for the help I have received from members of staff at EEECS. I would like to acknowledge the help of Dr Jesus Martinez Del Rincon and Professor Hans Vandierendonck. Special thanks to Mrs Katie Stewart for her unparalleled support and guidance throughout my time at the school. I am deeply indebted to Mrs Kathleen Ingram for her support, and help with proofreading my work.

In developing this thesis I have benefited from the constructive comments and warm encouragement of friends and colleagues at EEECS. I wish to thank Abdullah, Alimuddin, Ayesha, Maya, Michael, and Pritam. I enjoyed interacting and exchanging ideas with you all.

I am also immensely grateful to friends who have helped and supported me during this journey. Thank you, Abubakar, Akaoma, Chioke, Doreen, Michael, Victor, Victoria, and Udeme. My heartfelt thanks to Kigwab for your warmth and mirth.

Completing this thesis would not have been possible without the great personal support of my family. I am extremely grateful to my parents, who have always nourished and nurtured my curiosities, and to my siblings, for your profound belief in my abilities. You have contributed much more to my work than you realise.

Martins Dogo
February 2023

# CONTENTS

# LIST OF FIGURES

# LIST OF TABLES

## LIST OF ALGORITHMS

# ACRONYMS

ABM agent-based model

ADS American Dialect Society

BERT Bidirectional Encoder Representations from Transformers

BoW bag-of-words

CDT Context Dimension Tree

CNN Convolutional Neural Network

kNN $k$-Nearest Neighbour

LDA Latent Dirichlet Allocation

LIWC Linguistic Inquiry and Word Count

LSA Latent Semantic Analysis

LSTM Long Short-Term Memory

ML Machine Learning

NLP Natural Language Processing

NMF Non-negative Matrix factorisation

OED Oxford English Dictionary

OSN Online Social Network

PCA Principal Component Analysis

PCFG Probabilistic Context-Free Grammar

PoS parts-of-speech

RNN Recurrent Neural Network

RST Rhetorical Structure Theory

SVD Singular Value Decomposition

SVM Support Vector Machine

TFIDF Term Frequency-Inverse Document Frequency

tSNE $t$-distributed Stochastic Neighbor Embedding

UMAP Uniform Manifold Approximation and Projection

VAE Variational Autoencoder

Part I

# FAKE NEWS

# 1

## OVERVIEW

▸ THE IMPACT OF news on daily affairs is arguably greater than it has ever been. Its importance today can hardly be overstated. Although its ubiquity and influx make news readily available, the news—despite being mostly useful—is increasingly becoming skewed away from truth, towards more sensational headlines, as competition for readership becomes more difficult.[1]

The news plays many roles. It informs the populace, inspires hope, and initiates conversation, to name but a few. However, it sometimes leans towards rhetoric rather than facts, so it is not always reliable. The news can also overwhelm with its dizzying speed and endless breadth of topics. It can be argued that it would be almost impossible to detach the news from everything else. The world may not be able to function without it. Consequently, news can be leveraged for virtuous, or vicious activities.

The fabrication and dissemination of falsehood have become politically and economically lucrative endeavours, as well as tools for social and ideological manoeuvre. Thus, the present times have been labelled as a *post-truth* period. These endeavours have led to an intricately complex and constantly evolving phenomenon that is mainly characterised by *disinformation* (false information which is created or shared with malicious intent) and *misinformation* (also false, but shared with harmless intentions). The pair is commonly collectively referred to as *fake news*.[2]

[1] Wang (2012), Kavanagh et al. (2019)

[2] The uses and misuses of this term are discussed in more detail in §1.1.

▸ SEVERAL DOMAINS ARE affected by misinformation, especially in situations that involve or affect many people, where uncertainty and tensions are high, and resolutions are not forthcoming. These domains include but are not limited to:[3]

- Politics: misinformation has historically and globally played a consequential role in politics. A recent example of this is during the 2016 U.S. Presidential Elections.
- Healthcare: during an epidemic such as the Ebola disease outbreak in 2014, or a pandemic such as COVID-19, misinformation also spreads, particularly through social media.
- Natural disasters: examples include sharing falsified information (*e.g.*, following the Fukushima Daiichi nuclear disaster in Japan, in 2011); inade-quate (*e.g.*, during the earthquake in Nepal, in 2015); or mis-contextualised (*e.g.*, during an earthquake in Sicily, in 2014, whereby news of another one from 1908 was referenced).

[3] Allcott and Gentzkow (2017), U.S. Department of Homeland Security (2018), Waldman (2018), Wardle (2020), Chowdhury et al. (2021)

Central to this thesis is a computational investigation into some of the characteristics of fake news text that differentiate it from truthful news. The texts analysed in this work are in short and long forms—tweets related to news events and full-length news articles, respectively. The datasets cover diverse domains, including politics, sports, and conflict.

To better understand misinformation, it is important to first understand what news is. The Oxford English Dictionary (OED) (2022) defines[4] *news* as:

[4] This is the most relevant definition in the context of this thesis.

> *'The report or account of recent (esp. important or interesting) events or occurrences, brought or coming to one as new information; new occurrences as a subject of report or talk; tidings.'*

It is a somewhat subjective matter which events may qualify as important or interesting. However, it can objectively be stated, that the alteration of new information can distort a news report to the extent of falsifying events, thereby rendering the news false. Certainly, even if all the information is true and verified, the reportage can confuse or mislead, for instance, by means of rhetoric. It is clear then, that faithfully narrating a vapid event in an engrossing manner, does not constitute fake news, but perhaps is the result of skill or passion. On the other hand, regardless of how interesting an event is, its misrepresentation may misinform or disinform the reader. Therefore, it is important to categorise the various ways in which—and degrees to which—information can be falsified. This is because by doing so, a typology for distinguishing the different types of misinformation can be created. Such a typology helps to identify the specific kind of problem being dealt with, and in finding the optimum mitigation against it. In this thesis, a typology called *Information Disorder*, which captures the essence and full breadth of the mis- and disinformation landscape, is adopted. This is discussed in the next section (§1.1).

> *'The news knows how to render its own mechanics almost invisible and therefore hard to question.'*
>
> — Alain de Botton, "The News: A User's Manual"

▸ A PLETHORA OF sources now vie for the attention of readers—perhaps, more than ever before. As a result, editors and journalists may be incentivised to produce more sensational or emotionally charged pieces to invite, maintain or grow readership. This is not a critique of journalists, nor an assessment of their practices as measured against the principles and standards which apply to their field. Rather, it is simply an observation of a trend. In many fields, business and economic motives can clash with principles, and journalism is no exception. As discussed later in this chapter,[5] besides sensationalism and bias, advertising is one of the ways through which misinformation seeps into news pieces.

[5] See §1.1.3 and §1.3.2.

The lure of misinformation on social media typically begins with the title of an article—often heightened by accompanying photographs. In the so-called *attention economy*, attention is offered primacy because it is scant. With limited space, in adjacency with other publications, and within a publication itself, titles must therefore strive to be eye-catching. This is often done at the expense of high-quality information; at the same time, readers with limited attention tend to share tawdry information.[6]

[6] Menczer and Hills (2020), "The Attention Economy"

As for the main content of a piece, a compelling story is naturally more captivating than a list of factual statements. Facts tend to be either bland and predictable, and therefore boring—or strange and new, and therefore interesting. Most people possibly prefer the latter.

## 1.1 INFORMATION DISORDER

Mis- and disinformation are intertwined but essentially distinct phenomena. They form a part of the broader landscape of what's commonly, and often inaccurately, called 'fake news'. People, especially on the internet, have become accustomed to referring to the entire landscape of false information as fake news. Although the term suffices to indicate various types of false information and even sophistry in biased articles, its unbridled use is problematic.[7]

[7] A concise etymology of the term 'fake news' is related in §1.1.2, as well as examples of its misuse.

The misinformation landscape as a whole is so complicated, that there is currently no firm consensus on terminology, nomenclature, and definitions amongst researchers of the subject. Nonetheless, due to the acceleration of research in the area, the different types of fake news are becoming more firmly grouped.

*'When it was reported that Hemingway's plane had been sighted, wrecked, in Africa, the New York Mirror ran a headline saying, "Hemingway Lost in Africa," the word "lost" being used to suggest he was dead. When it turned out he was alive, the Mirror left the headline to be taken literally.'*

— Donald Davidson, "What Metaphors Mean"

Wardle (2017) was among the first to propose a typology of fake news. It consisted of seven main categories, in increasing order of harmfulness: satire or parody, false connection, misleading content, false context, imposter content, manipulated content, and fabricated content. The groups were based on three criteria: the type of information created and shared, the motivation behind the creation of the content, and how it is disseminated. Though it received some pushback, Wardle assiduously defended the inclusion of satire as a category in a revised edition of her typology.[8] Having acknowledged that satire (when intelligent) is a form of art, she explained that it is slyly used to veil canards and conspiracies, and thus divert the attention of fact-checkers. Moreover, should such a piece be later detected, its authors can simply claim that it was, after all, not intended to be taken seriously. Table 1.1[9] summarises the types of mis- and disinformation and their motivations, according to Wardle (2020).

[8] Wardle (2020), "Understanding Information Disorder"

[9] Adapted from Wardle (2020).

| TYPE | DESCRIPTION | POOR JOURNALISM | TO PARODY | TO PROVOKE/'PUNK' | PASSION | PARTISANSHIP | PROFIT | POLITICAL INFLUENCE | PROPAGANDA |
|---|---|---|---|---|---|---|---|---|---|
| Satire/Parody | No intention to cause harm but has intention to fool | ✓ | ✓ | | | | | | |
| False connection | When headlines, visuals or captions don't support the content | ✓ | | | | | ✓ | | |
| Misleading content | Misleading use of information to frame an issue or individual | ✓ | | | | ✓ | | ✓ | ✓ |
| False context | When genuine content is shared with false contextual information | ✓ | | | ✓ | ✓ | | ✓ | ✓ |
| Imposter content | When genuine sources are impersonated | | ✓ | ✓ | | | ✓ | | ✓ |
| Manipulated content | When genuine information or imagery is manipulated to deceive | | | ✓ | | | | ✓ | ✓ |
| Fabricated content | New content that is 100% false, made to deceive and do harm | | | ✓ | | | ✓ | ✓ | ✓ |

TABLE 1.1: The matrix of misinformation

### 1.1.1 *The ecosystem*

Wardle and Derakhshan (2017b) expand on Wardle's original typology in what can be regarded as one of the most in-depth explorations on the misinformation landscape to date. In this work, they present a conceptual framework that offers a useful perspective for understanding the misinformation ecosystem. In contrast to other authors, they use the term *information disorder* as a substitute for *fake news*, to encapsulate *mis-*, *dis-*, and, what they call *mal-information*—apt for conveying the mélange of problems faced in a post-truth world.

*'What had gone wrong was the belief in this untiring and unending accumulation of hard facts as the foundation of history, the belief that facts speak for themselves and that we cannot have too many facts, a belief at that time so unquestioning that few historians then thought it necessary-and some still think it unnecessary today-to ask themselves the question: What is history?'*

— E.H. Carr, "What Is History?"

Simply put, disinformation involves intentionally creating or sharing false information to cause harm. In other words, it contains deliberately and verifiably falsified information. Mal-information is genuine information shared with deceptive intent. Lastly, akin to a rumour, misinformation is false information but not originally intended to cause harm. This should not be confused with a *rumour*, which is 'an unverified or unconfirmed statement or report circulating in a community.'[10] A rumour may later be verified as true or false, whereas disinformation is false from the onset.

[10] Oxford English Dictionary (2021), *rumour | rumor, n.*

Manifold, acceptable definitions of basic terms can be found in the literature of misinformation; their misdefinitions can also be encountered. Therefore, paying close attention to definitions of terms related to the problem is critical. Otherwise, it may compound the problem. For example, Zubiaga et al. (2018) observed that Cai et al. (2014) and Liang et al. (2015) incorrectly defined a rumour.[11]

[11] Both define a rumour as false information, whereas the proper definition is as information whose veracity is not yet known.

Wardle further corraled a fairly comprehensive glossary in which she defined common terms and acronyms associated with the misinformation disorder landscape.[12] Attempts such as Wardle's, to clarify misinformation-related terms are crucial for aiding researchers and the general public in assimilating the scope of the problem of misinformation. Another such work is that of the media historian and theorist Caroline Jack, who created a lexicon for media content, aimed at educators, policymakers and others.[13]

[12] Wardle (2018), "Information Disorder: The Essential Glossary"

[13] Jack (2017), "Lexicon of Lies: Terms for Problematic Information"

### 1.1.2 *What is fake news?*

Fake news essentially means disinformation. It arguably is the term most widely used to refer to multiple categories of information disorder. Although the term was once used in a corrective and progressive manner,[14] its positive connotation has since split into a duality—it is now used to refer to disinformation, and critique and deride mainstream media.[15] Furthermore, 'fake news' is also used, ironically, to denounce or discredit factual information as misinformation. The phrase has recently become a tool for tactical subversion from truth and, especially in politics, for slander against dissenting opposition.

[14] Gelfert (2018), "Fake news: A definition"

[15] Wardle and Derakhshan (2017b), Caplan et al. (2018)

Documented uses of 'fake news' in writing date back to the 1890s; however, other terms denotative of misinformation go as far back as the 16th century.[16] Gelfert (2018) carried out an in-depth study of the etymology of fake news. The study lists examples of previous attempts at defining the phenomenon and why they are inadequate; it also gives historical examples of attempts to define fake news. The following definition from Gelfert is adopted in this thesis:

[16] Merriam-Webster Dictionary (2017), *How Is 'Fake News' Defined, and When Will It Be Added to the Dictionary?*

> *'Fake news is the deliberate presentation of (typically) false or misleading claims as news, where the claims are misleading by design.'*

The term 'fake news' may be unideal to refer to all kinds of misinformation. However, it is popular among the public and researchers of misinformation alike. Although 'fake news' may be a convenient catch-all term, it does not accurately reflect the nuances and complexities of misinformation. Therefore, it is crucial to exercise caution when using this terminology and consider alternative descriptors that may be more appropriate for specific contexts or types of disinformation.

As it is now used to denote the entire spectrum of information disorder, 'fake news' sometimes educes ambiguity. Understandably, most people will not be familiar with the minutia of a research area, no matter how relevant it is. Moreover, people may prefer simpler, more relatable terms for use in conversation. For these reasons, it may be permissible to call most kinds of misinformation fake news. However, the term is strictly inadequate and inaccurate. Misinformation need not even be news in the first place. It is essentially corrupted information. But if one generalises to say 'fake information', this is also problematic, because it negates correct information that is mistakenly shared with a false context.

▸ IN RESPONSE, SOME have proposed using the term 'false news' to refer to disinformation instead.[17] But what happens when those who use fake news in subversive ways also begin to use 'false news' in the same manner? Quite often, the intent of an actor who shares problematic information cannot be promptly proven or inferred. It appears, at least in research, that 'misinformation' is used as an umbrella term for information disorder. This is less obscure because although it does not strictly classify a piece of information, it still insinuates that the information is problematic.

[17] Oremus (2017), Habgood-Coote (2018)

In this thesis, 'fake news', 'false news' and 'misinformation' are used interchangeably, in a broader sense to refer to the scope of information disorder. Moreover, 'real', 'legitimate' and 'authentic' are used to refer to reliable and truthful news. The focus of this thesis is on finding an algorithmic solution to hindering the spread of fake news, and not its epistemology. The reader is referred to Tandoc et al. (2018), Torres et al. (2018) and Zannettou et al. (2019) for further in-depth studies of the typology and epistemology of fake news.

### 1.1.3 *A Brief History of Fake News*

While it is beyond the scope this thesis to expand on the chronology of fake news, some key events may serve to sum up its timelessness. This summary will centre on a few domains palpably affected by it: war, natural disasters, healthcare, and politics.

Fake news predates news itself, at least, news conveyed through newspapers. Dating back to the 17th century and originally called *newsletters*, newspapers were simply printed or handwritten letters used to exchange tittle-tattle. This activity grew and transformed into the production and consumption of modern newspapers.[18]

*'Since wit and fancy find easier entertainment in the world than dry truth and real knowledge, figurative speeches and allusion in language will hardly be admitted as an imperfection or abuse of it. I confess, in discourses where we seek rather pleasure and delight than information and improvement, such ornaments as are borrowed from them can scarce pass for faults.'*

— John Locke, "An Essay Concerning Human Understanding, Book III"

[18] Park (1923), "The Natural History of the Newspaper"

▸ LONG BEFORE NEWSLETTERS, however, the disinformation campaign had been a tactic in use. One example of note was in the Roman Empire. Following the demise of Julius Ceaser, Octavian and Antony launched disinformation campaigns against each other—employing propaganda, through the media of poetry, rhetoric and newly minted coins—in a bid to become emperor. This led up to the Battle of Actium, in 31 BC, out of which Octavian emerged the victor.[19] Though not its ultimate determiner, propaganda played a crucial role in the war. Misinformation has since been a prime weapon in the arsenal of warring entities. Or for inciting conflicts in the first place, as in the case of the Spanish-American War.[20]

The media and speed of disseminating fake news have drastically advanced. At the time of writing, Russia was continuing with its invasion of Ukraine. From the onset, the Russian state used various disinformation narratives to justify the invasion.[21] Its current model of propaganda is high-velocity and unremitting, high-volume and multichannel, and lacking in objective reality or consistency. This approach has been developing since the Soviet Cold War era—to Russia's invasion of Georgia in 2008—to its annexation of the Crimean peninsula in 2014—and it is, in all probability, now deployed in Russia's invasion of Ukraine.[22]

Clearly, then, misinformation has had an enduring influence on conflicts, but so has it on many other areas of life: another is natural disasters. During the 15th century the readership of news significantly expanded, thanks to the birth of the printing press. Fake news followed suit, expectedly.[23] After all, the original newsletters helped gossip to set sail. After an earthquake in Lisbon, in 1755, pamphlets containing fake news[24] were circulated around Portugal.[25] Today, there are various ways to fact-check news and other information. By contrast, fact-checking was a rarity then. In want of scientific understanding, several natural events, including natural disasters, were mystically interpreted.

▸ THE TERM 'INFODEMIC' was coined by Rothkopf (2003). It described the surge of information, true and false, related to the 2003 SARS epidemic. Mindful to not understate the severity of SARS itself, Rothkopf argued that the 'information epidemic' that resulted from it added a new and more worrisome dimension to the disease. Future global events would affirm Rothkopf's ominous piece. In the wake of the Coronavirus disease 2019 (COVID-19) pandemic, people all over the world frantically sought information and many hastily acted upon unverified information. To worsen matters, advice—including from the World Health Organization, governments, and other trusted and reputable was not only updated frequently, but at times, inconsistent. Naturally, some were stirred to doubt, anxiety, and confusion. Meanwhile, a steady flux of misinformation gushed out through Online Social Networks (OSNs) (also known as social media). This culminated in an infodemic. Its impacts included psychological issues, loss of public trust, loss of lives due to misinforming protective measures, and panic purchase.[26] In 2018, the Democratic Republic of Congo experienced multiple outbreaks of the Ebola virus. The adoption of preventive measures against it was hampered by misinformation and low institutional trust.[27] Other disease outbreaks such as the Zika virus, Middle East Respiratory Syndrome, and H1N1 Influenza (swine flu) were all adversely affected by misinformation.[28]

[19] Kaminska (2017), "A lesson in fake news from the info-wars of ancient Rome"

[20] Soll (2016), *The Long and Brutal History of Fake News*

[21] European External Action Service (2022),U.S. Department of State (2022)

[22] Paul and Matthews (2016), "The Russian "Firehose of Falsehood" Propaganda Model: Why It Might Work and Options to Counter It"

[23] Soll (2016), *The Long and Brutal History of Fake News*

[24] To be precise, this was a mixture of witnesses' accounts, false context and manipulated content. It brought forth a new genre of sensational news called *relações de sucessos*.

[25] Araújo (2006), "The Lisbon Earthquake of 1755 – Public Distress and Political Propaganda"

[26] Pian et al. (2021), "The causes, impacts and countermeasures of COVID-19 "Infodemic": A systematic review using narrative synthesis"

[27] Vinck et al. (2019), "Institutional trust and misinformation in the response to the 2018–19 Ebola outbreak in North Kivu, DR Congo: a population-based survey"

[28] Chowdhury et al. (2021), "Understanding misinformation infodemic during public health emergencies due to large-scale disease outbreaks: a rapid review"

▸ NEWSPAPERS HAVE HISTORICALLY carried misinformation—and occasionally, disinformation. Modern newspapers became more mainstream by the 19th century, and through them, true and false news travelled faster and farther. The news became more sensational too. For instance, in 1835, the New York Sun published multiple false articles claiming that there were aliens on the moon. This was known as the 'Great Moon Hoax'.[29] In the 1890s, Joseph Pulitzer and William Hearst, rival American news publishers, contended for a larger readership of their newspapers. Each sought to succeed by dubious practices—blatant reportage of rumours as facts. This practice was known as 'yellow journalism.'[30]

The dynamics of misinformation became more complex when news leapt from paper onto web pages. News became boundless, and so did misinformation. Meanwhile, sensationalism reigned on. By and by, news websites became interlaced with advertisements, and sometimes, *shockvertising* (*i.e.*, designed to shock and provoke).[31] And to increase advertisement revenue, some resorted to *clickbait*—attention-grabbing headlines designed to cajole readers into clicking links. Clickbait has taken up residence on the internet. To sell advertisements, 'drive traffic', 'increase engagement', or simply mislead, many websites resort to clickbait. It often misleads and can be acutely harmful. Yet, it remains inescapable on the web. In fact, it is believed that misinformation—largely in the form of clickbait shared on OSNs—influenced the outcome of the 2016 U.S. presidential election.[32]

Perhaps only coinciding with, rather than causing it, greater attention was being paid to misinformation, as the internet made strides. Or perhaps, misinformation simply grew too rapidly to be ignored. In 2005, the American Dialect Society (ADS) named *truthiness* its word of the year.[33] The Merriam-Webster Dictionary did the same the following year.[34] In 2016, the OED named *post-truth* its word of the year.[35] The next year, the ADS and the Collins Dictionary both announced *fake news* as their word of year.[36] In 2018, *misinformation* was named Dictionary.com's word of the year.[37]

[29] Soll (2016), *The Long and Brutal History of Fake News*

[30] Center for Information Technology & Society, UCSB (2022), *A Brief History of Fake News*

[31] Oxford English Dictionary (2022), *shockvertising, n.*

[32] Allcott and Gentzkow (2017), "Social media and fake news in the 2016 election"

[33] American Dialect Society (2006), *Truthiness Voted 2005 Word of the Year*

[34] Merriam-Webster Dictionary (2022), *What is 'Truthiness'?*

[35] Oxford Languages (2016), *Oxford Word of the Year 2016*

[36] Wright (2017), American Dialect Society (2018)

[37] Dictionary.com (2018), *Misinformation | Dictionary.com's 2018 Word of the Year*

▸ IT IS UNLIKELY that any medium for sharing information that is open to the general public will be immune to misinformation. To say nothing of sharing news. The minimal cost of creating accounts and posting, combined with economic and social incentives, particularly encourages bad actors.[38] One potential threat in the future could be the misuse of generative artificial intelligence. It is likely that deep fakes—in text, audio, image, and video forms—will become more sinister. At any rate, they are becoming more realistic.

While exacerbating misinformation, the internet, at the same time, may be the most effective tool for stifling it. Especially through the strategic use of OSNs. Inoculation or 'prebunking' is one such strategy. This means pre-empting oncoming information with facts. According to Pilditch et al. (2022), inoculating a critical mass of users in a network can inhibit the consolidation of falsehood.[39] Their experiments were carried out using agent-based models (ABMs). While their results are promising, it should be borne in mind that their setup, as well as

[38] Shu et al. (2017), "Fake News Detection on Social Media"

[39] They found that inoculating subsets of users at different times is also effective.

ABMs, are simplifications of the real-world. The same limitation applies to several other proposed approaches for stopping misinformation, including ML-based ones. Nonetheless, such research should be spurred on.

Via the internet, suspicious information can be scrutinised in near real-time. Likewise, facts corrective to them can be dispensed quickly. Indeed, on the internet, registers of facts abound and are at hand for swift withdrawal. But fake news is multiplicative. For regarding a single fact, countless false narratives can sprout up. Therefore, it is easier for lies to accrete than for truth to flow. Another strategy, nevertheless, is education: on how to spot and retard misinformation, and how to seek and interpret facts.

It would seem that society is in an endless battle with misinformation; that it may take one form or another, but can never be fully eradicated; and that it will always be one step ahead of the safeguards in place. This could be true as long as there remains insistence on velocity and volume rather than clarity and nuance, and on flimsy metrics as the measure of the effectiveness of communication. All these could come true as long as there remains insistence on velocity and volume rather than clarity and nuance, and on flimsy metrics as the measure of communication. Not every problem can be solved by a technological breakthrough, or simply more information, no matter how factual it is. Especially those problems entangled with people's identities, tightly held beliefs and opinions, and their daily lives and bread. It may be necessary to rethink the design of communication tools (both small and large scale), some of the incentives for these communications, and the online communities that foster them. Misinformation will like become an increasingly thorny issue in the future, and it is, therefore, crucial to think outside the box to find effective solutions.

## 1.2 FAKE NEWS ON SOCIAL MEDIA

Social media platforms provide a medium where the production and sharing of news is not limited to established news agencies, but also open to the general public.[40] News agencies used to be the main *creators* and *distributors* of news. Today, however, the general public is a lot more involved in that process.[41] In fact, so-called *content creators* (i. e., people from various fields who create media content for consumption, primarily on the internet) are thriving, particularly in technology news. Furthermore, people of all age groups and from all parts of the world interact, share and exchange information on OSNs. This makes it a suitable medium for rapidly spreading misinformation. Satisfactory solutions to counteracting this challenge have not yet been found.

[40] Campan et al. (2018), "Fighting fake news spread in online social networks: Actual trends and future research directions"

[41] Advances in smartphone and web technologies, now allow events to be broadcasted by members of the public with great speed and quality.

To sum up, misinformation on social media must be tackled. Given that this problem is multifaceted and dynamic, the ideal solution would equally be holistic and dynamic in its workings. Given its complexity, it must be approached pensively and with nuance. It is highly unlikely that the solution will be simple, if there is one at all—misinformation is a 'wicked problem'.[42]

[42] Rittel and Webber (1973) formulated the idea of a 'wicked problem' (in social policy) as one that is onerous or insoluble, characterised by 10 features, which Conklin (2006) generalised to the following six:

1. It is understood after a solution is developed.
2. It has no stopping rule.
3. Its solutions are not right or wrong.
4. It is novel and unique.

▸ WHETHER IN RESEARCH or deployment on the web, a multidisciplinary approach is ideally needed to combat misinformation. It has traditionally been combatted through manual fact-checking by experts. In addition to news agencies, other fact-checking organisations such as FactCheck[43], Snopes[44] and PolitiFact[45] employ such experts. Recently, however, computational techniques such as ML have been used to detect misinformation.

[43] https://www.factcheck.org
[44] https://www.snopes.com
[45] https://www.politifact.com

This potentially allows for automated, real-time detection which can alert people whilst or after engaging with misinformation. Furthermore, it can help in identifying social media accounts that spread misinformation. A lot of research work has been done in this area, but there remain limitations which hinder their application in real-world scenarios. One such limitation is the need for large news datasets annotated by experts.

## 1.3 WHY MACHINE LEARNING AND ONLINE SOCIAL NETWORKS?

### 1.3.1 *Access and participation on OSNs*

News is ubiquitous on OSNs and people access news through them. According to a survey by the Pew Research Center in the United States, 53% and 48% of US adults got their news from OSNs in 2020 and 2021, respectively.[46] The United Kingdom's Office of Communications (Ofcom) stated in its 2021 report on nationwide news consumption, that about half of adults in the UK access news on social media.[47] This trend transcends the Anglosphere. The Reuters Institute for the Study of Journalism at the University of Oxford, which aims to understand global news consumption, has been publishing its *Digital News Report* for the past decade. Its research focuses on countries with a high internet penetration and the 2021 report covered data from 46 countries across five continents. The report focused on the six largest OSNs—Facebook, YouTube, Twitter, Instagram, Snapchat, and TikTok—according to weekly use. In it, the Reuters Institute found that more than half of the Facebook and Twitter users surveyed encountered news on those platforms in the past week; for other networks, less than half of the users did.[48] They also found that for many Facebook users, the encounter with news on the platform is incidental rather than intentional. In fact, some people report avoiding it altogether.

[46] Pew Research Center (2021), *News Consumption Across Social Media in 2021*
[47] Ofcom (2021), *News consumption in the UK: 2021*
[48] Reuters Institute (2021), *Digital News Report 2021*

It should be expected that more people seek and find news on social media. After all, news sites and blogs share, and nudge people to share content on social media. Social media is apt for aggregating news from various sources, as well as for commentary and discussion. Furthermore, people themselves, now *create* news online by directly posting onto their profiles. In other words, social media activity sometimes *is* the news itself. Therefore, the creation of news is becoming more democratised and social media is continuously being reinforced as the global nucleus of news activity—from witnessing to disseminating, to assimilating.

However, along with this new-found voice and power follow ramifications. Most inimically, noise and lies compete with signal and truth, for space and attention.

▸ APART FROM READING news, people are generally spending more time on social media. It is also used to interact with friends and strangers, engage in public discourse or dissent, and more recently, to shop and donate to charitable causes, directly. It is not an overstatement, then, to say that social media has accrued an enormous value—or cost, depending on how one sees it—for nearly everyone. Misinformation arguably spreads the fastest on OSNs amongst news media. Now, if people's lives and livelihoods continue to be intricately intertwined with social media—if people are to find ways of navigating, or escaping, the real-world through it; to stay in touch and make new friends; to form and maintain communities and identities; find self-expression; to share memes and commiserate with one another—then it is worth protecting. Especially when it influences real-world events and politics. One of the consequences—or benefits, as the case may be—of wallowing in social media feeds is that it gradually shapes one's worldview. The design and resulting dynamics of OSNs make their users susceptive of a myriad of biases—information, political, cognitive, *etc.*[49] Fake news detection is currently mostly done by human experts. This is very expensive and time-consuming given the deluge of misinformation that parades OSNs daily. This work contributes to lessening the cost and effort spent by experts.[50]

[49] Menczer and Hills (2020), Barrett et al. (2021)

[50] See §1.3.3.

### 1.3.2 *Are OSNs doing enough to curb misinformation?*

At the ever-rising speed and scale of misinformation dissemination on social media, and considering that more and more people are reading news on them, the problem is proving to be insurmountable for human experts alone to deal with. The situation is critical, and the skills and resources needed for repair are limited. However, ML algorithms can augment the effort of experts combatting the problem. An example of how this can be done is explained in the next subsection (§1.3.3). Beyond intercepting misinformation, algorithms, more generally—as can be seen in this thesis and some of the works cited in it—are extending the capacity for unravelling the tangle of misinformation. A collection of algorithms, therefore, can act both as tools and as catalysts, matching the speed and scale at which misinformation propagates on OSNs and its complexity.

▸ WHILST EMPLOYING PEOPLE to spot problematic content including misinformation and false news ensures detection accuracy, this has been found to have detrimental effects on the moderators of social media content.[51] Firstly, repeated exposure to the kinds of disturbing media moderators scour out, can corrode a person's mental well-being. Besides that, reading false information repeatedly can lead one to believe it is true; this is a phenomenon called the *illusory truth effect*.[52] Finally, in spite of their invaluable contributions, content moderators are rather stingily remunerated for their work.

[51] Newton (2019), *The secret lives of Facebook moderators in America*

[52] Pennycook and Rand (2021), "The Psychology of Fake News"

In the case of Facebook, moderators are paid as low as $1.50 and $15 per hour, in Kenya and the United States, respectively. In both cases, these people are employed by contractors and not directly by Facebook. However, Facebook's own employees audit their work and periodically visit the contractors' offices for monitoring. Nonetheless, their pay is meagre and they are treated poorly, all in sharp contrast to the median salary of $240,000 and numerous additional perks, which Facebook employees enjoy.[53][54] Content moderators have reported struggling with mental trauma and indeed, some have been diagnosed with traumatic stress disorders. This is supposedly triggered by the appalling content they review. However, they have also reported facing intimidation and overwhelming pressure from their managers at work.[55] This compounds their work-related stresses rather than alleviating them. In 2019, a Facebook content moderator passed away, at work, at his desk. The management of the contracted company initially responded by dissuading their employees from discussing the tragedy, because they worried that it would dwindle productivity.[56] These findings raise some serious questions about the earnestness of social media platforms in fighting misinformation.

[53] Newton (2019), Perrigo (2022)

[54] This comparison does not mean to suggest that moderators should receive equal pay to other employees. (Although they definitely should be paid more and treated better.) Rather, it serves to elucidate that their role is regarded as subservient to those of others.

[55] Ibid.

[56] Newton (2019b), *Bodies in Seats: Facebook moderators break their NDAs to expose desperate working conditions*

Misinformation is a dynamic and convoluted problem. So much so that it seems misinformation will never be totally eradicated—but will always take one form or another—and can only be repeatedly extinguished. Such a volatile and amorphous nature demands supervision and intervention by experts. It is clear to see, then, the long-term significance of content moderation on OSNs, and the internet as a whole.

▸ IT WOULD BE unfair to social media companies if their efforts in combatting misinformation were not recognised. They have undertaken and funded numerous projects and initiatives, which demonstrate a sincere concern for the safety of their users. These also throw light on the multifaceted nature of the problem at issue.

Firstly, OSNs provide access to data for research purposes, through APIs or competitions. Research activities such as fake news detection using ML will not be practical without datasets, though some researchers have expressed a demand for additional data, *e.g.*, impression data.[57] In addition to data, OSNs platforms support researchers with grants. Therefore, notable contributions are made in support of research activities.

[57] Pasquetto et al. (2020), "Tackling misinformation: What researchers could do with social media data"

Secondly, OSNs are making it easier for people to flag or report posts they deem ill. Twitter, for instance, has taken this one step further through their Birdwatch pilot programme.[58] Users (in the U.S., for the time being) can directly annotate tweets they believe to be misleading, thereby providing context for the flag. These notes are publicly viewable by anyone. Beyond simplistic binary labels, this will provide more insight to Twitter users and researchers alike, for understanding the roots of misinformation.

[58] Coleman (2021), *Introducing Birdwatch, a community-based approach to misinformation*

Thirdly, these companies have established coalitions and partnerships with academia, media, fact-checking and other organisations, to work together to-wards achieving shared goals for the public good. Notable examples of such consortia include Social Science One,[59] the Content Authenticity Initiative,[60] and the Coalition for Content Provenance and Authenticity.[61] An outstanding, individual example is the Google News Initiative,[62] which boasts more than 7,000 partnerships and $300 million in funding to various organisations in over 120 countries.

[59] https://socialscience.one

[60] https://contentauthenticity.org

[61] https://c2pa.org

[62] https://newsinitiative.withgoogle.com

In addition, OSNs also:

- build in-house tools for detecting misinformation; they also incorporate new tools and expertise through company acquisitions (*e.g.*, Fabula AI being acquired by Twitter,[63] and Bloomsbury AI by Facebook[64])
- create robust, independent and transparent decision-making structures, which include external experts (*e.g.*, the Oversight Board[65] established by Facebook in 2018, which oversees critical content moderation on Facebook and Instagram)
- try to adapt their policies to current affairs and adhere to government policies around the world

[63] Agrawal (2019), *Twitter acquires Fabula AI to strengthen its machine learning expertise*

[64] Winick (2018), *Facebook's latest acquisition is all about fighting fake news*

[65] https://www.oversightboard.com

All that is mentioned here is not an exhaustive list of measures taken by social media platforms. But are they doing enough? While some of their efforts are commendable, there are areas where OSNs ought to improve.

▸ IT IS HELPFUL to constantly bear in mind that profit—primarily through advertising—is a top priority for social media companies. Notwithstanding the Google News Initiative's generous funding to various organisations, a noteworthy detail is that Google itself has a news product, Google News, which helps to drive user engagement with other Google products, such as search. Moreover, as of 2018 news accounted for 16-40% of Google Search results, and content crawled and scraped from news publishers drew in an estimated $4.7 billion according to the News Media Alliance (2019). Debate continues as to whether OSNs should reward publishers for their images and text which appear in search results, or if the publishers are better off for the additional web traffic. Publishers initially received nil from OSNs for their content, but this is no longer the case.[66]

[66] Google France (2021), *Le blog officiel de Google France: L'Alliance de la Presse d'Information Générale et Google France signent un accord relatif à l'utilisation des publications de presse en ligne*

Are OSNs willing to come up with tougher policies, which may hinder misinformation at the expense of some profit? Misinformation is common in advertising. According to Chiou and Tucker (2018), advertising makes a significant contribution to the spread of misinformation. To give some perspective, the U.S. Federal Trade Commission filed more than 150 instances of misinformation in adverts between 2015 and 2020; the settlements were as high as $191 million.[67]

[67] Fong et al. (2021), "Debunking Misinformation in Advertising"

▸ ALL IN ALL, users have a role—perhaps the biggest role, individually and collectively—to play in curbing misinformation. After all, users—businesses and individuals—generate most of the content on social media. In fact, according to Vosoughi et al. (2018), real human accounts *not* bots, are mostly responsible for sharing misinformation on Twitter.

### 1.3.3 *Alleviating the strain of labelling*

Supervised ML models for detecting fake news rely on labelled data. As such datasets are usually large, labelling them can be tedious. This process is expensive, exhausting and in some cases, detrimental to the well-being of those carrying out the task. These issues, as well as the low wages paid for labelling tasks, are discussed in the previous subsection. Further potential problems with labelling are that it does not scale, and it has an element of subjectivity.

Unsupervised ML models, on the other hand, do not rely on labelled data. Therefore, it can help to alleviate some of these issues. It would be ideal to minimise the effort required for labelling data while maintaining accuracy. In that sense, therefore, this thesis explores unsupervised learning as an alternative to supervised learning.

### 1.3.4 *Algorithms are versatile and catalytic*

Will ML algorithms someday be able to speedily and single-handedly spot every problematic content on OSNs? This is unlikely, for there will always be many borderline cases, and even humans sometimes disagree on how content should be classified. However, when the economic and psychological costs of human reviewing are considered, ML can make significant contributions to curbing problems such as information disorder. Moreover, it has, by and large, successfully been used to tackle other issues such as nudity on OSNs.

The scale of OSNs make them fertile grounds for the rapid spread of false news, with billions of people actively using them. History shows that OSNs have a revolutionary power. For instance, social media played a critical role in the Arab Spring of 2011[68], and more recently, in the 2016 U.S. Presidential Elections[69]. Further, OSNs are environments from which new culture (*e.g.*, memes) permeates into the real world, and they, therefore, influence the lives of individuals. As such, it is important to rid it of harmful actors and behaviours such as misinformation. Fortunately, the availability of datasets on false information in OSNs makes research on combatting the issue with algorithms feasible.

68 Brown et al. (2012), *The Role of Social Media in the Arab Uprisings*

69 Allcott and Gentzkow (2017), "Social media and fake news in the 2016 election"

## 1.4 PROJECT AIM

Existing implementations of semi-supervised and unsupervised ML are fewer and less varied than those of supervised learning. Work has been done aplenty in the supervised learning space, and good progress has been made. However, there are limitations which restrict its applicability, such as the need for labelled data.

There is also a need for new text representations that are robust for detecting misinformation. For example, it is common to use text features based on writing style. However, they may not be robust enough to identify false news written in a similar style to real news.

▸ THE AIM OF this research is to develop a novel approach for generating text representations from short and long-form texts. Furthermore, it aims to demonstrate the efficacy of such representations for misinformation detection, using unsupervised and supervised ML.

First, in Chapter 2, this thesis explores existing ways of utilising text features for misinformation detection. Second, in Chapter 3, experiments exploring how to harness text representations to detect fake news are presented. Chapter 4 introduces the concept of *thematic coherence*, based on analyses of topic features in news pieces. Finally, Chapter 5 shows results for detecting misinformation with topic representations using clustering and classification.

### 1.4.1 *Contributions*

Given their influence and harmfulness, a lot of research work has been done to address the elements of information disorder. This research focuses on mis- and disinformation. It mainly contributes to the existing body of work on misinformation detection using Natural Language Processing (NLP) and ML.

Firstly, an exploration of features for misinformation detection is carried out. A novel feature extraction approach, involving topic modelling, for classifying and clustering news articles is presented. Topic-based features are advantageous in situations where labelled data is difficult to acquire, available in a small quantity or non-existent. Additionally, topic features may be more robust when faced with machine-generated fake news, unlike the commonly used stylometric ones.[70]

Secondly, supervised and unsupervised ML methods are applied to detect misinformation, in multiple cross-domain datasets.

Lastly, the findings of this research may be applicable in other problem areas on the spectrum of information disorder in news text, *e.g.* hate speech detection. Also, this research more broadly contributes to the field of NLP. The experiments carried out and their results may be informative to other researchers in the field. The code for all the experiments presented in this thesis is available at `https://github.com/m-arti/mphil`.

[70] Schuster et al. (2020), "The Limitations of Stylometry for Detecting Machine-Generated Fake News"

# 2

# RELATED WORK

As information disorder continues to evolve, so too do surveys of research efforts in combatting it constantly diversify. This is markedly the case for scientific approaches. This diversity in perspectives and approaches indicates the complexity of information disorder. It also signifies the necessity for a holistic view of the problem.

## 2.1 MISINFORMATION DETECTION WITH MACHINE LEARNING

Existing approaches to misinformation detection using news text data generally involve two subtasks: feature extraction, and learning and classification.

### 2.1.1 *Feature extraction*

Feature extraction is a process by which attributes of news items on OSN posts are extracted and processed for classification. Shu et al. (2017) categorised these features into two groups, based on: (i.) news content, *i.e.*, text and image features; and (ii.) social context, *i.e.*, features based on users, posts, and networks.

Numerous papers use text features to detect fake news. Being the feature of interest in this work, this is expanded on in §2.2 and §2.3. In addition to images, videos and speeches are also used to extract features for fake news detection. Multiple features can be combined for detection, *i.e.*, in a *multimodal* fashion. Alam et al. (2022) surveyed multimodal fake news detection. Similarly, Cao et al. (2020) gave a comprehensive overview of the role visual content plays in fake news detection, while Shu et al. (2020b) did the same for user profiles. Zhou and Zafarani (2019), and Shu et al. (2020c) demonstrate the application and efficacy of network-based features.

### 2.1.2 *Learning and classification*

An ML model is then trained using the extracted features, to classify new, unseen news items or posts. The training process can be:

- Supervised: data with labels (typically 'real' or 'fake') are applied to train a classifier, *e.g.* neural network, decision tree, Support Vector Machine (SVM), *etc.*

- Semi-Supervised: this approach primarily aims to attenuate reliance on labelled data, which may be insufficient. It leverages unlabelled data to make predictions with higher accuracy than would have been attained using only labelled data.[71] Commonly known examples include:[72]
  - generative models, which initially learn features from a given task and are afterwards used in other tasks.
  - proxy-label methods, which utilise a model trained on a labelled dataset to generate more training data by labelling examples of unlabelled data.
  - graph-based methods, which model labelled and unlabelled data as nodes in a graph and try to propagate labels from the former to the latter.

  Semi-supervised learning also allows for a human-in-the-loop detection process. Some of the data are unlabelled in this case. An example is *active learning*, whereby labels for the most ambiguous training examples are sought from a human—a content moderator, for instance—to progressively improve the classification accuracy.

- Unsupervised: all data are unlabelled in this case. The task could be one of clustering, or anomaly detection whereby a fake news item is picked up as an outlier in the dataset.

Though wanting in visualisation, the overview of information disorder given by Zannettou et al. (2019), which is based on the work of Wardle (2017),[73] is sufficiently encompassing. It includes the various types of false information, its actors, as well as their motivations. Furthermore, works that analyse how false information propagates via different OSNs, as well as those which focus on how to detect them, are discussed. By comparison, Wardle and Derakhshan (2017b) equally gives a comprehensive view of the ecosystem—additionally, with a visual illustration to better the reader's understanding—of the types, actors, motives and phases of misinformation. However, a similar illustration for existing computational approaches is lacking in the literature.

Figure 2.1 shows a taxonomy of the different methods used to detect fake news and the sub-classification of tasks within each. In the following subsection, examples from the literature of each method and features utilised will be discussed.

[71] Ouali et al. (2020), "An Overview of Deep Semi-Supervised Learning"

[72] Ibid.

[73] See §1.1.

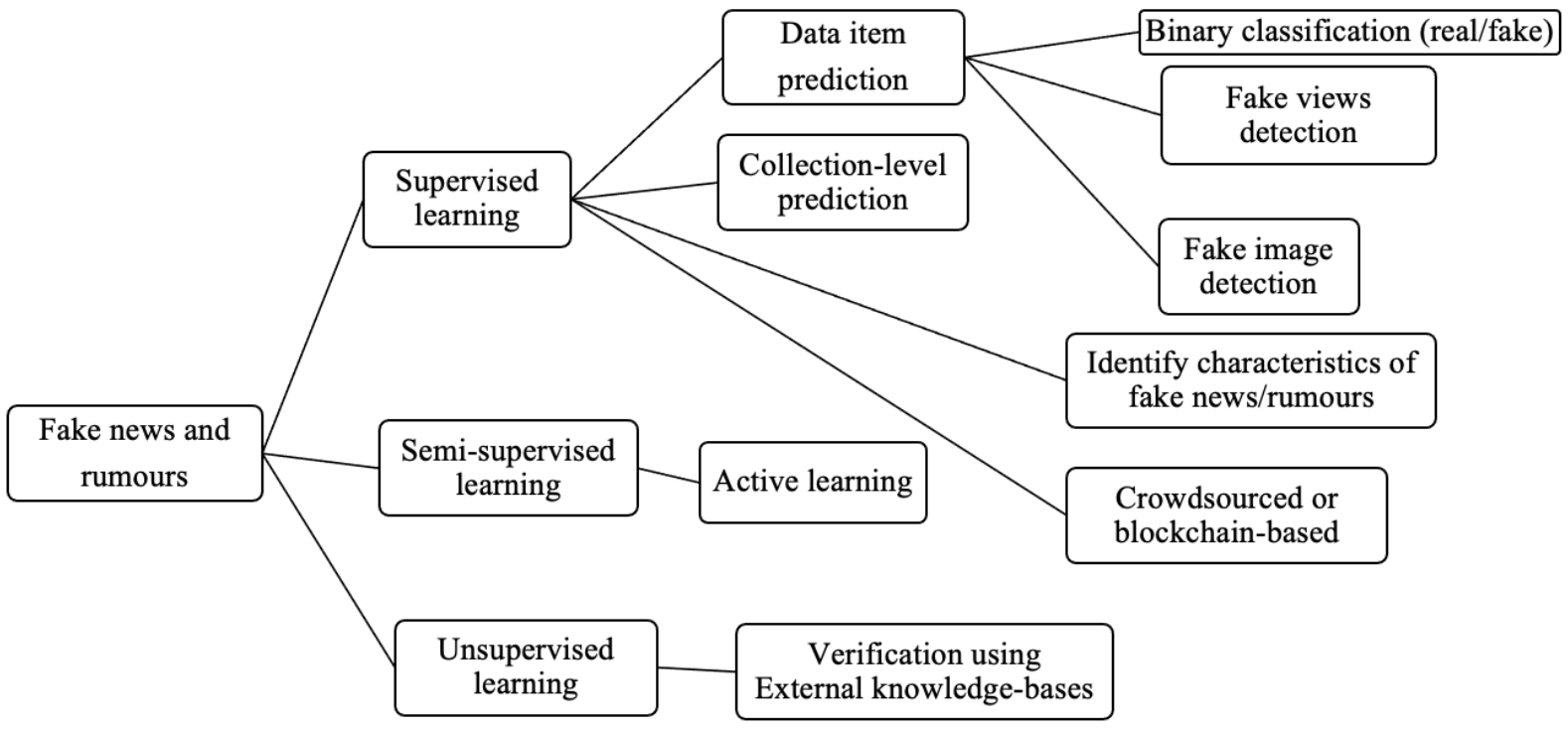

FIGURE 2.1: Fake news detection taxonomy.

## 2.2 TEXT REPRESENTATIONS FOR MISINFORMATION DETECTION

Although today it is produced and consumed in various media, news has pre-eminently been circulated in textual form. Over time, textual news developed into a general ossified structure:

- Source: the author and/or publisher of the article.
- Headline/title: this typically is a short sentence, descriptive of the principal news topic covered in the article.
- Body: the main text of the article, detailing the news story.
- Image/video: visual (or audio-visual) cue(s) included in the body of the article.

▸ IN AN IN-DEPTH study of the structure of news, van Dijk (1983) described the news as having two kinds of structure: thematic and schematic. The former represents the topical contents of a news item, while the latter describes the structure of the item's discourse. In other words, a news item is composed of themes bound together by a schema. These themes may vary (in nature and style of presentation) from one article to another, but the schema is firmly established.[74]

Online news presumably inherits its structure from that of traditional, printed news. Though the visual layout may be notably different. For instance, news on the web is typically laid out in a single, rather than multiple columns, has 'share' buttons, *etc.* However, its schema is identical to that of print news. Implicit in this schema, is a top-to-bottom outline of the news content, ranked from the most to the least important or newsworthy fragments.[75] It is conventional for journalists to produce—as it is for readers to assimilate—news in this manner. It is likely,

[74] van Dijk's study and conclusions were based on an empirical study of global press coverage of the 1982 assassination of the Lebanese president-elect, Bachir Gemayel. It covers 700 news articles published by 250 newspapers from 100 countries.

[75] van Dijk (1983), "Discourse Analysis: Its Development and Application to the Structure of News"

therefore, for the most relevant textual content for news analysis to be found closer to the top, rather than at the bottom of the news item.

Although published nearly four decades ago—long before the dawn of news detection using NLP and ML—van Dijk's paper provides some interesting and practical insights that could inform its current modus operandi. For instance, he gleaned from his study, that:

- the paramount topic of a news item is captured in its headline; and
- the opening sentences and paragraphs form the top of the schema—containing crucial details such as the time, location, parties, causes and outcomes of the main news events.

To summarise, the hierarchical structure of news embodies a linearly decreasing ordering of thematic information in a news item, from top to bottom. van Dijk attributes this order to "an implicit journalistic rule of the news organization." As news is written, so it is read. Thus, one can catch the scope of a news item by simply reading the headline and the main details in the opening section. This is true for most authentically produced news articles that adhere to journalistic standards. However, news produced with bad intent can exploit the hierarchy of its content for mal-intent.

Some have attempted to take advantage of this hierarchy for misinformation detection,[76] while most works using text data have relied on the body of articles. One of the main contributions of this thesis is that it presents a way to harness the inherent schematic of news, for detecting misinformation.

[76] For example, Biyani et al. (2016) used features, including titles, extracted from news webpages to detect clickbait; Sisodia (2019) extracted features from headlines to do the same; while Yoon et al. (2019) assessed the congruity between news headlines and body texts.

## 2.3 USE OF TEXT IN ML-BASED MISINFORMATION DETECTION RESEARCH

Whereas photos and videos were once merely accompaniments to news pieces, they are gradually taking centre stage in news dissemination, especially on OSNs. Nonetheless, text remains the predominant and most abundant form of news. Similarly, misinformation proliferating through social media and the web is typically in the form of text, and photos or videos are only recent developments. Besides, text can be extracted from news items disseminated in pictorial or video forms for analysis. For example, text extracted from a news video through speech-to-text technology can be used for NLP analysis. Text can similarly be extracted from photos. Expectedly, research in misinformation detection has mostly utilised text data as raw material, and NLP and ML techniques for extraction, enrichment and categorisation.

This research explores text representations for detecting online misinformation. In other words, it aims to find effective means of transforming text data into meaningful representations that can be used to characterize or identify fake news.

This section describes an overview of some key papers on text representations for misinformation detection. Most papers selected for this discussion focus their approach on two main strategies for exploiting textual data:

1. text-based features, generally extracted from the body text.
2. the schema of news: *i.e.*, papers which exploit features from a specific portion of news articles, such as headlines.

This section will focus on text-based features used for misinformation detection using ML. Shu and Liu (2019b) categorises such features into three groups: (i.) linguistic, (ii.) low-rank, and (iii.) neural text features. More elaborately, Zhou and Zafarani (2020) additionally categorise linguistic features into four groups: (i.) lexical, (ii.) syntactic, (iii.) discourse, and (iv.) semantic.

▸ LINGUISTIC FEATURES AIM to capture the *style* of writing in a piece of text. From this style, intent may be inferred (*i.e.*, whether to mislead or not),[77] or characterisation can be made (since fake news will likely have a style that differs from that of authentic news).

The following are some linguistic features and their applications for fake news detection:[78]

- Lexical features are generally concerned with the tallies or frequencies of character- or word-level features. Examples include *n*-grams, bag-of-words (BoW) methods, and Term Frequency-Inverse Document Frequency (TFIDF), which captures the relevance of a given word to a document in a corpus. Another is the Linguistic Inquiry and Word Count (LIWC), which calculates what percentage of words in a text fall into one of many categories, which indicate emotional and psychological properties, amongst others.
- Syntactic features are typically sentence-level features, including counts of punctuations, words, phrases, parts-of-speech (PoS) tagging, and Probabilistic Context-Free Grammar (PCFG) parse trees. Additional examples of these features are those specific to the news domain, such as quotations and links.
- Discourse features include applications of the Rhetorical Structure Theory (RST) and rhetorical parsers to extract rhetorical features from sentences.
- Latent features are primarily embeddings created using deep neural networks such as Convolutional Neural Networks (CNNs), Recurrent Neural Networks (RNNs) (particularly, using the Long Short-Term Memory (LSTM) architecture), and Transformers. These embeddings are dense vector representations of text at the word (most commonly), sentence, or document level. Commonly used word embedding models include `word2vec`[79], and more recently, transformer-based architectures such as BERT[80] and its variants. Another latent feature that is particularly relevant to this work (in Chapter 4) is the topic feature, extracted using topic modelling. Topic models identify themes latent in a group of documents by analysing the distribution of words and/or phrases across them.[81]

[77] Zhou and Zafarani (2020), "A Survey of Fake News: Fundamental Theories, Detection Methods, and Opportunities"

[78] Shu and Liu (2019b), Zhou and Zafarani (2020)

[79] Mikolov et al. (2013), "Distributed Representations of Words and Phrases and their Compositionality"

[80] Devlin et al. (2019), "BERT: Pre-training of Deep Bidirectional Transformers for Language Understanding"

[81] Blei (2012), "Probabilistic topic models"

Casillo et al. (2021) used a combination of topics, syntactic, and semantic features from news texts in three datasets to detect misinformation. They obtained the topic features using the LDA topic model. LDA is also used in this work and an in-depth explanation of its workings is given in §4.5.2. Stopwords were removed before feature extraction.[82] They used three syntactic features: (i.) the number of characters; (ii.) the Flesch Index, which is a measure of text readability; and (iii.) the Gunning Fog Index, which estimates text comprehensibility. These features are further processed using the Context Dimension Tree (CDT), which aids the selection of topics using temporal context. Next, they incorporate two semantic features—the probabilities of negative and positive news sentiment. Finally, the features are fed into a $k$-Nearest Neighbour (kNN) classifier for detection.

[82] stopwords are words such as *'just'*, *'do'*, and *'it'*, which are non-descriptive, and therefore, relatively less insightful with regard to generating or interpreting topics.

Another work that uses topic features is by Hosseini et al. (2022). Similar to the previous work, an LDA topic model was used to extract features. Before this, though, the texts are preprocessed into tokens, and non-English words and stopwords are removed. Word embeddings are obtained from the original news texts using the `word2vec` model. These embeddings are input into a bi-directional LSTM Variational Autoencoder (VAE) to form latent representations of the texts. The VAE representations are combined with the LDA topic representations to form the final features for classification. The combined features improved misinformation detection for classifiers compared to individual features.

Topic features can be extracted from non-English news texts. They can also be used for tasks other than detection. For example, Paixão et al. (2020) used BoW, word embedding, LIWC, PoS, and TFIDF features to differentiate between real and fake news in a Brazilian Portuguese news corpus. However, they further employed topic modelling to qualitatively study the two groups of articles in the dataset. They found the optimal number of topics to analyse using the coherence measure. This is also used in this work, although in a different way, in §4.7.1.

LDA is not the only topic modelling method available, but it is more commonly used in the literature. Ajao et al. (2019) experimented with a different method called Latent Semantic Analysis (LSA), but found LDA to perform better. They applied topic modelling to determine the 10 most prevalent topics in rumour and non-rumour tweets. The sentiment (positive, negative, or neutral) values of the words in each topic were then computed and used to calculate an *emotional ratio score*. This score was combined with linguistic features such as counts of user mentions, hashtags, and quotations, to form features for rumour detection.

Table 2.2 shows some of the commonly used text-based features for misinformation detection and examples of papers wherein they are implemented.

| TEXT FEATURE | PAPERS IMPLEMENTED IN |
|---|---|
| Lexical | BoW: Paixão et al. (2020), Zhou et al. (2020b);<br>LIWC: Pérez-Rosas et al. (2018), Paixão et al. (2020);<br>*n*-grams: Biyani et al. (2016), Ahmed et al. (2017), Potthast et al. (2018);<br>TFIDF: Biyani et al. (2016); Pérez-Rosas et al. (2018)<br>Others: Biyani et al. (2016), Potthast et al. (2018), Yang et al. (2019), Paixão et al. (2020) |
| Syntactic | PoS: Feng et al. (2012), Potthast et al. (2018), Paixão et al. (2020), Zhou et al. (2020b);<br>PCFG: Feng et al. (2012), Pérez-Rosas et al. (2018), Zhou et al. (2020b);<br>Others: Potthast et al. (2018) |
| Discourse | RST: Rubin and Lukoianova (2015);<br>Others: Karimi and Tang (2019), Zhou et al. (2020b) |
| Latent | CNN: Wang (2017b), Ajao et al. (2018), Yang et al. (2018);<br>RNN: Rashkin et al. (2017), Ruchansky et al. (2017), Ajao et al. (2018), Karimi and Tang (2019), Zhang et al. (2019), Hosseini et al. (2022);<br>Transformers: Vijjali et al. (2020), Kula et al. (2021), Raza and Ding (2022)<br>Topics: Bhattacharjee et al. (2018), Ajao et al. (2019), Benamira et al. (2019), Li et al. (2019) |

TABLE 2.1: Some of the main text representations for misinformation detection

▸ SIMILAR TO ZANNETTOU ET AL. (2019), Zubiaga et al. (2018) provide a comprehensive overview of research in this field, specifically focusing on rumours on OSNs. They categorise rumour classification architectures into four main types: *rumour detection*, *rumour tracking*, *stance classification*, and *veracity classification*. Additionally, they discuss examples of scientific approaches taken and datasets used by researchers to tackle each task—along with the state-of-the-art method for each task.

This research is primarily concerned with misinformation detection using machine learning. The availability of data is a prerequisite to achieving this goal. Furthermore, there are different ML approaches that can and have been used to solve this problem. In this section, existing datasets and ML approaches for misinformation detection are reviewed.

This thesis extends Zubiaga et al. (2018) by further categorising the ML approaches cited in it—and incorporating those cited in other papers supervised, semi-supervised and unsupervised, as is laid out in Table 2.2. It also expands on the applicable datasets for the respective tasks cited in Zubiaga et al. (2018). Their work focuses on rumours, while this research targets the broader ecosystem of misinformation. Note that the information in Table 2.2 does not constitute an exhaustive list of published research papers or datasets in each category.

| ML APPROACH | MISINFORMATION DETECTION | MISINFORMATION TRACKING | STANCE CLASSIFICATION | VERACITY CLASSIFICATION |
|---|---|---|---|---|
| Supervised | Wu et al. (2015), Zubiaga et al. (2016), Ahmed et al. (2017), Ruchansky et al. (2017), Wang et al. (2018), Wu and Liu (2018), Zhang et al. (2020) | Castillo et al. (2011), Ruchansky et al. (2017), Wang et al. (2017) | Kochkina et al. (2017), Shang et al. (2018) | Castillo et al. (2011), Kwon et al. (2017) |
| Semi-supervised | Bhattacharjee et al. (2018), Guacho et al. (2018), Shu et al. (2019) | — | — | — |
| Unsupervised | Chen et al. (2016), Zhang et al. (2016), Zhang et al. (2017), Chen et al. (2018), Hosseinimotlagh and Papalexakis (2018) | — | — | — |
| DATASET | Mitra and Gilbert (2015), Zubiaga et al. (2016), Zubiaga et al. (2016b), Zubiaga et al. (2016c), Kwon et al. (2017), Kochkina et al. (2018), Shu et al. (2018), Rubin (2019) | Kochkina et al. (2018) | Zubiaga et al. (2016c), Mohammad et al. (2016), Mohammad et al. (2017), Kochkina et al. (2018), Gorrell et al. (2019) | Zubiaga et al. (2016c), Kwon et al. (2017), Kochkina et al. (2018), Gorrell et al. (2019), Rubin (2019), Arslan et al. (2020) |

Table 2.2: A breakdown of existing ML architectures for misinformation classification.

▸ The literature on misinformation detection has been mostly focused on supervised learning. Castillo et al. (2011) were among the earliest to evaluate the veracity of OSN content using supervised learning. Their objective was to assess how believable tweets about global news events were over two months. They generated a dataset of 747 tweets, manually labelled ('true' or 'false') by expert judges. Extracted features were topic-based (*e.g.* textual length and sentiment of tweet), network-based (*e.g.* the number of users' followers), propagation-based (*e.g.* total number of tweets) and top-element (*e.g.* fraction of tweets containing the most popular hashtag). They tried four different supervised ML methods including SVMs and Bayes networks, but Decision Trees yielded the highest accuracy.

Ruchansky et al. (2017) created a deep learning model to detect fake news, using Twitter and Weibo data. It consists of three modules: *Capture*, *Score* and *Integrate*. The Capture module is built using a RNN which represents the temporal dynamics of a user's activities, and a `doc2vec` representation[83] of text posted therein. In the Score module, a neural network assigns a score to a user, based on their tendency of being the source of a fake news article. The third module combines information from the first two to classify the article. Supervised ML has also been used to detect rumours by analysing how they propagate. Wu et al. (2015) achieved this using an SVM classifier, while Wu and Liu (2018) used RNNs.

[83] Le and Mikolov (2014), "Distributed Representations of Sentences and Documents"

▸ GIVEN THAT IN real-world scenarios, labelled data is—at least immediately—lacking, some have tried to eliminate this restraint. Shu et al. (2019) proposed a novel semi-supervised approach, which models the interrelationship between the contents, publishers, and users (consumers) of news items (of which some are labelled). It predicts the unlabelled news items, using features extracted from the news articles, social relations between users, users' engagements with the news articles, and publishers' partisan associations. They collated fact-check data from BuzzFeed[84], PolitiFact[85] and Media Bias/Fact-Check[86], into two new datasets[87], which both included information on news contents, publishers and social interactions. They simplified the embeddings of their features using Non-negative Matrix factorisation (NMF) and devised an optimisation algorithm to classify the news articles.

[84] `https://github.com/BuzzFeedNews/2016-10-facebook-fact-check/tree/master/data`

[85] `https://www.politifact.com/factchecks`

[86] `https://mediabiasfactcheck.com`

[87] Shu et al. (2018), "FakeNewsNet: A Data Repository with News Content, Social Context and Dynamic Information for Studying Fake News on Social Media"

Bhattacharjee et al. (2018) used active learning to detect the veracity of news, using partially labelled datasets. Their system comprises two simultaneously running, independent modules. The first module $M_1$ begins with a Logistic Regression classifier and a copy of the labelled dataset. It selects and assigns weights to features by iteratively computing the Joint Mutual Information Maximisation between features and class labels, and gives higher weights to the most relevant ones in a greedy way. $M_1$'s dataset is updated to include the assigned weights, and the classifier is retrained. The second module $M_2$ begins with a copy of the unlabelled and labelled dataset. The latter was used to train an underlying classification model which is based on a CNN. Both modules iteratively classify each unlabelled sample, and they request labels from a human if their predictions do not attain a preset certainty threshold. $M_1$ and $M_2$ update their training sets to include the given labels, and then fine-tune their classification models. Finally, the predictions from both modules are combined into a decision profile and a fusion classifier was used to make a final decision on a sample.

The advantage of unsupervised learning is neither labelled data nor human input is needed. Zhang et al. (2016) considered fake news detection as an outlier detection problem. The rationale behind this is that the behaviours (related to style and timing) of a user when posting rumours and non-rumours will differ. Thus, rumours can be picked up as outliers in the user's feed. They used Principal Component Analysis (PCA) to detect rumours on Weibo. They initially collected verified rumour and non-rumour posts for analysis, to determine relevant features. The 13 selected features were numerical and categorical. When a post is flagged as a rumour, their model collects a set of $N$ recent posts (between 10 and 100) by the poster and extracts the aforesaid features from them. The model then performs PCA which transforms the $N$ posts into a matrix with $N$ rows (posts, the first of which denotes the original flagged post) and eight columns containing quantitative values. Eight was analytically chosen as the optimal number of primary components as it is the smallest number which captured at least 85% of the total variance in the recent posts, using varying $N$ sample sizes. The original post is considered an outlier (*i.e.*, a rumour) if it does not have at least zero neighbours within a given distance: calculated as the mean distance between pairs of posts divided by the standard deviation.

## 2.4 LIMITATIONS OF EXISTING METHODS

Given the significance of information disorder, a lot of work has been done to address many of its subproblems. However, some limitations remain unsolved. The following are some limitations related to this thesis:

1. One of the open challenges in contemporary fake news research is the lack of cross-domain, cross-topic, and cross-language studies.[88] This thesis partly addresses this limitation through the use of cross-domain datasets, that cover several different news topics, for fake news detection.

2. Although extensively used to engineer features for fake news detection,[89] stylometric features are ineffective for distinguishing between genuine news and fake news autogenerated by language models.[90] This limitation may be overcome by exploring features which transcend stylometry, such as topics, which are used in this work.

3. Large amounts of labelled data are needed to create accurate models, as observed by Wu and Liu (2018) and Wang (2017b). This is a motivation for using unsupervised ML, in which case a labelled dataset is not a prerequisite. The process of manually annotating datasets can be costly and very time-consuming. Furthermore, while some authors have employed Amazon Mechanical Turk workers to annotate their datasets, others[91] preferred to use trained annotators, claiming that they made more informed judgements on the veracity of examples. This ascribes an element of doubt to the reliability of manually labelled datasets.

[88] Zafarani et al. (2019), Zhou and Zafarani (2020)

[89] See §2.3.

[90] Schuster et al. (2020), "The Limitations of Stylometry for Detecting Machine-Generated Fake News"

[91] Castillo et al. (2011), Mitra and Gilbert (2015), Vosoughi et al. (2018), Zhang et al. (2018)

# Part II

# CONTRIBUTIONS

# 3

# WORD EMBEDDINGS FOR MISINFORMATION DETECTION

## 3.1 BACKGROUND

One of the prevailing approaches to solving tasks in NLP is based on the hypothesis that words which appear in close proximity tend to have a similar meaning.[92] This is known as the *distributional hypothesis* and was originally posited in 1954.[93] The distributional hypothesis has since led to the development of various methods to encode text in numeric form. Building on it, distributed representations for computing elements were introduced by Hinton et al. (1986) about three decades later. They were among the first to create numerical representations of words.

[92] Levy and Goldberg (2014), "Neural Word Embedding As Implicit Matrix Factorization"

[93] Harris (1954), "Distributional Structure"

More recent approaches in NLP problem-solving are based on neural network word embedding.[94] These models are constructed using neural nets that represent the similarity between words using dense continuous, real vectors of numbers. Today, the representations of words as vectors are generally referred to as *word embeddings*.[95] Semantically similar words will have numerically similar vectors—known as learned distributed feature vectors—which ideally have a much smaller dimension than the vocabulary. Embeddings can also be generated for whole sentences by aggregating word embeddings or using neural nets trained specifically for this task. When the dimensions of the learned vectors of words are reduced to two or three and visualised on a Cartesian plane, relationships between them become apparent. In this chapter, experiments were set up using word and sentence embeddings to find semantic differences between reactions to rumours and non-rumours. The experimental procedure, results, and conclusions are explained in the following subsections.

[94] Collobert and Weston (2008), Mikolov et al. (2013)

[95] Levy and Goldberg (2014), "Neural Word Embedding As Implicit Matrix Factorization"

## 3.2 RELATED WORK

Word embedding models have shown a better performance than classical methods, such as BoW, Term Frequency-Inverse Document Frequency and Distributional Embeddings.[96] However, according to Goldberg and Levy (2014), it remains unknown exactly why some models produce good word representations. Word embeddings have been used effectively for fake news detection, as demonstrated by Bhattacharjee et al. (2018), Shang et al. (2018) and Shu et al. (2019). They have also been used for automated fact-checking.[97]

[96] Mikolov et al. (2011), Mikolov et al. (2013)

Mikolov et al. (2013) observed that a limitation of word representations is their inability to capture idioms. For example, the phrase 'Washington Post' refers to a newspaper, and its meaning is not directly deducible by simply combining the individual meanings of the words 'Washington' and 'Post'. They, therefore, suggest using a Skip-gram model[98] to learn vector representations, as such a model is capable of representing phrases as vectors and is highly efficient, in terms of training time and accuracy. Nonetheless, word embeddings are now established and have been successfully applied to improve performance in various NLP tasks.[99] Examples of open-sourced word embedding tools include `word2vec` by Mikolov et al. (2013), Global Vectors for Word Representation (or `GloVe`) by Pennington et al. (2014), and `FastText` by Bojanowski et al. (2016). Word embeddings have effectively been used for fake news detection. Some papers which used these latent text features are listed in Table 2.2, in §2.3.

[97] Konstantinovskiy et al. (2018), "Towards Automated Factchecking: Developing an Annotation Schema and Benchmark for Consistent Automated Claim Detection"

[98] Mikolov et al. (2013b), "Efficient Estimation of Word Representations in Vector Space"

[99] Collobert et al. (2011), "Natural Language Processing (Almost) from Scratch"

▸ FALSE NEWS AND rumour content tend to be semantically distinct from authentic or non-rumour content.[100] The observation that the semantics of the two tend to differ partly motivates this investigation. Additionally, Choi et al. (2020) found that echo chambers tend to increase virality and accelerate the spread of rumours. They define an echo chamber as a collection of users that have shared at least two rumours in common. They analysed more than one hundred rumours from six fact-checking platforms. These rumours were the subject of nearly 300,000 tweets made by over 170,000 users. Therefore, it can be argued that those who retweet rumours are likely to be more driven to amplify a common message, than those who retweet non-rumour tweets. This amplification may also be in the form of replies that express agreement and may, therefore, be semantically similar.

[100] Parikh and Atrey (2018), Potthast et al. (2018)

The experiment presented in this chapter differs from some previous studies[101] —it focuses not on the rumours or non-rumours posted, but on the reactions which they attract. The goal here is to find out whether there is a difference in dispersion between people's reactions to rumours and non-rumour in tweets. Note that in this section, 'reactions' and 'comments' refer to the replies received by tweets.

[101] Wu et al. (2015), Zhang et al. (2016), Zhang et al. (2017)

## 3.3 PROBLEM DEFINITION

In this experiment, the aim is to determine whether or not there is any evidence to differentiate between rumours and non-rumours tweets, based on the reactions they receive. Latent text representations are used as the discriminant between the two groups. This study is carried out using statistical hypothesis testing. The hypotheses can be stated as follows:

**Hypothesis $H_0$** (Null)**:** *the semantic similarities between rumour and non-rumour tweet reactions are equal.*

**Hypothesis $H_1$** (Alternative)**:** *the semantic similarities of rumour tweet reactions are greater than those of non-rumour tweet reactions.*

For both Hypotheses $H_0$ and $H_1$, the semantic similarities are measured using `InferSent`[102] sentence embeddings. It is expected that there will be a greater similarity amongst rumour tweet reactions, compared with non-rumours ones. This is in line with the aforementioned observations. The method through which Hypothesis $H_1$ will be tested against Hypothesis $H_0$ is explained in the next section.

[102] Conneau et al. (2017), "Supervised Learning of Universal Sentence Representations from Natural Language Inference Data"

## 3.4 METHODOLOGY AND MATERIALS

### 3.4.1 *Experimental procedure*

Let $P_{ALL} \in \{P^R, P^F\}$ represent a dataset containing all $M$ rumour $N$ and non-rumour (factual) posts, $P^R = \{p_1^R, \ldots, p_M^R\}$ and $P^F = \{p_1^F, \ldots, p_N^F\}$, respectively. Each rumour post $p_i^R$ has received comments $c_i^R = \left[c_{i,1}^R, c_{i,2}^R, \ldots, c_{i,m_i}^R\right]$, where $m_i$ is the total number of comments that follow, and $i = \{1 \ldots M\}$. Likewise, each factual tweet $p_i^F$ has received $c_i^F = \left[c_{i,1}^F, c_{i,2}^F, \ldots, c_{i,n_i}^F\right]$ comments (a total of $n_i$), with $i = \{1 \ldots N\}$. Therefore, $C^R = \bigcup_{i=1}^{M} c_i^R$ and $C^F = \bigcup_{i=1}^{N} c_i^F$ are all the reactions to rumours and non-rumours, respectively. Algorithm 1 summarises the computations for this experiment.

Before the experiment, the data was cleaned as follows: (i.) all datasets were cleaned to remove usernames and hashtags; (ii.) comments less than three words long were removed.[103] Next, a pre-trained `InferSent` word embedding model was used to generate 4096-length vectors for each rumour and non-rumour reaction. This gives us matrices for rumour and non-rumour embeddings, $E^R = (e_{i,j}) \in \mathbb{R}^{M\times 4096}$ and $E^F = (e_{i,j}) \in \mathbb{R}^{N\times 4096}$, respectively.

[103] This helps to make the word embedding more accurate. A post may have only a single comment, but all comments must be more than three words long, or else that post is excluded.

▸ THE AVERAGE PAIRWISE cosine similarities, $cs_{avg}$, between the embeddings for rumour and non-rumour reactions are separately calculated. Self-comparisons between items (having a similarity of 1) are excluded. Therefore, an $l \times 4096$ matrix of embeddings is inputted and the output is a $l$-length vector in return,[104] after finding the mean. In this vector, each item is the mean of cosine distances between the embedding of a comment and all other comments.

[104] The row-wise mean is calculated first in Line 6 of Algorithm 1 to compare each comment with every other comment except itself.

Lastly, the average of each vector is calculated, as $A^R$ and $A^F$, and the difference between the two is found as $\Delta A = A^R - A^F$. The higher $\Delta A$ is, the more similar rumours are as compared with non-rumours, and vice versa.

`InferSent` is trained on natural inference data and it generates semantic representations for sentences in English.[105] Embeddings of phrases and sentences

[105] Conneau et al. (2017), "Supervised Learning of Universal Sentence Representations from Natural Language Inference Data"

**Algorithm 1** Comparison of rumour and non-rumour comments using `InferSent` embeddings

**Input:** Comments $C^R$ and $C^F$
**Output:** $\Delta A$

**function** InferSentEmbed(*text*)
    **return** `InferSent` embedding vector for *text*, **e** ▷ $|\mathbf{e}| = 4096$
**end function**
**function** PairwiseCosSim($X = (e_{i,j}) \in \mathbb{R}^{l \times 4096}$)
    Pairwise cosine similarity matrix for $X$ is $S = (a_{i,j}) \in \mathbb{R}^{l \times l}$
    **return** $mean(S)$ ▷ do row-wise mean first
**end function**
**for all** $c_i \in C^F$ **do**
    $E_i^F$ = InferSentEmbed($c_i$)
**end for**
**for all** $c_i \in C^R$ **do**
    $E_i^R$ = InferSentEmbed($c_i$)
**end for**
$E^F = \left\{E_1^F, \ldots, E_M^F\right\} = (e_{i,j}) \in \mathbb{R}^{M \times 4096}$
$E^R = \left\{E_1^R, \ldots, E_N^R\right\} = (e_{i,j}) \in \mathbb{R}^{N \times 4096}$
$A^F$ = PairwiseCosSim($E^F$)
$A^R$ = PairwiseCosSim($E^R$)
$\Delta A = A^R - A^F$
**return** $\Delta A$

are typically obtained by averaging their constituting word embedding vectors. However, `InferSent` is advantageous because it takes the order of words into account to produce embeddings for whole sentences, as it is built on a RNN.[106]

[106] Conneau et al. (2017), Konstantinovskiy et al. (2018)

### 3.4.2 *Datasets*

The PHEME dataset created by Zubiaga et al. (2016b) was used to evaluate Hypotheses $H_0$ and $H_1$. It contains nearly 6,000 tweets concerning five fatal incidents that occurred in North America and Europe between 2014 and 2015. A breakdown of this dataset is shown in Table 3.1.

| EVENT | RUMOURS | NON-RUMOURS | RUMOUR COMMENTS | NON-RUMOUR COMMENTS |
|---|---|---|---|---|
| Charlie Hebdo Shooting (Jan. 2015) | 458 (22%) | 1620 (78%) | 422 (22.0%) | 1493 (88%) |
| Ferguson Unrest (Aug. 2014) | 284 (24.8%) | 231 (75.2%) | 257 (25.8%) | 740 (74.2%) |
| Germanwings Crash (Mar. 2015) | 238 (50.7%) | 231 (49.3%) | 160 (49.0%) | 166 (51.0%) |
| Ottawa Shooting (Oct. 2014) | 470 (52.8%) | 420 (47.2%) | 426 (54.2%) | 360 (45.8%) |
| Sydney Siege (Dec. 2014) | 522 (42.8%) | 699 (57.2%) | 486 (42.6%) | 656 (57.4%) |

TABLE 3.1: Breakdown of PHEME dataset

### 3.4.3 *Results and discussion*

The tendency for rumour and non-rumour content to differ semantically was introduced in §3.2. This experiment hypothesises that rumour reactions will generally be similar to each other—rather than to non-rumour reactions—and vice versa. Similarity, here, is evaluated by computing and comparing the sentence embeddings of the two groups of tweets. It is expected, therefore, that the mean of the pairwise distances between rumours will generally be greater than that between non-rumour comments.

To verify this scientifically, a statistical test was carried out on the experimental results. The differences in the similarities of rumour and non-rumour reactions were evaluated using the Wilcoxon–Mann–Whitney test, at 5% significance level. This test was chosen because the resulting data, for all datasets, did not pass the test for normality, and therefore, could not be assumed to be normally distributed. A Shapiro-Wilk test showed that the distribution of rumour and non-rumour similarity values departed significantly from a normal distribution (R — rumour, NR — non-rumour):

- Charlie Hebdo: R ($W = 0.971, p < 0.01$), NR ($W = 0.915, p < 0.01$)
- Ferguson: R ($W = 0.890, p < 0.01$), NR ($W = 0.915, p < 0.01$)
- Germanwings: R ($W = 0.929, p < 0.01$), NR ($W = 0.970, p < 0.01$)
- Ottawa: R ($W = 0.914, p < 0.01$), NR ($W = 0.901, p < 0.01$)
- Sydney: R ($W = 0.918, p < 0.01$), NR ($W = 0.883, p < 0.01$)

Therefore, the differences between the median similarities, rather than the mean, were conclusively analysed, by applying the Wilcoxon–Mann–Whitney test.

Table 3.2 summarises the results of this experiment, while detailed plots are presented in Figure 3.1. Recall from Algorithm 1, that $A^R$ and $A^F$ are the mean similarities between rumour and non-rumour (factual) tweet reactions, respectively. $Med^R$ and $Med^F$ ($\Delta Med = Med^R - Med^F$) are the median similarities between rumours and non-rumours, respectively.

| EVENT | $A^R$ | $A^F$ | $\Delta A$ | $Med^R$ | $Med^F$ | $\Delta Med$ | $p$-$value$ |
|---|---|---|---|---|---|---|---|
| Charlie Hebdo | 0.603 | 0.599 | 0.004 | 0.603 | 0.608 | -0.005 | 0.487 |
| Ferguson | 0.632 | 0.617 | 0.015 | 0.629 | 0.618 | 0.012 | $1.686 \times 10^{-4}$ |
| German-wings | 0.603 | 0.593 | 0.001 | 0.608 | 0.599 | 0.009 | 0.114 |
| Ottawa | 0.615 | 0.613 | 0.002 | 0.614 | 0.604 | 0.009 | 0.136 |
| Sydney | 0.606 | 0.614 | -0.008 | 0.607 | 0.616 | -0.009 | 0.998 |

TABLE 3.2: Summary of experimental and statistical results for comparisons between sentence embeddings.

The results show that there are no significant, consistent differences between rumours and factual comments when comparing the two using sentence embeddings. Most values of $\Delta A$ are positive as expected, except for the Sydney Siege dataset. The $\Delta Med$ values are also positive, except for the Charlie Hebdo and Sydney Siege datasets. The two measures may suggest that the rumour reactions are generally semantically more similar to each other than non-rumours, but $\Delta A$ and $\Delta Med$ indicate that such a conclusion cannot be made. Except for the Ferguson dataset, the null hypothesis (Hypothesis $H_0$)—that the average semantic similarities between rumour and non-rumour tweet reactions are equal—is not rejected at the 5% level, based on the Wilcoxon–Mann–Whitney test. Therefore, in summary, the Hypothesis $H_0$ tested in this experiment is not rejected.

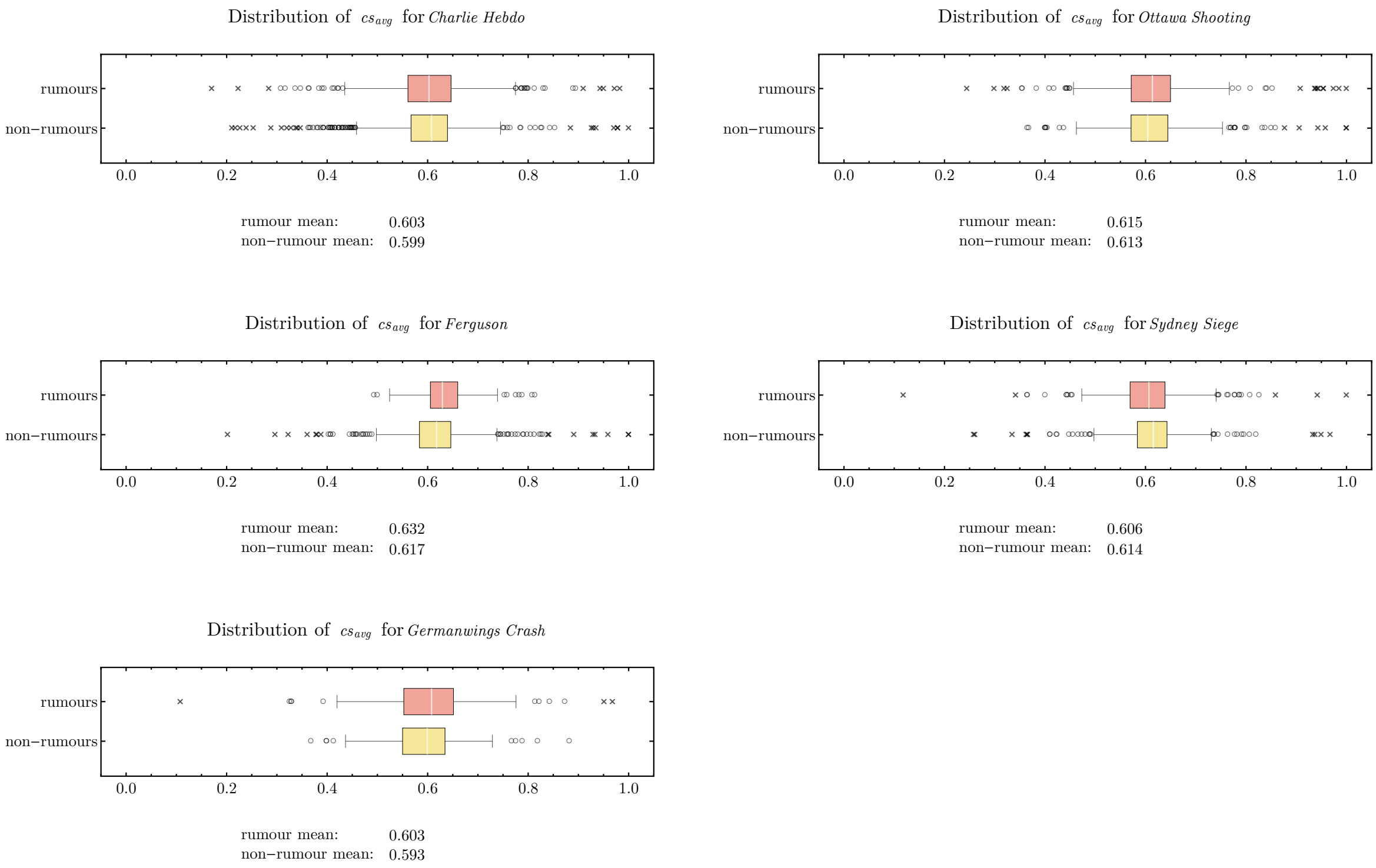

FIGURE 3.1: Box plots of distributions of $cs_{avg}$ for rumour and non-rumour reactions.

There are some possible factors which may have affected the outcome of this experiment. First, sentence embeddings work better with well-written sentences. However, short texts of just 140 characters—not necessarily forming complete sentences—were used here. Second, the datasets used are concerned with separate events in different countries, which may have been discussed in different ways. For example, the Charlie Hebdo event occurred in France, and some of the tweets in that dataset are in French. Similarly, the Germanwings Crash data contains some tweets in German. However, an `InferSent` model for English texts was used to get the sentence embeddings, as most of the tweets are in English. Lastly, the small size of the dataset, coupled with the imbalance of the number of examples in each class, possibly influenced the outcome of this experiment.

▸ FURTHER EXPERIMENTS WERE conducted with minor changes made to the methodology. In one of the follow-up studies, the aim was to determine whether rumour posts are more similar to the comments they attract, compared with non-rumours. In other words, the goal is to compare the semantic differences between rumours and their reactions, and non-rumours and their comments. The following steps were carried out, for each event:

- Embed each post (rumour or non-rumour) into a 300-length vector.
- Embed its corresponding comments also into 300-length vectors.
- Find the averages of pairwise cosine similarities and Euclidean distances between each post and its set of comments. Here, $cs_{avg}$ represents the similarity between a post and the comments it attracted.

The results of this experiment (see Figure 3.2, Figure 3.3 and Table 3.3) also did not yield conclusive results. They show that for some events, the rumour posts are more similar to their comments, while the opposite is the case for others. Even when the embeddings of the comments are averaged before being compared with the embeddings of their original post, significant differences were not found between rumour and non-rumour tweets.

| EVENT | EUCLIDEAN DISTANCE |
|---|---|
| Charlie Hebdo Shooting | -1.365 |
| Ferguson Unrest | 0.924 |
| Germanwings Crash | 1.156 |
| Ottawa Shooting | 0.076 |
| Sydney Siege | -1.608 |

TABLE 3.3: Difference between the averages of the Euclidean distances of rumour and non-rumour comments.

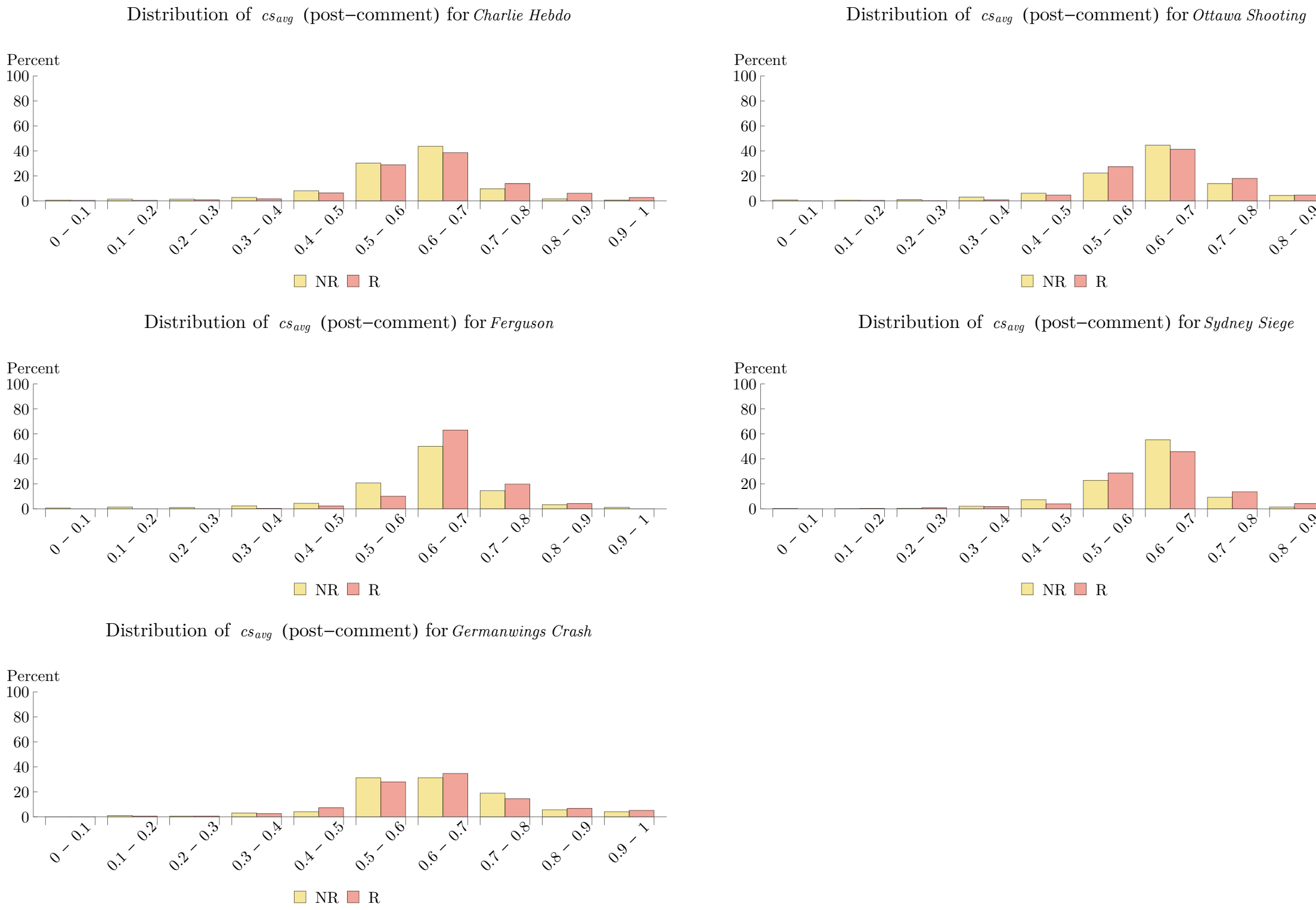

FIGURE 3.2: Distributions of average pairwise cosine similarities between posts and their comments. $NR$ = non-rumours, $R$ = rumours.

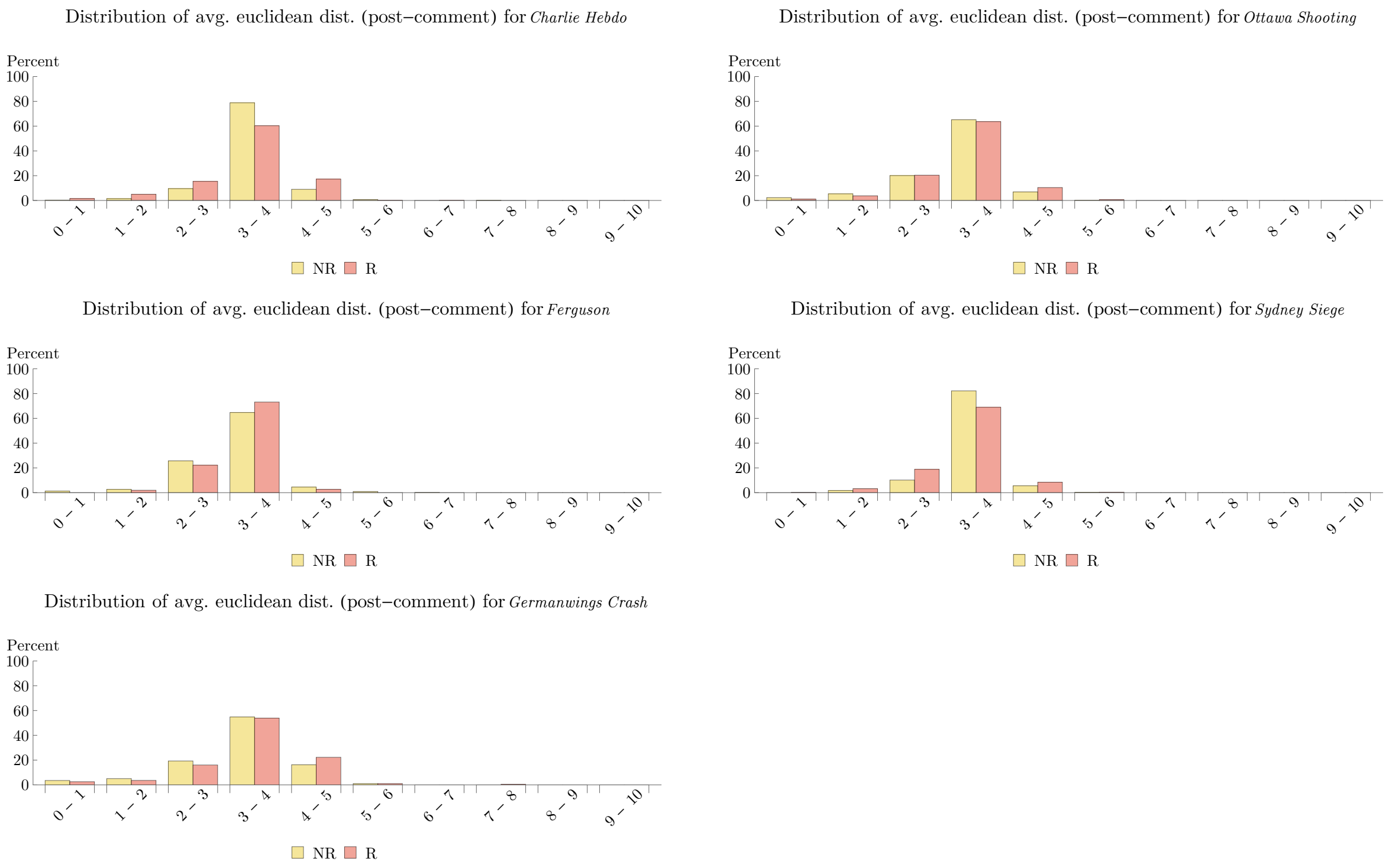

FIGURE 3.3: Distributions of average pairwise Euclidean distances between posts and their comments. $NR$ = non-rumours, $R$ = rumours.

## 3.5 DISPARITIES IN SENTIMENT

As mentioned earlier (see §1.2), authors of false news sometimes seek to arouse emotional responses from readers, as has been observed through studies of their writing style. This observation served as the basis for an experiment which aimed to distinguish between rumours and non-rumours by analysing the sentiment expressed in both sets of tweets.

In this experiment, the Stanford NLP tool[107] was used to analyse the sentiment scores of posts and comments in the PHEME dataset. For a given text, the tool computes one of the following sentiment scores: 0 (very negative), 1 (negative), 2 (neutral), 3 (positive), or 4 (very positive). In the first variant of this experiment, the sentiment scores of the posts and comments of rumour and non-rumour tweets were computed and compared.

[107] Socher et al. (2013), "Recursive Deep Models for Semantic Compositionality Over a Sentiment Treebank"

The results (plotted in Figure 3.4 and Figure 3.5) do not show significant variations between the sentiment scores of rumours and non-rumours, for posts or comments. Further analyses were carried out but the results did not significantly distinguish between rumour and factual comments or posts based on inferred sentiment.

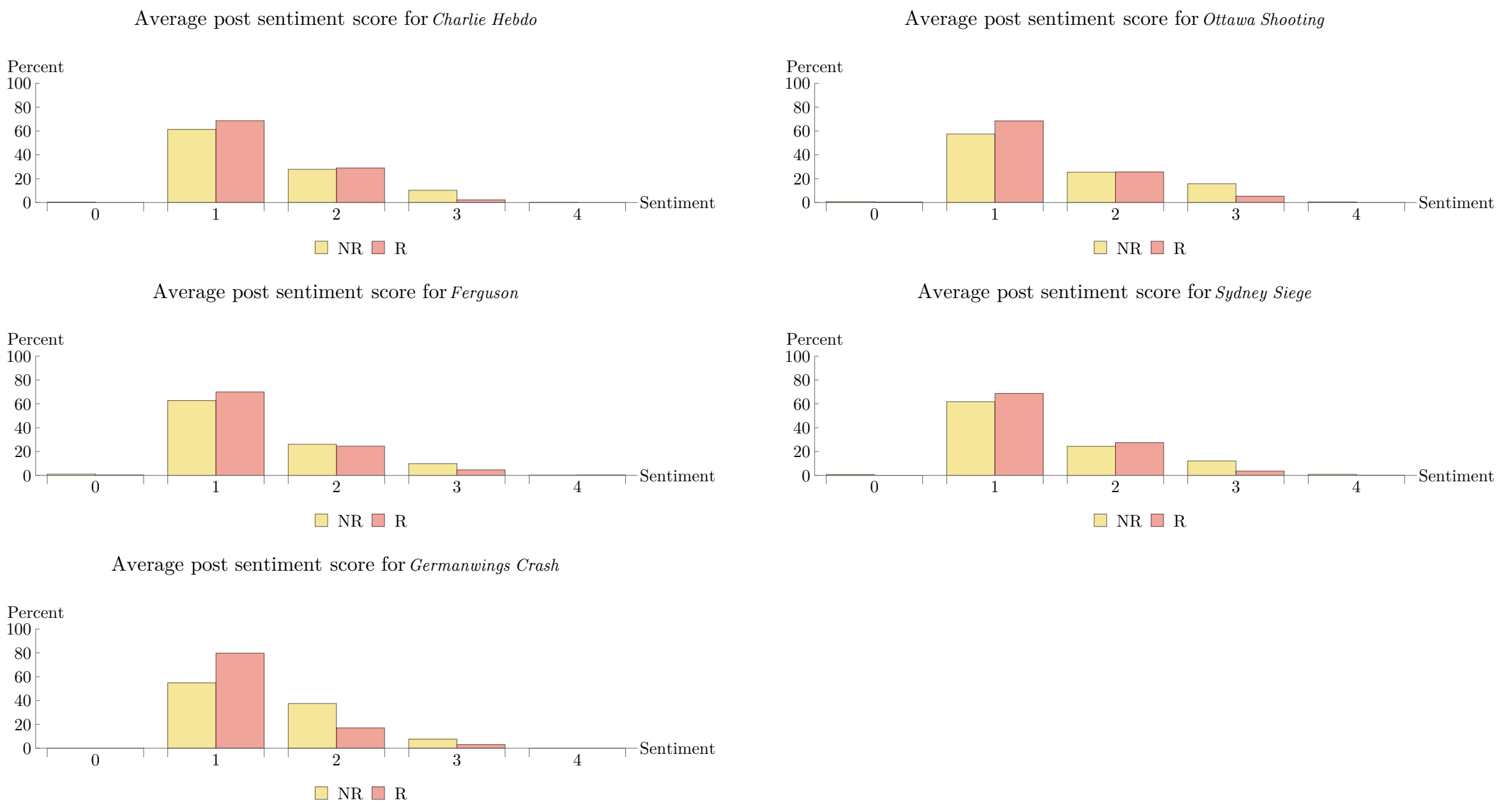

FIGURE 3.4: Average sentiment scores of posts. $NR$ = non-rumours, $R$ = rumours.

▶ ONE FINAL EXPERIMENT was performed regarding text embeddings and sentiment. In it, $K$-Means clustering (with $K = 2$) was used to analyse the sentiment scores for rumour and non-rumour comments. This was repeated with

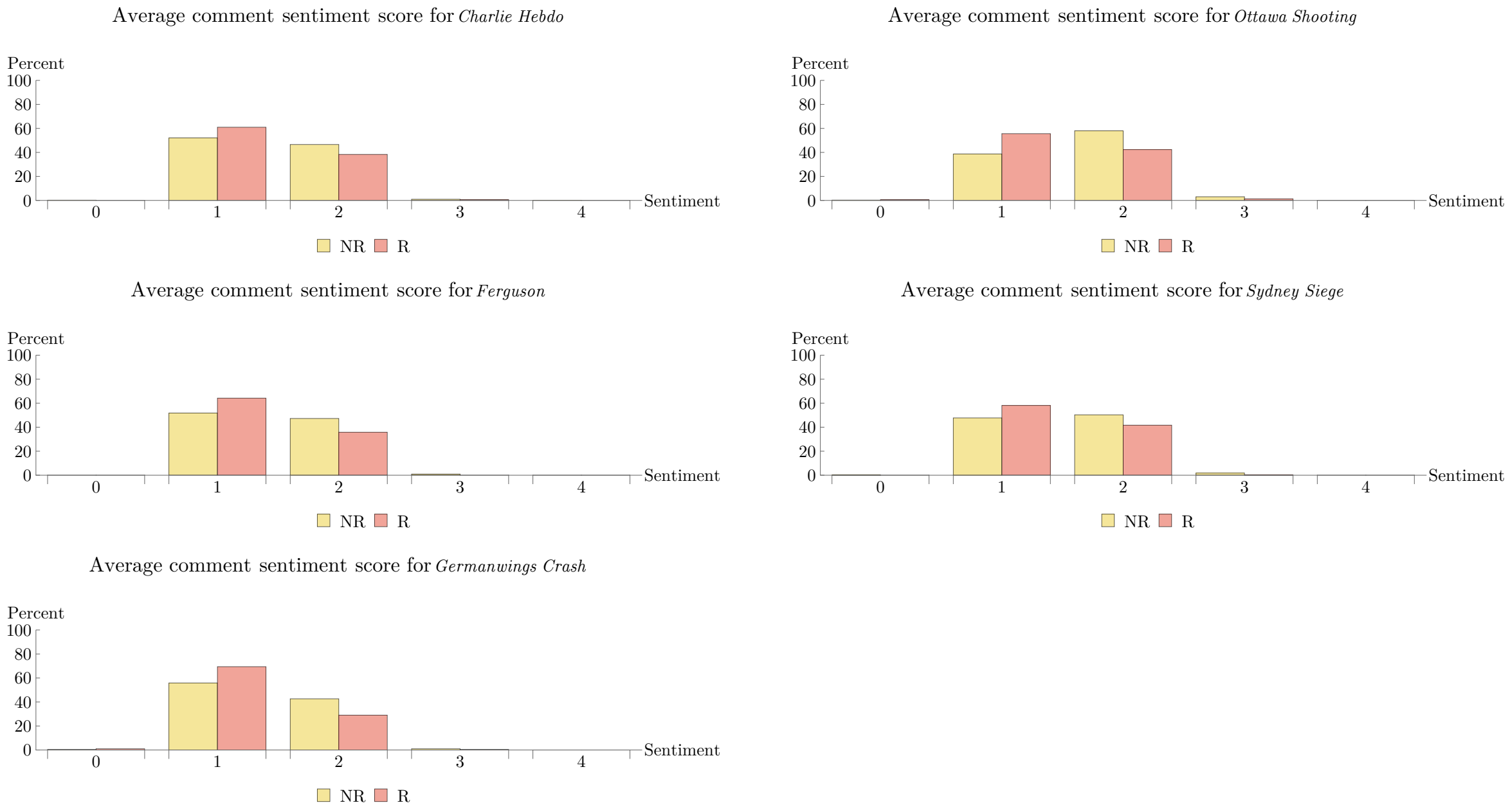

Figure 3.5: Average sentiment scores of comments.
$NR$ = non-rumours, $R$ = rumours.

`InferSent`, as well as `word2vec` (300-length vectors) word embeddings, instead of sentiment scores. The results from clustering showed that there is no clear distinction between rumour and non-rumour tweets.

## 3.6 Conclusion

The series of experiments discussed in this section do not conclude that word or sentence embeddings, or sentiment, can reliably distinguish rumour tweets from factual ones (at least, in the datasets used here), or the reactions that either receive. Nonetheless, these findings are limited, and probably only apply, to the set-up of the experiments presented here. Others have successfully used these text representations in different ways to differentiate between the two types of tweets. To further test the methodology followed here will require substantially larger datasets. Furthermore, as opposed to using simplistic sentiment measures (positive, neutral, and negative), a more granular and precise measure of emotions in tweets can be explored. For example, Vosoughi et al. (2018) examined a range of positive emotions (such as joy and trust) as well as negative ones (such as fear and anger) in both false and true news. They found that comments to false rumours had greater surprise and disgust expressed in them, while reactions to true ones expressed more sadness and anticipation. Similarly, Kolev et al. (2022) carried out fake news detection by using the predicted six emotions (anger, disgust, fear, joy, sadness, and surprise) in the titles of news articles as features.

# 4

# THEMATIC COHERENCE IN FAKE NEWS

## 4.1 BACKGROUND

*'The construction of life is at present in the power far more of facts than of convictions, and of such facts as have scarcely ever become the basis of convictions.'*

— Walter Benjamin, "One-Way Street"

This chapter deals with the exploration of thematic coherence of fake news. News readers are often enticed by the headlines of articles, or their opening sentence(s). False news is written for many different reasons, including propaganda, provocation and profit,[108] and therefore, often in catchy or emotive language. Given the deluge of information which competes daily for people's attention, most people would now skim through news pieces that they would otherwise carefully read—perhaps to save their time—or in an attempt to spend it on stories of greater interest to them. However, this inattention can be exploited by propagators of misinformation, as they can make the headlines or openings of false news captivating. An indication of a misleading article could, therefore, be that its headline or starting paragraph thematically deviates from the rest of the article.

[108] Shu et al. (2017), "Fake News Detection on Social Media"

In this chapter, the focal point is fake news that appears in the form of long online articles and explores the extent of internal consistency within fake news vis-à-vis legitimate news. In particular, these experiments aim to determine whether *thematic deviations—i.e.*, a measure of how dissimilar topics discussed in different parts of an article are—between the opening and remainder sections of texts can be used to distinguish between fake and real news across different news domains. Put simply, this is a measure of the distance between the distributions of topics extracted from two sections of an article, the opening and the remainder. The dissemination of fake news is increasing, and because it appears in various forms and self-reinforces,[109] it is difficult to erode. Hence, there is an urgent need for increased research in understanding and curbing it.

[109] Wardle (2017), Waldman (2018), Zhou and Zafarani (2018)

▸ ONE STUDY BY Gabielkov et al. (2016) found that, as of 2016, 59% of links shared on OSNs have never been clicked before. This indicates that people share information without actually reading it. A more recent study by Anspach et al. (2019) suggests that some readers may skim through an article instead of reading the whole content because they overestimate their political knowledge, while others may hastily share news without reading it fully, for emotional affirmation. This presents bad actors with the opportunity to deftly intersperse news content with falsity. Moreover, the production of fake news typically involves the collation of disjointed content and lacks a thorough editorial process.[110]

The limitation of existing misinformation detection methods not adequately capturing the subtle differences between false and legitimate news motivates the experiments presented in this section.

[110] Karimi and Tang (2019), "Learning Hierarchical Discourse-level Structure for Fake News Detection"

▸ TOPICS DISCUSSED IN news pieces can be studied to ascertain whether an article *thematically deviates* between its opening and the rest of the story, or if it remains coherent throughout. In other words, does an article open with one topic and finish with a different, unrelated topic? Thematic analysis is useful here for two reasons. First, previous studies show that the coherence between units of discourse (such as sentences) in a document is useful for determining its veracity.[111] Second, analysis of thematic deviation can identify general characteristics of fake news that persist across multiple news domains.

Topics have been employed as features for misinformation detection using ML.[112] However, they have not been applied to study the unique characteristics of fake news. Research efforts in detecting fake news through thematic deviation have thus far focused on spotting incongruences between pairs of headlines and body texts.[113] Yet, thematic deviation can also exist within the body text of a news item. The focus is to examine these deviations to distinguish fake from real news.

To the best of the author's knowledge, this is the first work that explores thematic deviations in the body text of news articles to distinguish between fake and legitimate news.

[111] Rubin and Lukoianova (2015), Karimi and Tang (2019)

[112] Bhattacharjee et al. (2018), Benamira et al. (2019), Li et al. (2019)

[113] Chen et al. (2015), Ferreira and Vlachos (2016), Sisodia (2019), Yoon et al. (2019)

## 4.2 RELATED WORK

The coherence of a story may be indicative of its veracity. For example, Rubin and Lukoianova (2015) demonstrated this by applying RST[114] to study the discourse of deceptive stories posted online. They found that a major distinguishing characteristic of deceptive stories is that they are disjunctive. Furthermore, while truthful stories provide evidence and restate information, deceptive ones do not. This suggests that false stories may tend to thematically deviate more due to disjunction, while truthful stories are likely to be more coherent due to restatement. Similarly, Karimi and Tang (2019) investigated the coherence of fake and real news by learning hierarchical structures based on sentence-level dependency parsing. Their findings also suggest that fake news documents are less coherent.

[114] Mann and Thompson (1988), "Rhetorical Structure Theory: Toward a functional theory of text organization"

▸ TOPIC MODELS ARE unsupervised algorithms that aid the identification of themes discussed in large corpora. With them, these texts can be understood, organized, summarised and searched for automatically.[115] One example of topic models is LDA, which is a generative probabilistic model that aids the discovery of latent themes or *topics* in a corpus.[116] Vosoughi et al. (2018) used LDA to show that false rumour tweets tend to be more novel than true ones. Novelty was evaluated using three measures: Information Uniqueness, Bhattacharyya Distance, and Kullback-Leibler Divergence. Likewise, Ito et al. (2015) used LDA to assess the credibility of Twitter users by analyzing the topical divergence of their tweets from those of other users. They also assessed the veracity of users' tweets by comparing the topic distributions of new tweets against historically discussed topics. Divergence was computed using the Jensen-Shannon Divergence, Root Mean Squared Error, and Squared Error. This work primarily differs from those two, in that here, full-length articles are analysed instead of tweets.

[115] Blei (2012), "Probabilistic topic models"

[116] Blei et al. (2003), "Latent Dirichlet Allocation"

## 4.3 PROBLEM DEFINITION

Building on the previous subsections, the aim is to establish whether or not there is evidence to distinguish between fake and authentic news, based on the coherence of topics discussed in them. Similar to Chapter 3, the statistical hypothesis testing approach is found to be appropriate for carrying out this study. The following hypotheses are tested:

**Hypothesis $H_0$** (Null)**:** *False and authentic news articles are similarly coherent thematically.*

**Hypothesis $H_1$** (Alternative)**:** *the thematic coherence of authentic news articles is greater than that of false news articles.*

Specifically, the thematic drift between the opening part and the remaining part of an article is measured, to see how they differ. The primary tool used to measure this is LDA topic modelling. The opening section of an article is defined using a hyperparameter, $l$, which is the number of sentences at the start of it. To test Hypotheses $H_0$ and $H_1$, experiments are carried out in the manner outlined in Algorithm 2. The differences in mean and median coherence values of fake and real articles are evaluated using an Independent Samples T-test, at 5% significance level.

## 4.4 RESEARCH GOAL AND CONTRIBUTIONS

The research presented in this chapter aims to assess the importance of internal consistency within articles as a high-level feature to distinguish between fake and real news stories across different domains. This chapter sets out to explore whether the opening segments of fake news thematically deviate from the rest of it, significantly more than in authentic news. Experiments are conducted using seven datasets which collectively cover a wide variety of news domains, from business to celebrity, to warfare. Deviations are evaluated by calculating the distance between the topic distribution of the opening part of an article, to that of its remainder. The first five sentences of an article are taken as its opening segment.

The following summarise the contributions of this chapter:

- It presents new insights towards understanding the underlying characteristics of fake news, based on thematic deviations between the opening and remainder parts of news body text.
- Experiments are carried out on five cross-domain misinformation datasets; the results demonstrate the effectiveness of thematic deviation for distinguishing fake from real news.

## 4.5 METHODOLOGY AND MATERIALS

### 4.5.1 *Latent Dirichlet Allocation*

Given a text document, an LDA model generates words by selecting a topic from the document-topic distribution, and then selecting a word from the topic-word distribution.[117] A brief description of how LDA works is given here, following the notation used by Maiya and Rolfe (2015) for its clarity.

[117] Blei (2012), "Probabilistic topic models"

Let $D = \{d_1, \ldots, d_N\}$ be a corpus, consisting of $N$ documents which collectively cover $T = \{t_1, \ldots, t_K\}$ latent topics. Each document, $d_i$, is made up of a sequence of words. That is, $d_i = \langle w_{i,1}, w_{i,2}, \ldots, w_{i,W_i} \rangle$, where $i \in \{1 \ldots N\}$ and $W_i$ is the total number of words in—called the *vocabulary* of—$d_i$. Therefore, the vocabulary of $D$ is $\mathscr{W} = \bigcup_{i=1}^{N} d_i$.

In addition to some hyperparameters, probabilistic topic models such as LDA require only two inputs: (i.) corpus $D$; and (ii.) desired number of topics $K$. They output two matrices: (i.) the document-topic distribution matrix, $\theta \in \mathbb{R}^{N \times K}$, which represents the topics drawn from each document;[118] and (ii.) the topic-word distribution matrix, $\phi \in \mathbb{R}^{K \times |\mathscr{W}|}$, which represents the distribution of words within each topic. The model assumes that each row in both matrices is a Dirichlet probability distribution, hence its name. The optimal value of $K$ is typically found by iteration. If $K$ is overly high, the resulting topics may be uninterpretable, and should ideally have been merged; and if it is too low, the topics will be too broad, *i.e.*, covering many differing concepts.[119]

[118] This is the 'Allocation' in LDA.

[119] Syed and Spruit (2018), "Full-Text or abstract? Examining topic coherence scores using latent dirichlet allocation"

A topic found in a document $d_i$ is usually shown as a combination of a word $w_i$ and its probability $p_i$ in the distribution $\phi_i$, as $(p_i * w_i)$. For example, $(fact * 0.01)$ or $(fake * 0.001)$. Each topic distribution contains the entire vocabulary, with varying probabilities assigned to each word. The word with the highest probability in the distribution is usually used to label a topic.[120] Words that have higher probabilities within a topic would tend to co-occur in the corpus as a whole.

[120] Maiya and Rolfe (2015), "Topic similarity networks: Visual analytics for large document sets"

LDA generates document-topic distributions $\theta_d$ and word-topic distributions $\phi_t$. Figure 4.1[121] shows a graphical model of LDA. The box labelled $D$ represents the documents in a corpus. While boxes $\mathscr{W}$ and $K$ represent the repeatedly selected words and topics within a document, respectively. The circles are random variables in the generative process. The Dirichlet parameter $\alpha$ controls the sparsity of topics within documents, while $\beta$ controls the sparsity of words within topics.

[121] Adapted from Blei et al. (2003) (Fig. 1) and Blei (2012) (Fig. 4).

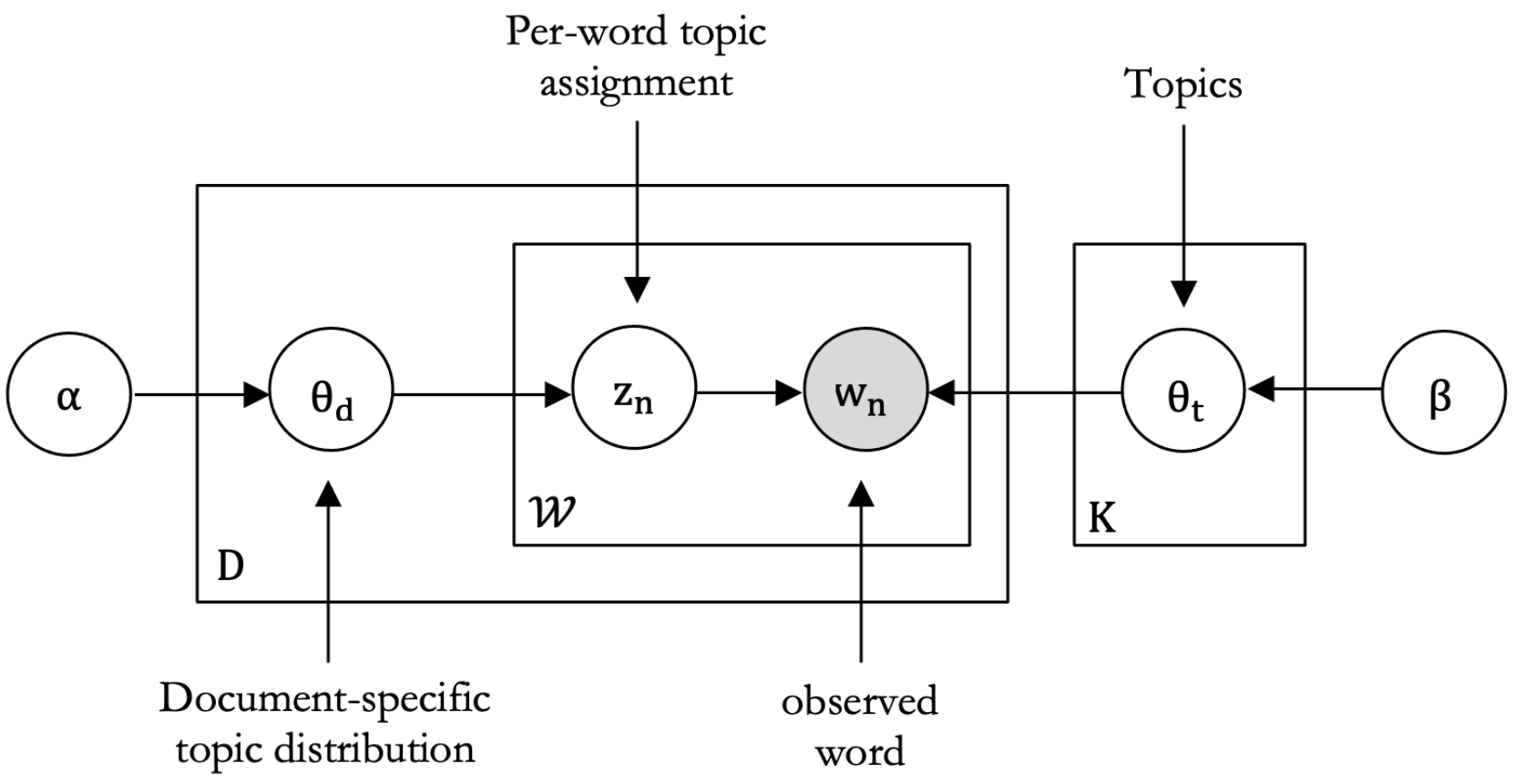

FIGURE 4.1: Graphical representation of LDA

The hidden variables (topics, topic proportions, and assignments) are unshaded, while the observed variable (words in a document) is shaded.

### 4.5.2 *Distance measures*

Distributional similarity/distance measures are commonly used to compare the similarities, differences and overlaps between topics extracted from corpora.[122]

All articles $S_{bg}$ are split into two parts: its first $x$ sentences[123], and the remaining $y$. Next, $N$ topics are obtained from $x$ and $y$ from an LDA model trained on the entire dataset. For $i = (1, \ldots, m)$ topics, let $p_x = (p_{x1}, \ldots, p_{xm})$ and $p_y = (p_{y1}, \ldots, p_{ym})$ be two vectors of topic distributions, which denote the prevalence of a topic $i$ in the opening text $x$ and remainder $y$ of an article, respectively. Finally, the average and median values of each distance are calculated across all fake ($S_f$) and real ($S_r$) articles. These steps were repeated with varying values of $N$ (from 10 to 200 topics) and $x$ (from 1 to 5 sentences).

The following are the data required for this procedure: a corpus $S_{bg} = S_f \bigcup S_r$ of full-length fake ($S_f = \left\{d_1^f, d_2^f, \ldots, d_F^f\right\}$) and real ($S_r = \left\{d_1^r, d_2^r, \ldots, d_R^r\right\}$) documents.

The following measures were considered for calculating the topical divergence between parts $x$ and $y$ of an article:

1. Cosine distance ($D_C$):

$$D_C\left(p_x, p_y\right) = \frac{p_x \cdot p_y}{\left\|p_x\right\| \left\|p_y\right\|} = 1 - \frac{\sum_{i=1}^{m} p_{x_i} p_{y_i}}{\sqrt{\sum_{i=1}^{m} {p_{x_i}}^2} \sqrt{\sum_{i=1}^{m} {p_{y_i}}^2}} \tag{4.1}$$

[122] Omar et al. (2015), Vosoughi et al. (2018)

[123] Only articles with at least $x + 1$ sentences are used.

2. Chebyshev distance ($D_{Ch}$):

$$D_{Ch}\left(p_{x_i}, p_{y_i}\right) = \max_{i=1 \ldots m} \left|p_{x_i} - p_{y_i}\right| \quad (4.2)$$

3. Euclidean distance ($D_E$):

$$D_E\left(p_{x_i}, p_{x_i}\right) = \left\|p_{x_i} - p_{x_i}\right\| = \sqrt{\sum_{i=1}^{m}\left(p_{x_i} - p_{x_i}\right)^2} \quad (4.3)$$

4. Hellinger distance ($D_H$):

$$D_H\left(p_x, p_y\right) = \frac{1}{\sqrt{2}}\sqrt{\sum_{i=1}^{m}\left(\sqrt{p_{x_i}} - \sqrt{p_{y_i}}\right)^2} \quad (4.4)$$

5. Jensen-Shannon divergence ($D_{JS}$):

$$D_{JS}\left(p_x \parallel p_y\right) = \frac{1}{2}\left[D_{KL}\left(p_x \parallel p_z\right) + D_{KL}\left(p_y \parallel p_z\right)\right] \quad (4.5)$$

where $p_z = \frac{1}{2}\left(p_x + p_y\right)$

6. Kullback-Leibler (KL) divergence ($D_{KL}$)[124]:

$$D_{KL}\left(p_x \parallel p_y\right) = \sum_{i=1}^{m} p_{x_i} \log \frac{p_{x_i}}{p_{y_i}} \quad (4.6)$$

7. Squared Euclidean distance ($D_{SE}$):

$$D_E\left(p_{x_i}, p_{x_i}\right) = \left\|p_{x_i} - p_{x_i}\right\|^2 = \sum_{i=1}^{m}\left(p_{x_i} - p_{x_i}\right)^2 \quad (4.7)$$

These measures were all used in the preliminary explorations carried out for this chapter. Eventually, however, only three ($D_{Ch}$, $D_E$, and $D_{SE}$) were proceeded with in the main experiments. Intuitively, the cosine distance indicates the *angular* gap between two vectors (distributions of topics, in this case). Chebyshev distance is the greatest difference found between any two topics in $x$ and $y$. The Euclidean distance measures how *far* the two topic distributions are from one another, while the Squared Euclidean distance is simply the square of that *farness*. The other measures ($D_H$, $D_{JS}$, and $D_{KL}$) were considered as they were originally developed to deal directly with probability distributions.

[124] $D_{KL}$ is not symmetric and therefore not a metric, but it can be transformed into one—to form the Jensen-Shannon divergence, $D_{JS}$ (Equation 4.5).

**Algorithm 2** Evaluation of thematic divergence in news articles

**Input:** (i.) Pairs of first $l = [1, 2, \ldots, 5]$ sentences and remainder $y$ of each fake ($d_i^f = \left\langle d_{i_x}^f, d_{i_y}^f \right\rangle$; $\left| d_{i_x}^f \right| = l$) and real article ($d_i^r = \left\langle d_{i_x}^r, d_{i_y}^r \right\rangle$; $\left| d_{i_x}^r \right| = l$);
(ii.) LDA model $\mathscr{M}_{bg}$ generated using $S_{bg}$;
(iii.) Number of topics $N \in \{10, 20, 30, 4050, 100, 150, 200\}$;
(iv.) Divergence function $\mathscr{D} \in \{D_{Ch}, D_E, D_{SE}\}$

**Output:** $\left\{ D_{avg}^f, D_{avg}^r, D_{med}^f, D_{med}^r \right\}$

**for all** $l = [1, 2, \ldots, 5]$ **do**
  **for all** fake articles $\left\langle d_{i_x}^f, d_{i_y}^f \right\rangle$ **do**
    get $N$ topics in $d_{i_x}^f$ and $d_{i_y}^f$ using $\mathscr{M}_{bg}$
  **end for**
  $T_{i_x}^f = (p_i^x, \ldots, p_N^x)^f$ ▷ Topics in opening of fake article
  $T_{i_y}^f = \left(p_i^y, \ldots, p_N^y\right)^f$ ▷ Topics in remainder of fake article
  $D_i^f = \mathscr{D}\left(T_{i_x}^f, T_{i_y}^f\right)$
  **for all** real articles $\left\langle d_{i_x}^r, d_{i_y}^r \right\rangle$ **do**
    get $N$ topics in $d_{i_x}^r$ and $d_{i_y}^r$ using $\mathscr{M}_{bg}$
  **end for**
  $T_{i_x}^r = (p_i^x, \ldots, p_N^x)^r$ ▷ Topics in remainder of real article
  $T_{i_y}^r = \left(p_i^y, \ldots, p_N^y\right)^r$ ▷ Topics in remainder of real article
  $D_i^r = \mathscr{D}\left(T_{i_x}^r, T_{i_y}^r\right)$
  $D_{avg}^f = mean\left(D_i^f \big|_{i \in \{1,\ldots,F\}}\right)$; $D_{avg}^r = mean\left(D_i^r \big|_{i \in \{1,\ldots,R\}}\right)$
  $D_{med}^f = median\left(D_i^f \big|_{i \in \{1,\ldots,F\}}\right)$; $D_{med}^r = median\left(D_i^r \big|_{i \in \{1,\ldots,R\}}\right)$
**end for**
**return** $\left\{ D_{avg}^f, D_{avg}^r, D_{med}^f, D_{med}^r \right\}$

## 4.6 EXPERIMENT

### 4.6.1 *Preprocessing*

All computational operations in this experiment were performed using Python and freely available packages. Preprocessing for each dataset was done in the following steps:

1. Articles are split into sentences using the `NTLK`[125] package. Each sentence is tokenised, lowercased and normalised (*i.e.*, accentuations are removed) to form a list of words, from which stopwords are removed. The union of the built-in stopwords in the `NLTK` and `spaCy` toolkits, as well as the MySQL Reference Manual,[126] was used to filter irrelevant words. Furthermore, additional words typically found in news text but can be considered to be unimportant were added. Examples of such words include long and short forms of days of the week, and months, and others such as *'says'*, *'said'*, *'Reuters'*, *'Mr'*, and *'Mrs'*.

2. Bigrams were created from two consecutive words which appeared several times in the corpus. A minimum count of such instances was set to five and a threshold score as explained in Mikolov et al. (2013b) of 100 was used. The bigrams are then added to the vocabulary.

3. Next, each document is lemmatized using `spaCy`[127], and only noun, adjective, verb, and adverb lemmas are retained. A dictionary is formed by applying these steps to $S_{bg}$.

4. Each document is converted into a BoW format,[128] which is used to create an LDA model $\mathscr{M}_{bg}$. The models were created with `Gensim`.[129]

5. Fake and real articles are subsequently preprocessed likewise (*i.e.*, from raw text data to BoW format) before topics are extracted from them.

Although there is no consensus on whether the inclusion or omission of stopwords yields better topic models,[130] stopwords can affect the interpretability of topics as they can diminish the appearance of other more important words. In this experiment, the goal is to find differences between topics extracted from legitimate and false news. As false news content often cunningly mimics true news, it is important to remove words which are contextually irrelevant and focus on words which can help us tell the two apart.

125 `https://www.nltk.org`

126 `https://dev.mysql.com/doc/refman/8.0/en/fulltext-stopwords.html`

127 `https://spacy.io/models/en`

128 A list of (`token_id`, `token_count`) tuples.

129 `https://radimrehurek.com/gensim`

130 Shi et al. (2019), "A new evaluation framework for topic modeling algorithms based on synthetic corpora"

### 4.6.2 *Datasets*

Table 4.1 summarizes the datasets (after preprocessing) used in this study and lists the domains (as stated by the dataset provider) covered by each. An article's sentence length (Avg. sent. length) is measured by the number of words that remain after preprocessing. The article's maximum sentence length (Max. sent length) is measured in terms of the number of sentences. The following datasets were used:

1. BuzzFeed-Webis Fake News Corpus 2016 (BuzzFeed-Web)[131]
2. BuzzFeed Political News Data (BuzzFeed-Political)[132]
3. FakeNewsAMT + Celebrity (AMT+C)[133]
4. Falsified and Legitimate Political News Database (POLIT)[134]
5. George McIntire's fake news dataset (GMI)[135]
6. University of Victoria's Information Security and Object Technology (ISOT) Research Lab[136]
7. Syrian Violations Documentation Centre (SVDC)[137]

[131] Potthast et al. (2018), https://zenodo.org/record/1239675

[132] Horne and Adali (2017), https://github.com/BenjaminDHorne/fakenewsdata1

[133] Pérez-Rosas et al. (2018), "Automatic Detection of Fake News"

[134] http://victoriarubin.fims.uwo.ca/news-verification/access-polit-false-n-legit-news-db-2016-2017

[135] https://github.com/GeorgeMcIntire/fake_real_news_dataset (Accessed 5 November 2018)

[136] Ahmed et al. (2017), https://www.uvic.ca/engineering/ece/isot

[137] Salem et al. (2019), https://zenodo.org/record/2532642

| DATASET (DOMAIN) | NO. OF FAKE | NO. OF REAL | AVG. SENT LENGTH (F) | AVG. SENT LENGTH (R) | MAX. SENT LENGTH (F) | MAX. SENT LENGTH (R) |
|---|---|---|---|---|---|---|
| AMT+C (business, education, entertainment, politics, sports, tech) | 324 | 317 | 14.7 | 23.2 | 64 | 1,059 |
| BuzzFeed-Political (politics) | 116 | 127 | 18.9 | 43.9 | 76 | 333 |
| BuzzFeed-Web (politics) | 331 | 1,214 | 21.7 | 26.4 | 117 | 211 |
| GMI (politics) | 2,695 | 2,852 | 33.9 | 42.8 | 1,344 | 406 |
| ISOT (government, politics) | 19,324 | 16,823 | 18.0 | 20.3 | 289 | 324 |
| POLIT (politics) | 122 | 134 | 19.2 | 34.9 | 96 | 210 |
| SVDC (conflict, war) | 312 | 352 | 14.0 | 14.6 | 62 | 49 |

TABLE 4.1: Summary of datasets after pre-processing (F – Fake, R – Real).

## 4.7 RESULTS AND DISCUSSION

As can be seen in the first line of Algorithm 2, experimented with were varying values of hyperparameter for the length of the opening section, $l$, from 1 to 5. Results for $l = 5$ are reported in this section because during initial analyses it yielded the best results (*i.e.*, the greatest disparity between fake and real deviations) for most datasets and measures. This is likely due to the first five sentences containing more information. For example, five successive sentences are likely to entail one another and contribute more towards a topic than a single sentence.

The outcomes of the experimental evaluation using the different divergence measures are shown in Table 4.2.[138] It was observed that fake news is generally likely to show greater thematic deviation (lesser coherence) than real news in all datasets. Nonetheless, the mean and median values for fake news are lower than those of real news for these datasets. Table 4.3, shows the mean and median $D_{Ch}$ deviations of fake and real articles across all values of $N$, while Figure 4.2 shows results for comparing topics in the first five and remaining sentences. Results for values of $N$ not shown are similar (with $D_{Ch}$ gradually decreasing as $N$ increases).

As the results for all three measures are alike, $D_{Ch}$ is focused on for the rest of the analysis. This is because the choice of divergence measure is not critical to the outcome of the experiment. Rather it is only a means for estimating thematic divergence. Table 4.3 shows the mean $D_{Ch}$ deviations of fake and real articles across $N = \{10, 20, 30, 40, 50, 100, 150, 200\}$ topics.

[138] The average of each $N$ group was found before doing the T-test.

AMT+C and BuzzFeed-Web are not statistically significant according to the T-test. However, the results for all other datasets are. Therefore, for all datasets except AMT+C and BuzzFeed-Web, the null hypothesis (Hypothesis $H_0$)—that false and authentic news articles are similarly coherent thematically—is rejected at the 5% level, based on the T-test. In summary, it has been shown statistically that thematic coherence is generally greater in real news articles compared to fake ones.

| DATASET | P-VALUE ($D_{Ch}$) | P-VALUE ($D_E$) | P-VALUE ($D_{SE}$) |
|---|---|---|---|
| AMT+C | 0.144 | 0.126 | 0.116 |
| BuzzFeed-Political | 0.0450 | 0.0147 | 0.0287 |
| BuzzFeed-Web | 0.209 | 0.209 | 0.207 |
| GMI | 0.0480 | 0.00535 | 0.0106 |
| ISOT | 0.00319 | 0.000490 | 0.000727 |
| POLIT | 0.000660 | 0.0000792 | 0.0000664 |
| SVDC | 0.000684 | 0.0000112 | 0.0000789 |

TABLE 4.2: Results of T-test evaluation based on different measures of deviation used.

| DATASET | MEAN ($D_{Ch}$) (F) | MEAN ($D_{Ch}$) (R) | MEDIAN ($D_{Ch}$) (F) | MEDIAN ($D_{Ch}$) (R) |
|---|---|---|---|---|
| AMT+C | 0.2568 | 0.2379 | 0.2438 | 0.2285 |
| BuzzFeed-Political | 0.2373 | 0.2149 | 0.2345 | 0.2068 |
| BuzzFeed-Web | 0.2966 | 0.2812 | 0.2863 | 0.2637 |
| GMI | 0.4580 | 0.4241 | 0.4579 | 0.4222 |
| ISOT | 0.3372 | 0.2971 | 0.3369 | 0.2989 |
| POLIT | 0.2439 | 0.1939 | 0.2416 | 0.1894 |
| SVDC | 0.2975 | 0.2517 | 0.2934 | 0.2435 |

TABLE 4.3: Mean and median $D_{Ch}$ deviations of $N = \{10, 20, 30, 40, 50, 100, 150, 200\}$ topics combined for fake and real news (F – Fake, R – Real).

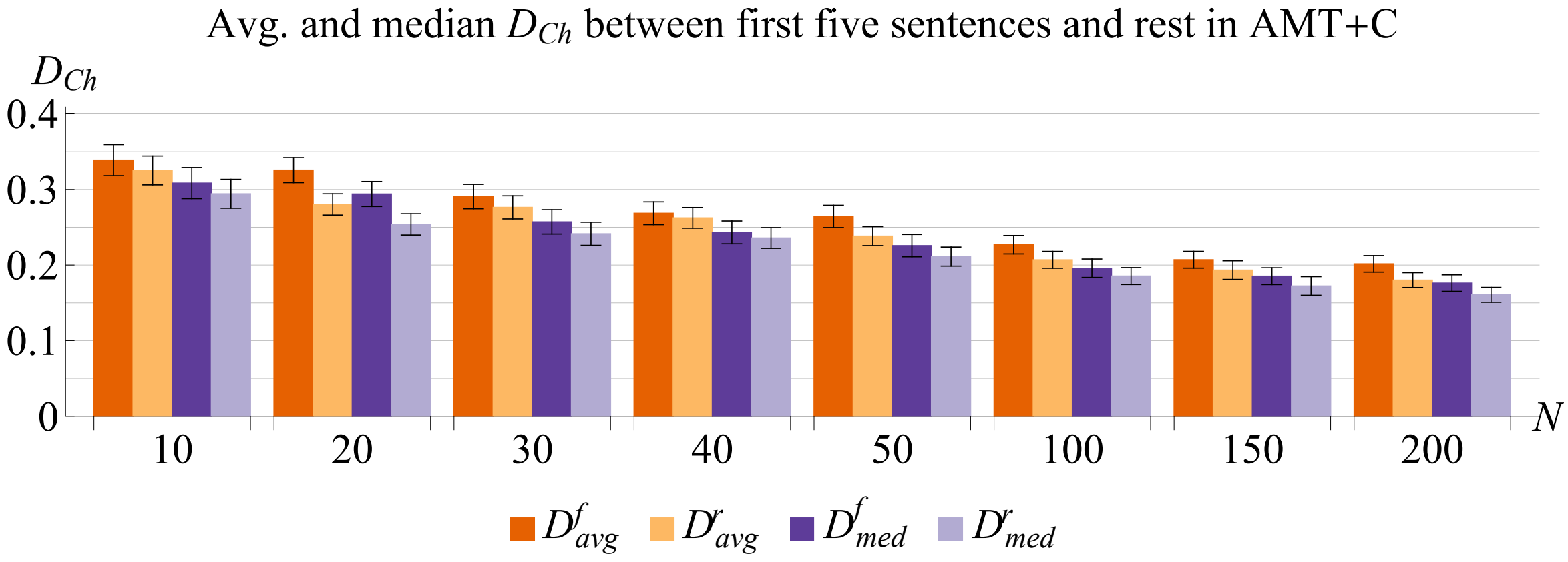

(a) Results for AMT+C

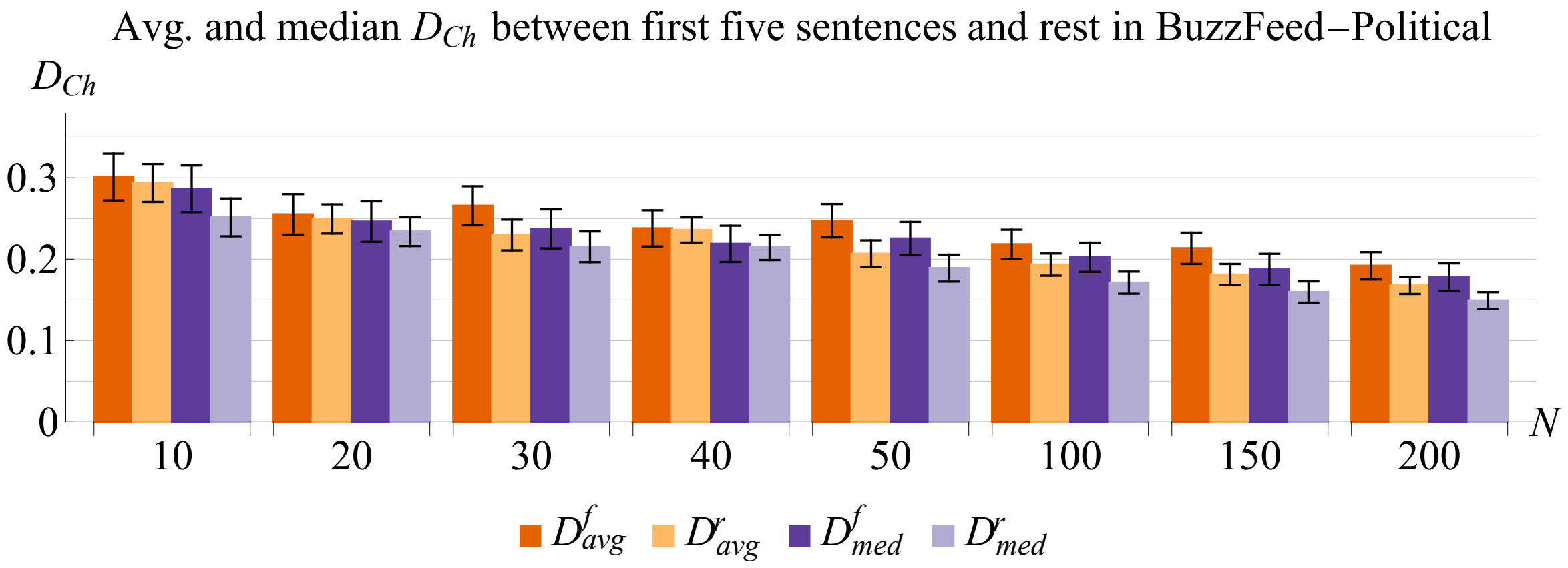

(b) Results for BuzzFeed-Political

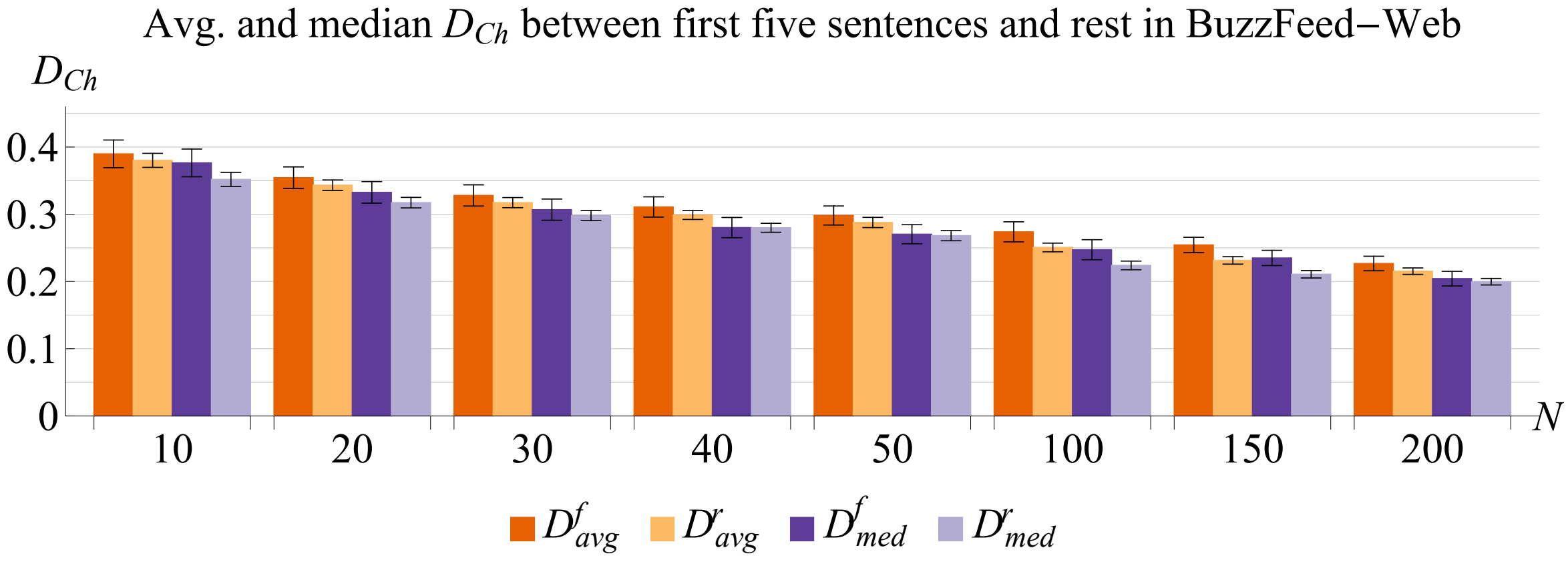

(c) Results for BuzzFeed-Web

FIGURE 4.2: Average and median Chebyshev distances in fake and real news, when comparing topics in the first five sentences to the rest of each article. Error bars show 95% confidence interval.

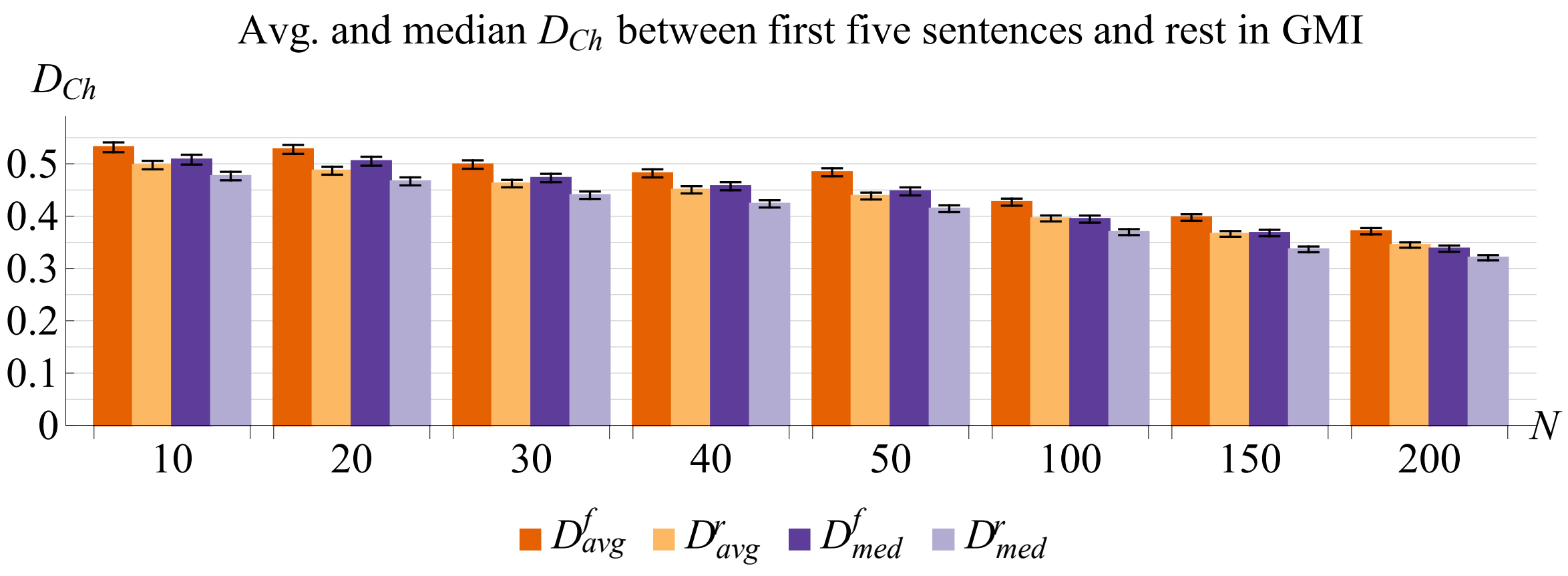

(d) Results for GMI

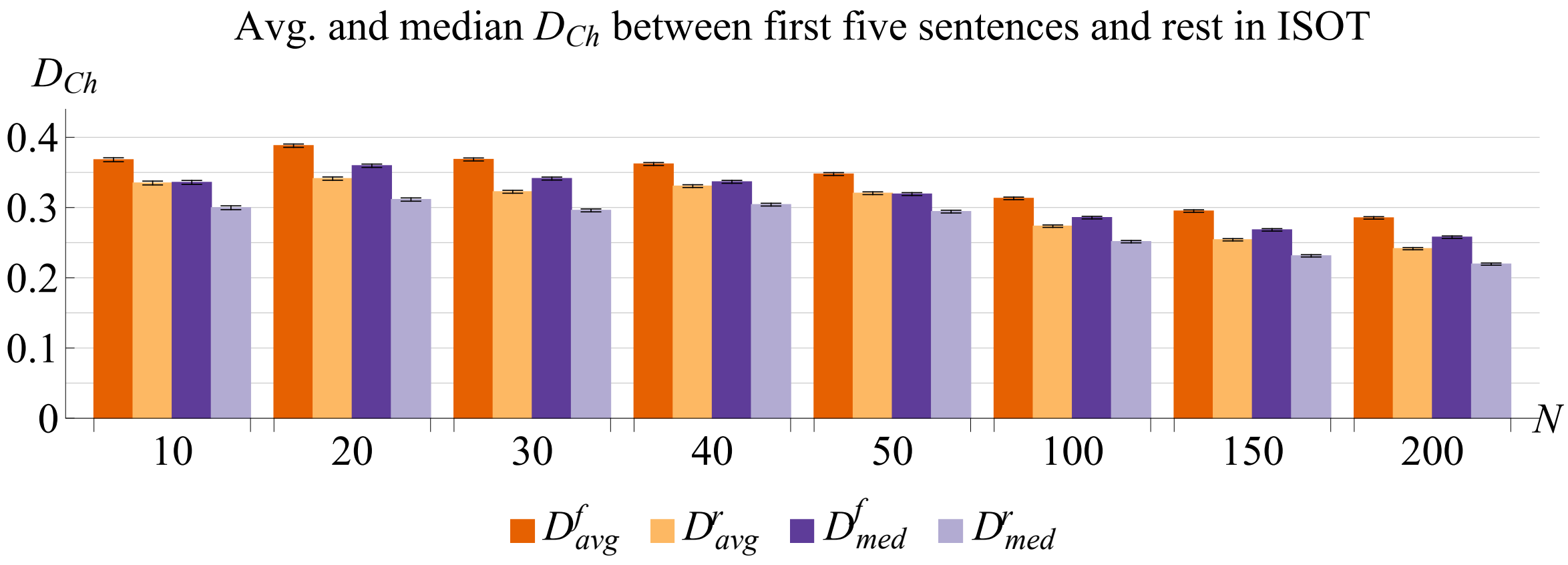

(e) Results for ISOT

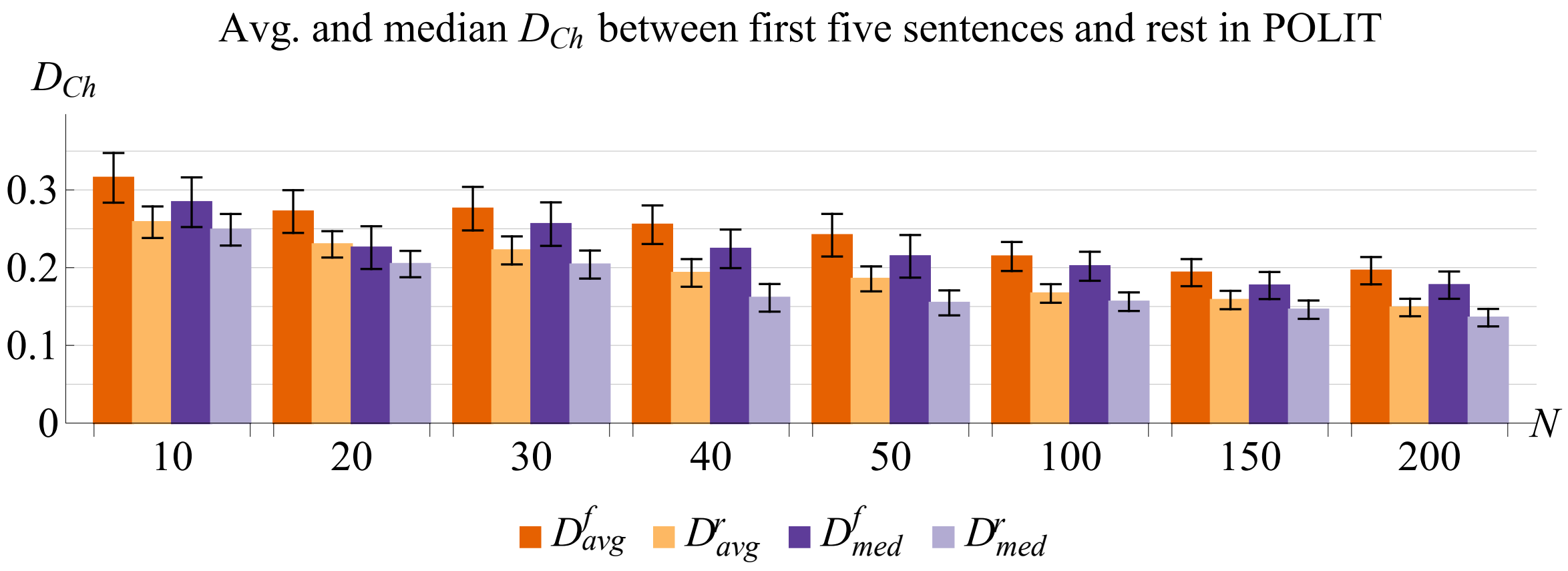

(f) Results for POLIT

FIGURE 4.2: Average and median Chebyshev distances in fake and real news, when comparing topics in the first five sentences to the rest of each article. Error bars show 95% confidence interval. (Cont.)

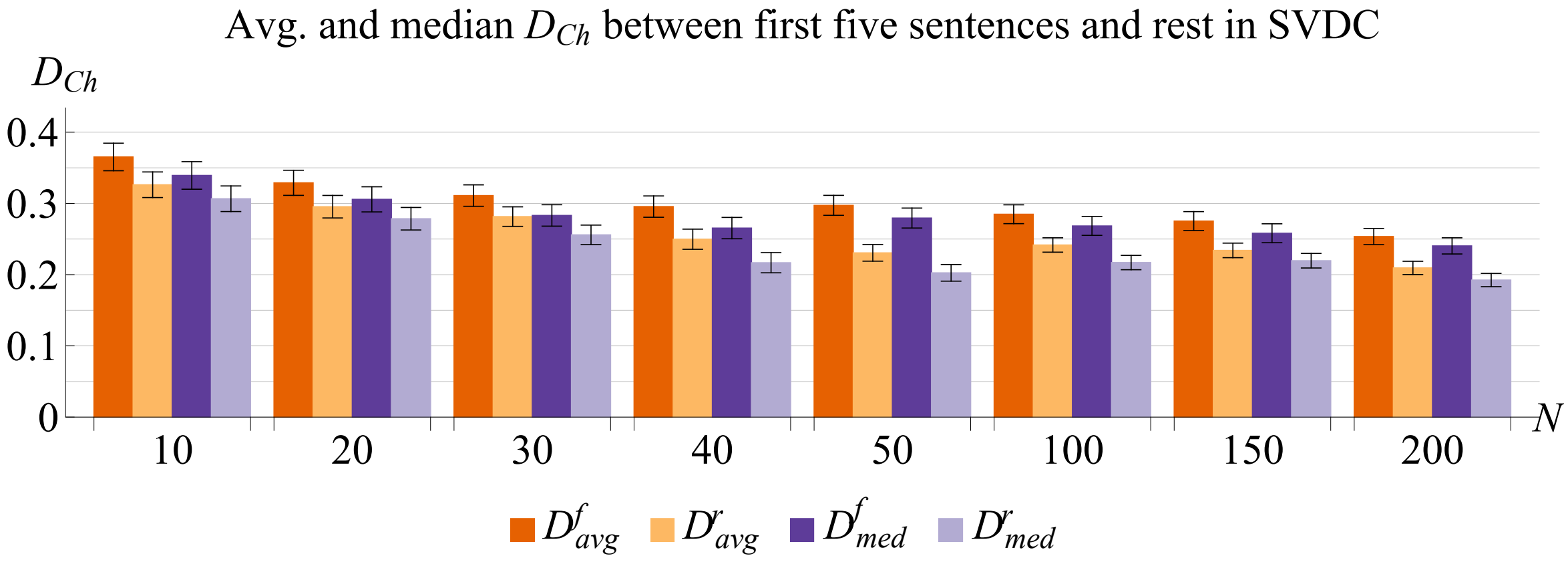

(g) Results for SVDC

FIGURE 4.2: Average and median Chebyshev distances in fake and real news, when comparing topics in the first five sentences to the rest of each article. Error bars show 95% confidence interval. (Cont.)

It is worth highlighting the diversity of datasets used here, in terms of domain, size, and the nature of articles. For example, the fake and real news in the SVDC dataset have a very similar structure. Both types of news were mostly written with the motivation to inform the reader of conflict-related events that took place across Syria. However, fake articles are labelled as such primarily because the reportage (*e.g.*, on locations and the number of casualties recorded) in them is insufficiently accurate.

▸ TO GAIN INSIGHT into possible causes of greater deviation in fake news, the five most and least diverging fake and real articles (according to $D_{Ch}$) were qualitatively inspected. A small set of low and high numbers of topics ($N \leq 30$ and $N \geq 100$) were also compared. It was observed that fake openings tend to be shorter, vaguer, and less congruent with the rest of the text. By contrast, real news openings generally give a better narrative background to the rest of the story. Horne and Adali (2017) reported similar findings regarding the comparison in length, between authentic and false news articles: *i.e.*, the former is generally longer than the latter, as shown in Table 4.1. Furthermore, the same study also found that fake articles are highly redundant and contain less substantial information.

Although the writing style in fake news is sometimes unprofessional, this is an unlikely reason for the higher deviations in fake news. Additionally, both fake and real news may open with the most newsworthy content, and expand on it with more context and explanation. This is the conventional hierarchical structure of news,[139] as discussed in §2.2.

[139] van Dijk (1983), "Discourse Analysis: Its Development and Application to the Structure of News"

Indeed, in this study, it was observed that real news tends to have longer sentences, which give more detailed information about a story and are more narrative. It can be argued that the reason behind this is that fake articles are designed to get readers' attention, whereas legitimate ones are written to inform the reader. For instance, social media posts which include a link to an article are sometimes displayed with a short snippet of the article's opening text or its summary. This section can be designed to capture readers' attention.

It was also observed that fake articles include more question and exclamation marks, as well as words and phrases in all capitals. Although this is inconsequential to forming topics, it supports the claim that false news is written in an attention-grabbing style. While Horne and Adali (2017) state that punctuation is less likely to be found in fake news text, Rubin et al. (2016) suggest that it is a differentiating factor between fake and real news. Punctuation marks, including question and exclamation marks, have also been used as a feature in fake news detection.[140]

[140] Pérez-Rosas et al. (2018), "Automatic Detection of Fake News"

Furthermore, it is conceivable that a bigger team of people working to produce a fake piece may contribute to its vagueness. They may input different perspectives that diversify the story and make it less coherent. This may be compared with real news, whereby there is one professional writer, perhaps two, and therefore, better coherence.

### 4.7.1 *Quantitative analysis of coherence and perplexity*

The observation of greater thematic deviation was further explored experimentally. The qualitative findings previously discussed can be more reliably verified through quantitative analyses, using empirical measures for *topic coherence*.[141]

Topic coherence assigns a score to each topic by evaluating the semantic similarity between top words in the topic.[142] It is capable of reflecting people's perception of latent topics in a given text.[143] Thus, topic coherence is adopted here as an *indicator* of the amount of vagueness in an article. The intuition behind this is that topics with high coherence constitute words which allow a reader to infer the general topic(s) the text is about. Conversely, those with very low coherence are hardly interpretable,[144] and hence, are likely to arise from vaguer text.

Topic coherence measures fall into two groups:

1. Intrinsic measures, which capture model semantics and are based on human evaluation of topics' interpretability.

2. Extrinsic measures, which indicate how good a topic model is at performing predefined tasks such as classification.

As intrinsic measures are based on human evaluations, they are more apt for indicating how a person might assess the coherence of an article they are reading. Moreover, intrinsic measures have been shown to correlate better with human judgement.[145] Therefore, one such measure called UMass[146] is used. This is defined in Equation 4.8 as was done by Stevens et al. (2012). From a set of top words used to describe a topic, UMass measures the extent to which a common word is a good predictor of a less common word on average.[147]

$$score_{UMass}\left(v_i, v_j, \epsilon\right) = log\,\frac{D\left(v_i, v_j\right) + \epsilon}{D\left(v_j\right)} \tag{4.8}$$

where: $D(x)$ = number of documents which contain word $x$

$D(x, y)$ = number of documents containing words $x$ and $y$

$\epsilon$ = smoothing factor that ensures $score_{UMass}$ is a real number

Topic coherence was evaluated in two ways: (i.) the openings of fake ($S1$) and authentic articles ($S2$); and (i.) the whole articles. In both cases, the numbers of topics ($N$) studied are $10, 20, \ldots, 140, 150, 200$.[148]

▸ $S2$ ARTICLES IN the AMT+C dataset have greater coherence than $S1$ ones. This becomes more apparent when $N \geq 40$. However, focusing on the opening sections, it can be seen that $S1$ opening sentences are only slightly more coherent than $S2$ ones. This means that although $S2$ in this dataset are more topically coherent, the opening sections of $S1$ are more coherent.

[141] This should not be confused with the concept of *thematic coherence* earlier introduced in this chapter. Whereas the former is used in this thesis to denote consistency in the subject(s) discussed throughout a news article, the latter is a measure for evaluating topics in general.

[142] Stevens et al. (2012), "Exploring topic coherence over many models and many topics"

[143] Blair et al. (2019), "Aggregated topic models for increasing social media topic coherence"

[144] Röder et al. (2015), "Exploring the space of topic coherence measures"

[145] Chang et al. (2009), "Reading tea leaves: How humans interpret topic models"

[146] Mimno et al. (2011), "Optimizing semantic coherence in topic models"

[147] Mimno et al. (2011), Hemmatian et al. (2019)

[148] Note that a wider range of $N$ is used here, compared with Algorithm 2.

In the BuzzFeed dataset, it can be seen that $S2$ articles are more coherent than $S1$ articles, both for opening and whole texts. For the openings in BuzzFeed Political, $S2$ articles are only slightly more coherent. Nonetheless, whole $S2$ articles are noticeably more coherent than $S1$ articles. Considering the opening sections, it is clear that the $S2$ coherence scores are generally—though only marginally in most cases—higher than those of $S1$.

Figure 4.2 shows UMass scores for the first five sentences of $S1$ and $S2$ articles, calculated over the training set (a combination of all $S1$ and $S2$ full articles). Higher values indicate higher topic coherence, *i.e.*, words associated with each topic in that model are more likely to co-occur. As expected, more topics are generally less coherent than fewer ones.

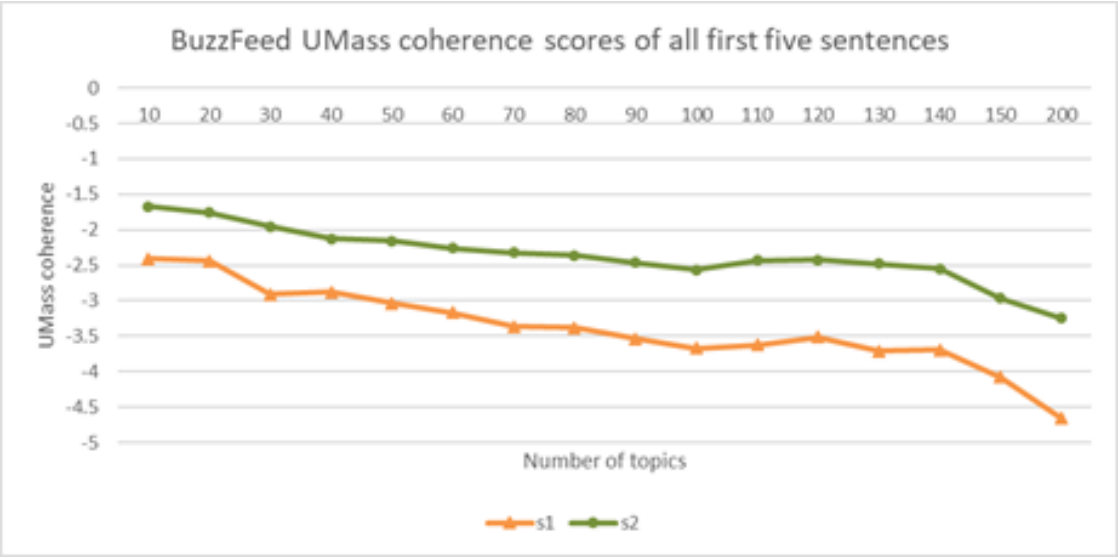

(a)

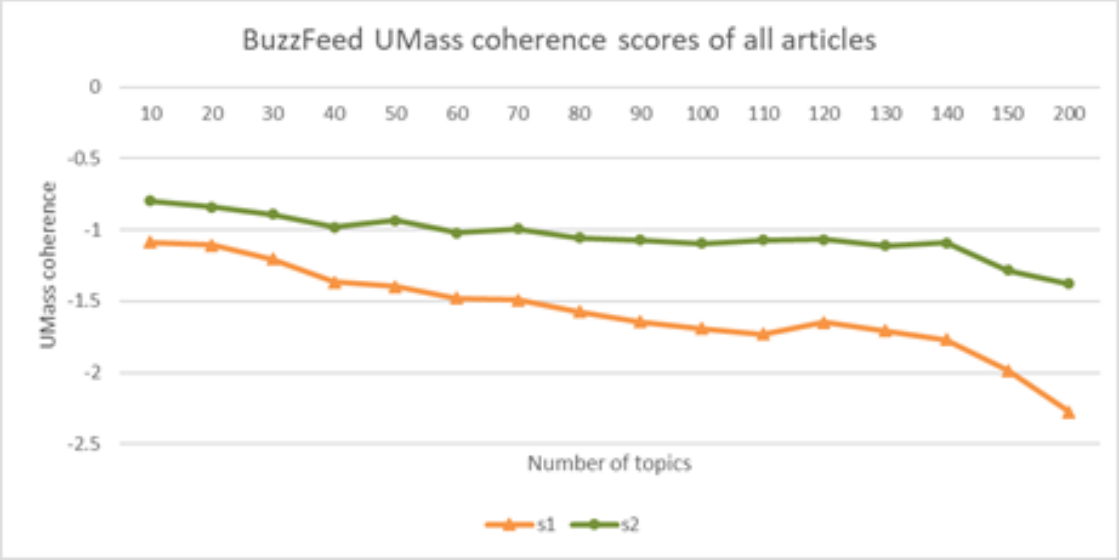

(b)

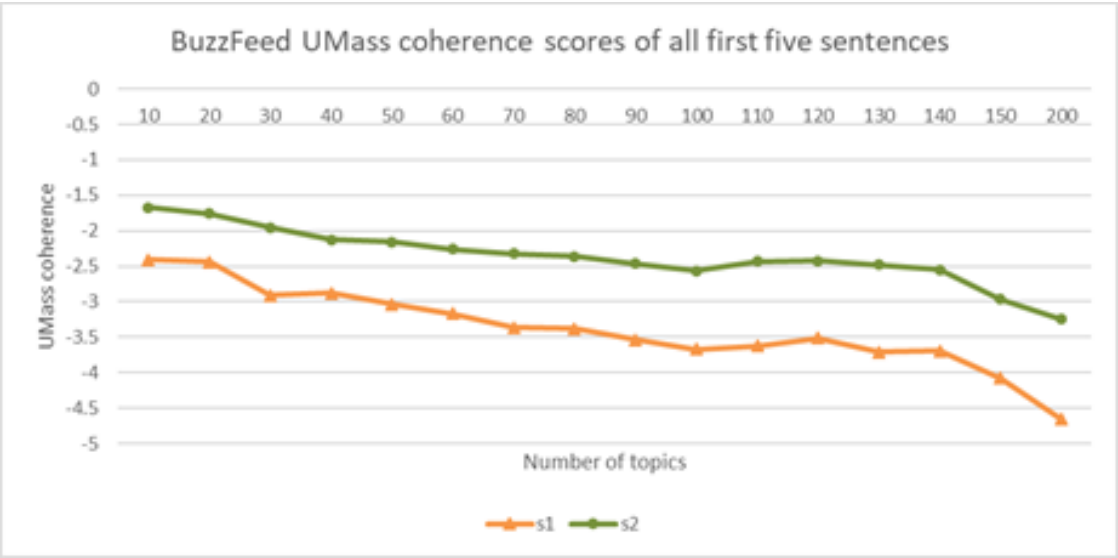

(c)

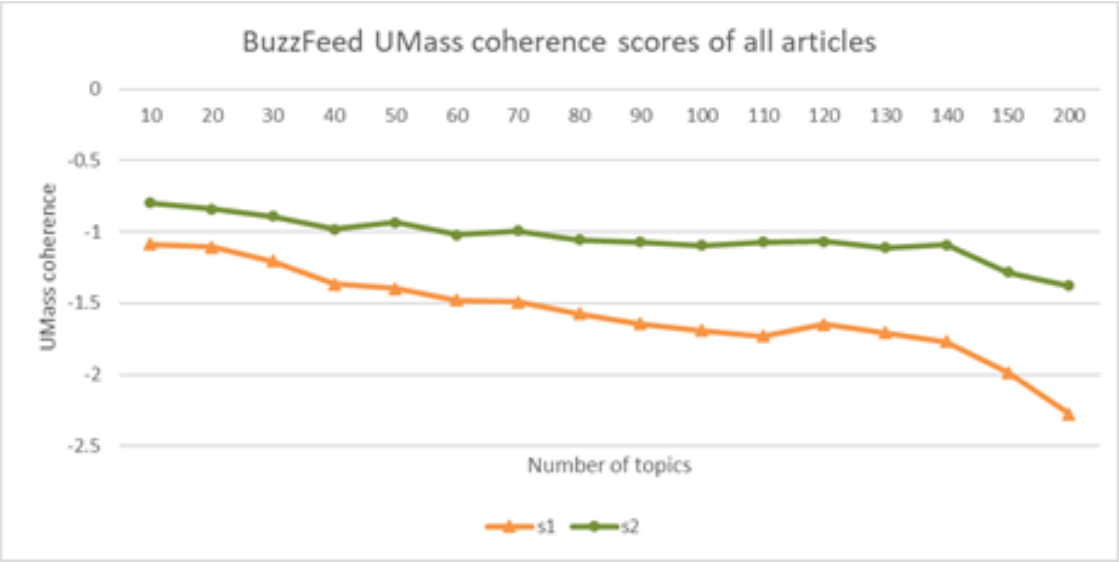

(d)

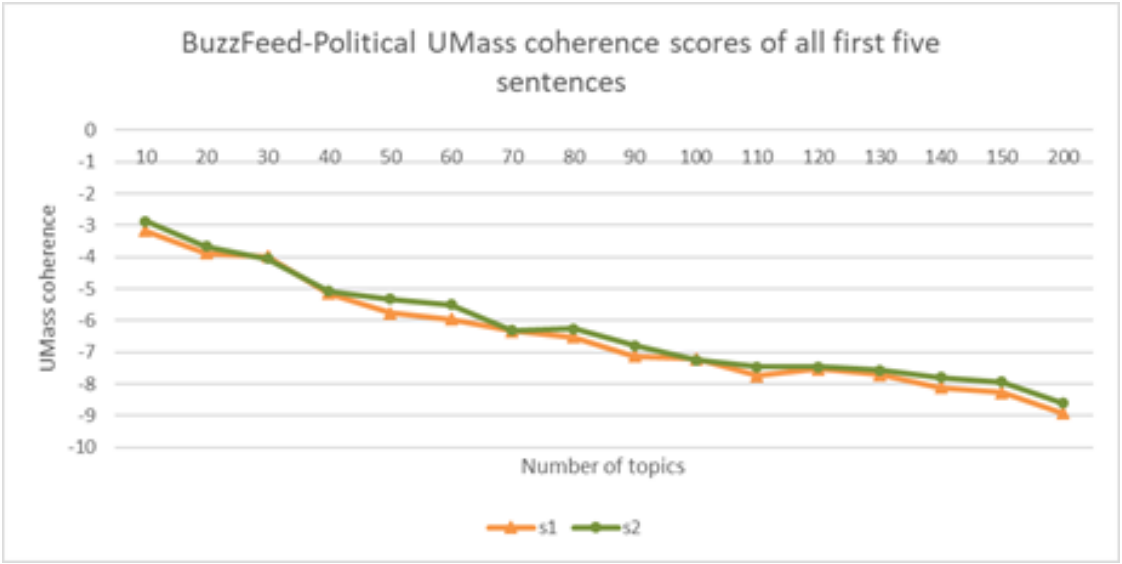

(e)

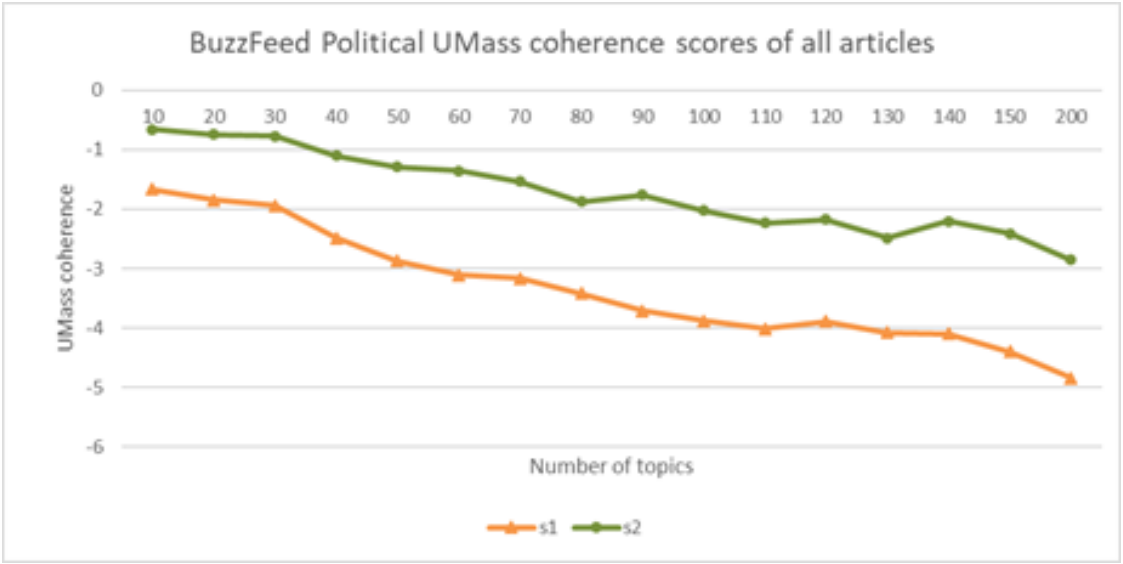

(f)

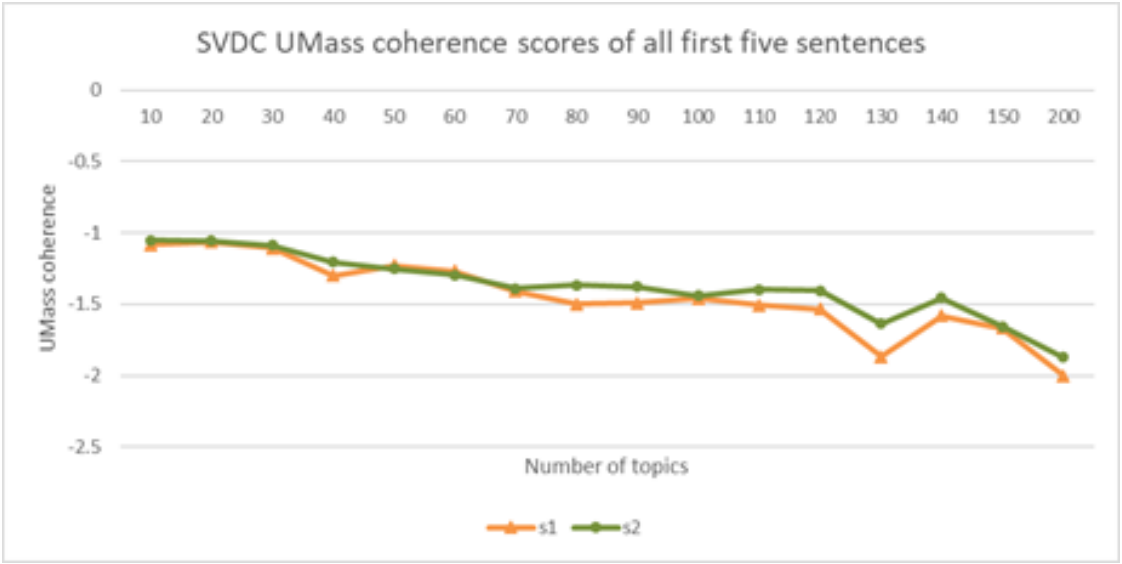

(g)

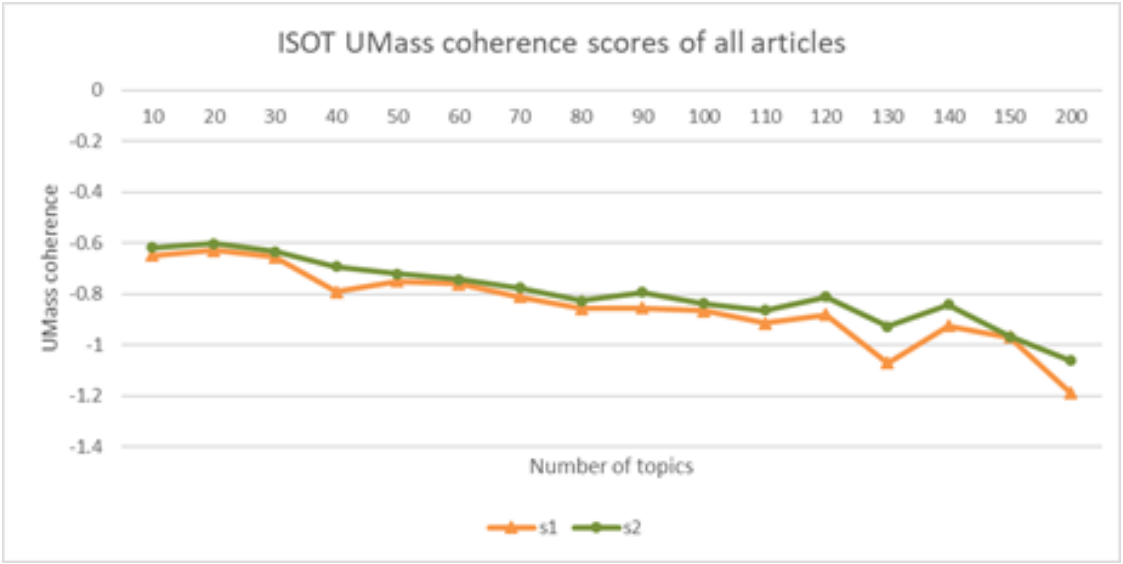

(h)

FIGURE 4.2: UMass topic coherence scores

▸ IN SUMMARY, THE topic coherence of authentic news is generally greater than that of misinformation, in all datasets except AMT+C. This is the case in the articles' opening sections, and when considering the whole article. The UMass coherence scores suggest that true articles are less vague, compared with fake ones, as they form more coherent topics. This corroborates earlier qualitative findings on the coherence of real and fake articles. Nonetheless, manual inspection of the top words in each topic may still be required. While some datasets show a clear distinction between false and true articles' coherence scores, the disparity is not clear in others.

Ideally, applying insights from Röder et al. (2015), a different topic coherence score called $C_V$, should be used. The authors found this to be the best amongst topic coherence measures in their study. The $C_V$ score uses Normalized Pointwise Mutual Information and cosine similarity measure (see Equation 4.1) in its workings. One drawback of the $C_V$ score is its runtime—it takes more than twenty times longer to run compared with UMass. In any case, the performance of UMass suffices in this exploration.

The datasets analysed here cover a broad range of themes and contain articles with different structures and writing styles. Furthermore, their constituent false and real articles have sentences of varying lengths and vocabularies of varying sizes. These findings show that regardless of the individual attributes of datasets, fake news articles appear to have some high-level features which can be used to systematically tell them apart from real news.

## 4.8 CONCLUSION

Fake news and deceptive stories tend to open with sentences which may be incoherent with the rest of the text. It is worth exploring if the consistency of fake and real news can distinguish between the two. Accordingly, the thematic deviations of seven cross-domain fake and real news, using topic modelling were investigated. The findings presented in this chapter suggest that the opening sentences of fake articles topically deviate more from the rest of the article, in contrast to real news. The next step is to find possible reasons behind these deviations through in-depth analyses of topics. In conclusion, this paper presents valuable insights into thematic differences between fake and authentic news, which may be exploited for fake news detection.

### 4.8.1 *Future work*

Future work can extend this research in two main ways. Firstly, experimenting with topic modelling methods other than LDA may improve the results. One example that has been demonstrated to outperform LDA in the task of learning insightful topics is `top2vec`[149]. This topic modelling algorithm combines document and word vectors to find topics. Both are commonly used features for fake news detection. With `top2vec`, topic vectors are indicative of the semantic similarity between documents. Additionally, it automatically finds the optimal number of topics. By comparison, this is done iteratively with LDA, by evaluating metrics such as *perplexity* across a varying number of topics.

[149] Angelov (2020), "Top2Vec: Distributed Representations of Topics"

Secondly, other techniques beyond splitting an article into multiple parts can also be investigated. Ideally, the feature extraction method should be resilient to minor alterations in the news text. It is worth stating that experiments in parallel with this idea were also carried out in this research. For example, the semantic coherence between the extractive summaries and opening paragraphs of articles was evaluated, using word embeddings obtained from Bidirectional Encoder Representations from Transformers (BERT).[150] Further experiments were carried out to find the amount of overlap between the summaries and opening paragraphs, and the positions of sentences in each article, which constitute its extractive summary. This branch of experiments did not show semantic coherence—found in this particular way—to be a robust marker of misinformation detection. Nonetheless, it can be further investigated in future research.

[150] Devlin et al. (2019), "BERT: Pre-training of Deep Bidirectional Transformers for Language Understanding"

# 5

# CLUSTERING AND CLASSIFICATION USING TOPICS

The various experiments in the preceding chapters culminated in demonstrating that topics can be used to tell apart fake from authentic news texts. This can be achieved using the unique representation of topics from the opening and remainder sections of news articles. This conclusion was arrived at following statistical tests, which showed that there is some evidence that fake and real news may differ thematically.

In this study, the utility of topic representations using simple methods is analysed. First, through unsupervised learning, clustering; and second, through supervised learning, classification. The most straightforward possible ML methods are selected here because the goal is to evaluate the utility of these representations. It should be noted that the topic distributions themselves are used as features in this case, rather than the divergence scores calculated from them. Therefore, the experiments in this chapter are not based on the calculated variance between topics in the opening and remainder parts of fake and real articles per se. Nonetheless, this information is still retained in the distributions.

As related in Chapter 2, both approaches (*i.e.*, clustering and classification) have been used by several works in the literature on misinformation detection. Their advantages and shortcomings, particularly in the context of misinformation, are also discussed therein.

Clustering is done using the $K$-means algorithm. Having performed feature extraction in an unsupervised way, it is additionally beneficial to further detect misinformation likewise. A wide range of classifiers was experimented with, including Decision Trees, Random Forest and SVM.

## 5.1 CLUSTERING

This experiment was carried out on whole topic distributions from the opening and remainder sections of articles, as well as their reduced 2D vectors (called the *Aggregate* method here).

PROBLEM DEFINITION: The $K$-means algorithm requires us to specify the number of clusters outputted, $K$. The evaluation for this experiment, detailed later in this subsection, will focus on determining whether the clusters are becoming *pure* or not.

DATASETS AND DATA: The datasets used in this experiment are the same as in Chapter 4[151] except for GMI because it was only considered as a first step in the other experiments. The same topic data was used too, except that it is shortened here, which is to say, $N = \{10, 20, 30, 40, 50\}$.[152] Therefore, there is a $(1 \times 150)$ topic distribution for each article.

[151] See §4.6.2.

[152] See Algorithm 2 in §4.5.2.

CONJECTURES AND BASELINES: Three baselines were formulated to assess and compare the utility (based on the clustering metric used hereinafter) of extracting topic features from articles in different ways. For example, whether using reduced dimensions of the topic features improve the clustering. Or, if extracting features from two sections of an article is any better than taking topics from the entire text.

**Conjecture 1:** *Topics extracted from the opening and remainder sections of articles improve clustering (Aggregate method), compared with topics extracted from the whole document.*

*Baseline 1* was created to evaluate Conjecture 1. Here, topics (10, 20, 30, 40, and 50) are extracted from entire documents, instead of from their openings and remainders. Therefore, each document is represented as a single 150-dimensional vector. $K$-means (with $K = 2$ and the maximum number of iterations set to 500) is run on the original 150D topic distribution, as well as on their reduced dimension (2D) vectors. The projection-based dimensionality reduction methods experimented with are: Autoencoder, tSNE, and Uniform Manifold Approximation and Projection (UMAP). The component-based methods used are: Linear, NMF, PCA, and Singular Value Decomposition (SVD).

**Conjecture 2:** *Combining multiple topic distributions also improves clustering, compared with individual topics on their own.*

*Baseline 2* tests Conjecture 2: individual topic distributions (for 10, 20, 50, and 100 topics) from the opening and remainder sections form the clustering data. For example, the vector for 10 topics will be a $(1 \times 20)$ vector.

**Conjecture 3:** *Clustering performs better than simply assigning examples to classes randomly.*

*Baseline 3* tests Conjecture 3: here, the quality of clustering is calculated based on random assignment to each class, *i.e.*, half of each type of news forms a cluster.

EVALUATION: There are two main ways to evaluate the quality of clustering: *internal* and *external* criteria.[153] Ideally, articles within a given cluster should be similar (inter-cluster similarity), and those from different clusters should be dissimilar (intra-cluster similarity). This is the basis of the internal criterion. On

[153] Manning et al. (2008), *Introduction to Information Retrieval*

the other hand, the external criterion requires a benchmark created by people who can expertly categorise each item. As it takes account of the nuances of a given application, the external criterion is more reliable, especially if a model is to be deployed in the real world. It is applicable in this case since the data is fully labelled. One such criterion, *Purity*, was used in this experiment to evaluate the aforementioned baselines. Following Manning et al. (2008), it can be defined as:

$$purity\ (\Omega, C) = \frac{1}{D} \sum_{k=1}^{K} \max_{j=1\ldots J} \left|\omega_k \cap c_j\right| \quad (5.1)$$

where: $\Omega = \{\omega_1, \omega_2, \ldots, \omega_K\}$ is a set of clusters

$C = \{c_1, c_2, \ldots, c_J\}$ is a set of classes

$D$ = total number of documents

$\omega_k$ = set of documents in $\omega_k$

$c_j$ = set of documents in $c_j$

The most frequent class of articles–fake or real–in a cluster is assigned as the label of that cluster. Therefore, the accuracy of the clustering is the sum of fractions of correct assignments in each cluster. This summarises purity as a metric.

Figure 5.0 shows plots of the concatenated 300D data, with their dimensions reduced to 2D, using the *Linear* method for dimensionality reduction in Wolfram Mathemematica.[154] Each data point is coloured according to its class. It can be observed from the figure that reducing the dimensions removes superfluous information while preserving the essential information that apparently differentiates the two types of news. Significant variations can be observed in the topic distributions of fake and real news in all datasets, except for BuzzFeed.

[154] Wolfram Research (2021), *"Linear" (Machine Learning Method)*

Put simply, when it comes to both fake and real articles talking about the same subject, even across various domains, there are noticeable variations in how they approach the topic in the beginning and the rest of the articles. This observation is in agreement with the outcomes of previous experiments, that topics are an effective feature for detecting misinformation. Naturally, the boundaries between the clusters are not clear-cut. For example, there is a discernible overlap among the clusters in the ISOT dataset, indicating the presence of both counterfeit and genuine articles that nsimilarly narrate stories. However, notable differences can be seen when examining multiple fabricated and legitimate articles in general.

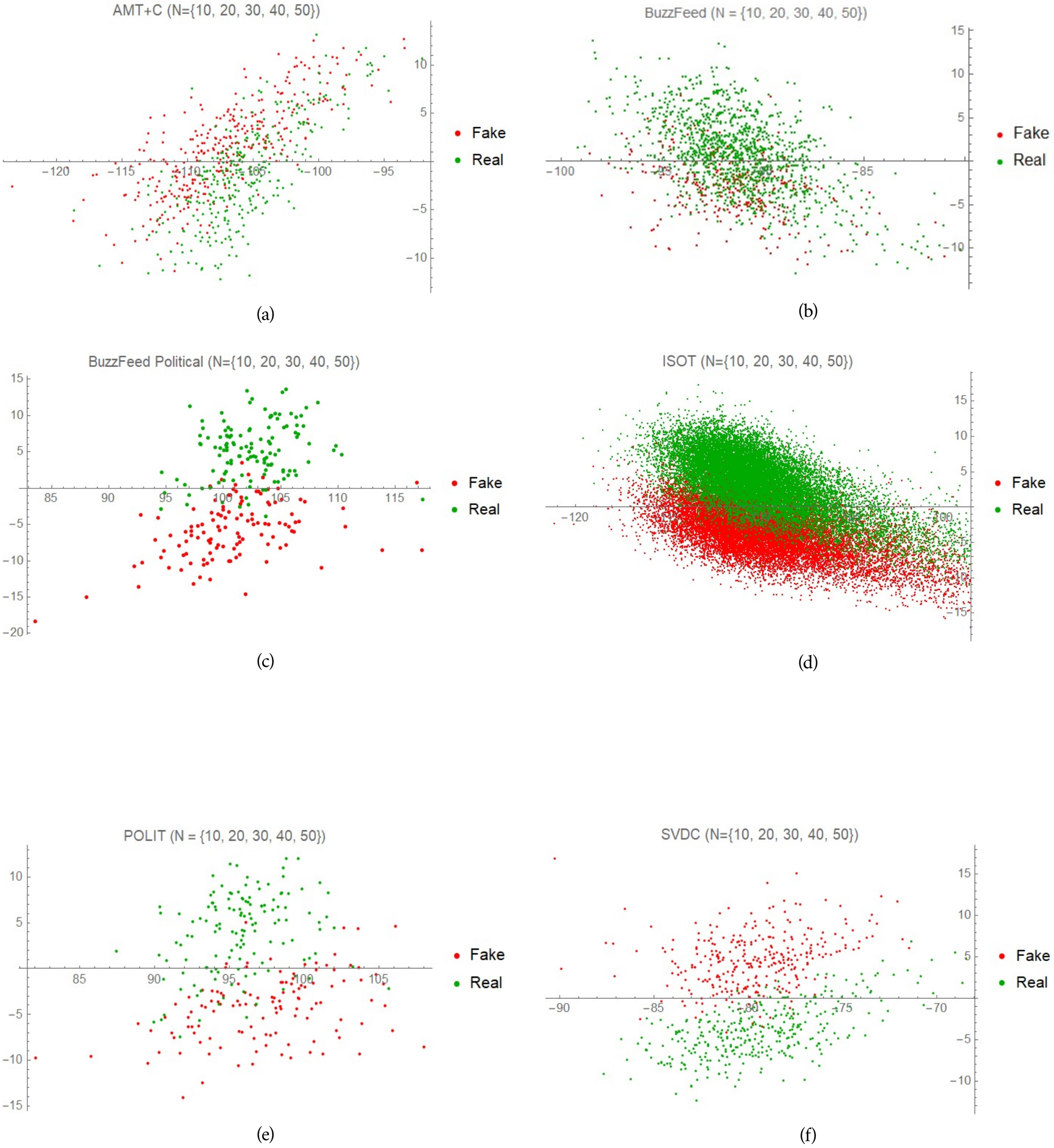

Figure 5.0: 2D plots of dimension-reduced topic distributions for datasets used.

**RESULTS AND DISCUSSION:** Table 5.1 shows results for the evaluation of Baseline 1 on the different data dimensions experimented with. Two values, 100 and 200, were used for the tSNE parameter *perplexity*.[155] For UMAP, two values, 50 and 100, were used for the number of neighbours. Table 5.2 shows results for the evaluation of Baseline 2.

[155] Note that this notion of perplexity is different from the one discussed in 4.7.1.

| Dimension | AMT+C | | Buzzfeed | | Buzzfeed-Political | | ISOT | | Polit | | SVDC | |
|---|---|---|---|---|---|---|---|---|---|---|---|---|
| | Agg | B1 | Agg | B1 | Agg | B1 | Agg | B1 | Agg | B1 | Agg | B1 |
| 150D / 300D | 0.5179 | 0.5367 | 0.7858 | 0.7858 | 0.5205 | 0.5226 | 0.5679 | 0.5346 | 0.6211 | 0.5234 | 0.7169 | 0.5452 |
| Autoencoder | 0.5179 | 0.5320 | 0.7858 | 0.7858 | 0.5220 | 0.6872 | 0.5453 | 0.5345 | 0.5234 | 0.6523 | 0.6221 | 0.5301 |
| 2D, Linear | 0.5133 | 0.5211 | 0.7858 | 0.7858 | 0.8150 | 0.7078 | 0.5897 | 0.5579 | 0.7031 | 0.6602 | 0.5949 | 0.5301 |
| 2D, NMF | 0.5086 | 0.5226 | 0.7858 | 0.7858 | 0.5410 | 0.5225 | 0.5844 | 0.5409 | 0.5820 | 0.5234 | 0.7334 | 0.5557 |
| 2D, PCA | 0.5195 | 0.5413 | 0.7858 | 0.7858 | 0.6271 | 0.5226 | 0.5346 | 0.5346 | 0.6211 | 0.5234 | 0.7500 | 0.5437 |
| 2D, SVD | 0.5117 | 0.5288 | 0.7858 | 0.7858 | 0.5574 | 0.5226 | 0.5608 | 0.5346 | 0.6094 | 0.5234 | 0.7425 | 0.5422 |
| 2D, TSNE [$p = 100$] | 0.5055 | 0.5273 | 0.7858 | 0.7858 | 0.9016 | 0.5226 | 0.5591 | 0.5346 | 0.7648 | 0.5625 | 0.7892 | 0.5723 |
| 2D, TSNE [$p = 200$] | 0.5179 | 0.5273 | 0.7858 | 0.7858 | 0.8893 | 0.5597 | 0.5826 | 0.5346 | 0.7617 | 0.5430 | 0.8298 | 0.5723 |
| 2D, UMAP [$nn = 50$] | 0.5226 | 0.5445 | 0.7858 | 0.7858 | 0.5205 | 0.5597 | 0.5828 | 0.5346 | 0.5742 | 0.5391 | 0.5301 | 0.5768 |
| 2D, UMAP [$nn = 100$] | 0.5304 | 0.5445 | 0.7858 | 0.7858 | 0.5205 | 0.5597 | 0.5815 | 0.5346 | 0.5547 | 0.5391 | 0.6054 | 0.5768 |

Table 5.1: Purity scores for Baseline 1 (B1) and Aggregate (Agg) methods. $p$ = perplexity, $nn$ = number of neighbours.

| DATASET | $T = 10$ | $T = 20$ | $T = 50$ |
|---|---|---|---|
| AMT+C | 0.5101 | 0.5070 | 0.5663 |
| BuzzFeed | 0.7858 | 0.7858 | 0.7858 |
| BuzzFeed-Political | 0.5820 | 0.5328 | 0.5697 |
| ISOT | 0.5373 | 0.5595 | 0.5346 |
| POLIT | 0.5898 | 0.5430 | 0.5860 |
| SVDC | 0.5979 | 0.5919 | 0.7063 |

Table 5.2: Purity scores for Baseline 2

With regards to Baseline 1, clustering on combined (*i.e.*, concatenated) topic distributions from the opening and remainder of documents (Aggregate method) generally performs better, than clustering on whole documents. The only exceptions to this are in AMT+C where the opposite result is observed; and BuzzFeed Political, where there is no significant difference, probably owing to class imbalance. As for Baseline 2, the results show that the combination of multiple topic distributions also gives better clustering performance, than individual topics, except in BuzzFeed Political. Baseline 3 results show that clustering outperforms random assignment.

| DATASET | BASELINE 1 | | BASELINE 2 | | | BASELINE 3 | AGGREGATE | |
|---|---|---|---|---|---|---|---|---|
| | 150D | 2D | $T = 10$ | $T = 20$ | $T = 50$ | | 300D | 2D |
| AMT+C | 0.5367 | 0.5273 | 0.5101 | 0.5070 | **0.5663** | 0.5023 | 0.5179 | 0.5179 |
| BuzzFeed | 0.7858 | 0.7858 | 0.7858 | 0.7858 | 0.7858 | 0.7864 | 0.7858 | 0.7858 |
| BuzzFeed Political | 0.5226 | 0.5597 | 0.5820 | 0.5328 | 0.5697 | 0.5204 | 0.5205 | **0.8893** |
| ISOT | 0.5346 | 0.5346 | 0.5373 | 0.5595 | 0.5346 | 0.5346 | 0.5679 | **0.5826** |
| POLIT | 0.5234 | 0.5430 | 0.5898 | 0.5430 | 0.5860 | 0.5234 | 0.6211 | **0.7617** |
| SVDC | 0.5452 | 0.5723 | 0.5979 | 0.5919 | 0.7063 | 0.5301 | 0.7169 | **0.8298** |

Table 5.3: Comparison of clustering purity scores. tSNE with $p = 200$ is used to obtain 2D data. The best purity scores for each dataset are in bold.

All results for clustering are summarised in Table 5.3. They show that clustering on a combination of multiple topic distributions from the opening and remainder of articles generally performs better than the formulated baselines. The exceptions are AMT+C and BuzzFeed. The latter dataset has a significant class imbalance (331 fake articles and 1,214 real ones), which is a likely explanation for the absence of variation in its results. Dimensionality reduction appears to retain important thematic information and improve clustering. Misinformation detection is typically done using a variety of high-level and low-level (*shallow*) features. For example, combining thematic features with semantic or linguistic ones may yield improvements in the current clustering results.

In conclusion, the clustering experiments presented in this chapter have demonstrated that features obtained through topic modelling may be exploited for misinformation detection. Crucially, unsupervised learning is advantageous for problems such as this. Therefore, the combination of multiple unsupervised learning methods, such as topic modelling, dimensionality reduction, and clustering, allows for an end-to-end unsupervised pipeline for detecting misinformation. However, such an implementation would not be without constraints. For though clustering algorithms such as $K$-means are efficient, topic modelling and dimensionality reduction can be time-consuming, depending on the size of the dataset.

In future work, the experimental methods can be improved. Firstly, as mentioned in Chapter 4, the topic modelling method for feature extraction can be improved. Secondly, other clustering methods can also be experimented with. In this research, spectral clustering was also considered, but $K$-means gave better results. In a semi-supervised scenario, rather than setting $K = 2$, articles may be clustered into three authentic, false, and indeterminable—and human experts can review and label items in the third group.

## 5.2 CLASSIFICATION

Classification was also applied to assess the effectiveness of topic representations as markers for distinguishing between authentic and false news. Classification is the most prominent ML method in the literature.

The models used are Decision Trees, Gradient Boosted Trees, Logistic Regression, Markov Model, Naive Bayes, kNN, Neural Network, Random Forest, and SVM. First, a dataset is trained using all types of classifiers simultaneously. To do this, the data is split 80% for training and validation, and 20% for testing. Multiple versions of some classifiers, with different parameters, were created and trained using the 80% portion. Next, the best classifier (based on loss) and its parameters are selected. Note that the test set was not used in selecting the hyperparameters of the classifiers, but only for testing later on.

Finally, this classifier is recreated, applying its parameters, to train the dataset afresh, using five-fold cross-validation. Classification was done using the Wolfram Language `Classify`[156] function. Unless stated otherwise, the default parameters for all classifier types were used.

[156] Wolfram Research (2021), *Classify*

DATASETS AND DATA: The datasets used here are the same as for clustering.[157] This time though, an additional dataset, FakeNewsNet,[158] was also used. This dataset contains 4,443 and 13,433 fake and real articles, respectively. Similar to

[157] See 4.6.2, Table 4.1 for more information on datasets.

[158] Shu et al. (2020), https://github.com/KaiDMML/FakeNewsNet

FakeNewsNet, the BuzzFeed dataset also has a significant class imbalance—with 331 and 1,214 fake and real articles, respectively). These two datasets were balanced, by randomly sampling the bigger class to select the same number of articles from the smaller one. Therefore, the final FakeNewsNet and BuzzFeed datasets used had 4,443 and 331 articles, respectively, in both classes.

The original representation used for clustering contained $N = \{10, 20, 30, 40, 50\}$ topics, extracted from the opening and remaining text of each article. Therefore, for $m$ articles, the dimension of the data is $(5 \times 2 \times m)$. This data was modified to obtain the following topic data representations for classification:

1. The original topic representation, *i.e.*, a $(5 \times 2 \times m)$ tensor.
2. Flattened topic representation, *i.e.*, the original tensor concatenated to a 300D vector. The sum of $N$ dimensions for each section of the article is 150; concatenating the distributions for both sections gives 300D.)
3. Dimension reduced representation, *i.e.*, the 300D vector reduced to 2D using tSNE (with the parameter $perplexity = 200$).

RESULTS AND DISCUSSION: Table 5.4 shows results for the original topic representation. After the initial training, the best classifier on each dataset is a variant of a logistic regressor. Four datasets (FakeNewsNet, GMI, ISOT, and SVDC) were classified with more than 90% accuracy, and all others with more than 80%. Accuracy is satisfactory as an indicator of the overall classification performance because the datasets do not have huge class imbalances.

| DATASET | ACCURACY | F1 | PRECISION | RECALL |
|---|---|---|---|---|
| AMT+C[a] | 0.8393 | 0.8384 | 0.8380 | 0.8413 |
| BuzzFeed[b] | 0.8384 | 0.8376 | 0.8375 | 0.8411 |
| BuzzFeed-Political[c] | 0.8933 | 0.8912 | 0.8953 | 0.8906 |
| FakeNews Net[d] | 0.9487 | 0.9314 | 0.9307 | 0.9323 |
| GMI[e] | 0.9171 | 0.9169 | 0.9172 | 0.9168 |
| ISOT[f] | 0.9352 | 0.9346 | 0.9364 | 0.9335 |
| POLIT[g] | 0.8479 | 0.8451 | 0.8494 | 0.8462 |
| SVDC[h] | 0.9353 | 0.9345 | 0.9358 | 0.9343 |

[a] Logistic Regressor ($l2_reg = 100$), [b] Logistic Regressor ($l2_reg = 100$), [c] Logistic Regressor ($l2_reg = 10$), [d] Logistic Regressor ($l2_reg = 100$), [e] Logistic Regressor ($l2_reg = 100$), [f] Logistic Regressor ($l2_reg = 100$), [g] Logistic Regressor ($l2_reg = 10$), [h] Logistic Regressor ($l2_reg = 10$).

TABLE 5.4: Evaluation metrics for the best-performing classifier on each dataset, using the original representation. $l2_reg = L2$ regularisation

Table 5.5 shows the results for the dimension-reduced (2D) topic representation. The accuracy scores are generally lower compared with when using the original dimensions, but they remain at 80% or higher in the BuzzFeed Political, FakeNewsNet, POLIT, and SVDC datasets.

| DATASET | ACCURACY | F1 | PRECISION | RECALL |
|---|---|---|---|---|
| AMT+C[a] | 0.5289 | 0.5261 | 0.5307 | 0.5301 |
| BuzzFeed[b] | 0.6285 | 0.6274 | 0.6281 | 0.6283 |
| BuzzFeed-Political[c] | 0.8853 | 0.8838 | 0.8885 | 0.8845 |
| FakeNews Net[d] | 0.8004 | 0.8002 | 0.8017 | 0.8005 |
| GMI[e] | 0.6358 | 0.6285 | 0.6410 | 0.6322 |
| ISOT[f] | 0.6779 | 0.6778 | 0.6799 | 0.6803 |
| POLIT[g] | 0.8515 | 0.8495 | 0.8556 | 0.8597 |
| SVDC[h] | 0.9217 | 0.9208 | 0.9248 | 0.9195 |

[a] Random Forest ($leaf_size = 2, num_trees = 100$), [b] Logistic Regressor ($l2_reg = 1$), [c] Logistic Regressor ($l2_reg = 0.01$), [d] kNN ($nn = 20, method = KDtree$), [e] kNN ($nn = 50, method = KDtree$), [f] kNN ($nn = 500, method = KDtree$), [g] kNN ($nn = 20, method = KDtree$), [h] kNN ($nn = 10, method = KDtree$).

TABLE 5.5: Evaluation metrics for the best-performing classifier on each dataset, using the 2D representation. $l2_reg = L2$ regularisation, $nn$ = number of neighbours, $num_trees$ = number of trees.

Table 5.6 shows the results for the flattened 300D topic representation. They are generally better than the results of the 2D representation but not as good as those of the original representation. The accuracy scores only drop below 80% in the AMT+C and BuzzFeed datasets.

| DATASET | ACCURACY | F1 | PRECISION | RECALL |
|---|---|---|---|---|
| AMT+C[a] | 0.7628 | 0.7607 | 0.7624 | 0.7609 |
| BuzzFeed[b] | 0.7372 | 0.7360 | 0.7376 | 0.7372 |
| BuzzFeed-Political[c] | 0.8813 | 0.8768 | 0.8798 | 0.8765 |
| FakeNews Net[d] | 0.9102 | 0.9101 | 0.9102 | 0.9101 |
| GMI[e] | 0.9037 | 0.9036 | 0.9042 | 0.9034 |
| ISOT[f] | 0.9242 | 0.9238 | 0.9240 | 0.9236 |
| POLIT[g] | 0.8124 | 0.8111 | 0.8160 | 0.8140 |
| SVDC[h] | 0.9247 | 0.9241 | 0.9237 | 0.9260 |

[a] Gradient Boosted Trees ($leaf_size = 35, max_depth = 6, num_leaves = 110, l2_reg = 0, max_tr_rounds = 50, lr = 0.04$), [b] Logistic Regressor ($l2_reg = 1 \times 10^{-6}$), [c] Logistic Regressor ($l2_reg = 1$), [d] Logistic Regressor ($l2_reg = 100$), [e] Logistic Regressor ($l2_reg = 100$), [f] Logistic Regressor ($l2_reg = 10$), [g] Logistic Regressor ($l2_reg = 1$), [h] Gradient Boosted Trees ($leaf_size = 35, max_depth = 6, num_leaves = 110, l2_reg = 0, max_tr_rounds = 50, lr = 0.1$).

Table 5.6: Evaluation metrics for the best-performing classifier on each dataset, using the 300D representation.
$l2_reg = L2$ regularisation, $lr$ = learning rate, $max_tr_rounds$ = maximum training rounds, $num_trees$ = number of trees.

In summary, using the original topic representation, without dimension reduction or flattening, gives the best classification performance. However, this requires a noticeably longer training time compared with the two other representations. Dimension reduction appears to lose some important information that could enhance classification. Nonetheless, the retained information is adequate for classification on four of the datasets used. Topics are yet to be fully exploited as features in the misinformation detection literature. The classification results suggest that topic data representations can be used in different ways as features for this task. They can be used as standalone features, as demonstrated here, or combined with other kinds of features to improve the generalization ability of an ML model for misinformation detection. The latter is a worthwhile direction to pursue in future work.

## 5.3 CONCLUSION

In the previous chapter, topic representations of articles were introduced and explored as potentially viable features for misinformation detection. This chapter has presented experiments aimed at demonstrating the utility of topic representations. Simple implementations of clustering and classification have been used to separate authentic news articles from false ones. The results suggest that these representations are efficacious for this task.

In future work, a deeper exploration can be carried out using more sophisticated ML methods, to find out whether fake news detection can be improved even further using topic representations.

Part III

# EPILOGUE

# 6

# CONCLUSION

One way to detect misinformation is by manual fact-checking. This task is typically done by trained experts who tend to be accurate in spotting fake news. There are, however, a few issues with this approach. Firstly, the high quantity and speed of fake news make it difficult for fact-checkers to keep up. Secondly, continuous exposure to misinformation can be harmful to an individual, or even lead them to believe it is true. Finally, fact-checkers have a degree of subjectivity, which can lead to inconsistent results, especially when dealing with complex or controversial topics.

An alternative way of identifying fake news is by using computational methods. The application of NLP and ML techniques to do so is delivering increasingly better results so far. Supervised ML models are more commonly used to detect fake news, but they rely on a large amount of labelled data.

Fake news is becoming more cunning and even more similar to authentic news. Some of the features currently used to tell apart fake from real may not work well when the two are highly similar. For instance, it has been shown that representations based on writing style, are impractical when applied to machine-generated fake news. Therefore, there is also a need to innovate new text representations for distinguishing this type of news from legitimate news.

▸ This thesis makes the following main contributions to the current state of fake news detection:

1. It develops a novel approach for obtaining robust text features from news articles based on the topics they discuss. This is particularly useful in circumstances where labelled data is scant or unavailable.
2. It demonstrates the effectiveness of this new representation in distinguishing between fake and real news articles. This is shown using both supervised (classification) and unsupervised (clustering) ML. The latter approach helps in minimising the reliance on labelled datasets.

These contributions were achieved through the design and implementation of three main experiments:

1. This thesis explored word embeddings and sentiment features on short rumour and non-rumour texts. They did not show evidence of being

capable of differentiating between the groups. Nonetheless, this study can be extended in different ways to better understand semantic and sentiment relations between rumours and non-rumours.

2. It investigated the coherence in the themes discussed in the opening and remaining sections of fake and authentic news articles. The themes were represented in the form of latent topics. This study culminated in the development of a novel text representation, which showed evidence of being able to distinguish fake from real news.

3. It exploited the topic features for misinformation detection, using classification and clustering methods. Although these experiments are preliminary, the results are promising and to some degree, substantiate the efficacy of topic representations.

In its totality, this thesis contributes to researchers' ability to detect fake news computationally. However, further and deeper studies remain to be done.

## 6.1 FUTURE WORK

There now exist word embedding models that are more advanced than `word2vec` and `InferSent`. The experiments carried out in Chapter 3 can be extended to take advantage of state-of-the-art language models such as BERT. Language models pre-trained on short texts or tweets, such as `BERTweet`[159], may perform better than the ones used in this research. Furthermore, concerning sentiment, it has only explored limited categories of it (positive, neutral, and negative). In future work, an expanded range of emotions can be studied.

Other topic modelling tools may perform better than LDA for a study similar to the one in Chapter 4. Such a tool may generate better topics as assessed through intrinsic and extrinsic measures. This will increase confidence in ascertaining the robustness of thematic coherence as a text representation. For example, Egger and Yu (2022) carried out a detailed study of the strengths and weaknesses of different topic modelling methods for investigating OSN text data. In this work, LDA, NMF, `top2vec`,[160] and `BERTopic`[161] were compared. Additionally, other ways of extracting topics from articles can be explored. For instance, the articles could be split into multiple sections, rather than just two. This may improve the robustness of topic text representations, to make it more resilient to changes that can be made to increase the coherence of topics in fake news.

Chapter 5 presents preliminary yet promising results on the evaluation of the utility of topic representations. Simple classification and clustering algorithms were used for this. In future work, however, a more novel technique can be devised to fully take advantage of the features used in this work. For example, beyond evaluating the purity of clusters, a complete unsupervised fake news detection model can be created and its performance can be evaluated against the state-of-the-art.

159 Nguyen et al. (2020), "BERTweet: A pre-trained language model for English Tweets"

160 Angelov (2020), "Top2Vec: Distributed Representations of Topics"

161 Grootendorst (2022), "BERTopic: Neural topic modeling with a class-based TF-IDF procedure"

Finally, the text representations explored in this thesis can be combined, or used with those in other studies, to detect fake news. Experiments can be set up to compare the performances of the topic and stylometric features, or to evaluate the utility of their combination. Future research can adopt multimodal misinformation detection to combine the text representations presented in this work with features from other types of media. Especially image and video,[162] which are becoming increasingly easier to fabricate using tools such as Generative Adversarial Networks.[163] Future work may also evaluate if topic features are robust enough to accurately detect machine-generate fake news. This type of misinformation has the potential of becoming the primary way to create mis- and disinformation in the future.

[162] Mirsky and Lee (2021), "The Creation and Detection of Deepfakes"

[163] Goodfellow et al. (2014), "Generative Adversarial Nets"